\documentclass[a4paper,11pt]{article}

\usepackage{mathrsfs}
\usepackage{bm}
\usepackage{amsmath, amsthm, amssymb,amsfonts,latexsym,mathtools}
\usepackage{graphicx}
\usepackage{float}
\usepackage{lineno}

\usepackage{hyperref}
\usepackage{url}
\usepackage{hyperref}
\usepackage{color}

\newcommand{\Part}[3]{ \frac{ \partial^{#3} #1 }{ \partial #2^{#3} } }
\newcommand{\V}[1]{\bm{#1} } 

\newcommand{\Ave}[1]{\left\langle {#1} \right\rangle} 
 
\newcommand{\sgn}[1]{{\rm sgn}\left({#1} \right)}

\newcommand{\Extr}[1]{ \mathop{\rm Extr}_{ #1 } }

\newcommand{\mR}{\mathbb{R}}

\newcommand{\lb}{\left(}
\newcommand{\rb}{\right)}
\newcommand{\lbb}{\left\{}
\newcommand{\rbb}{\right\}}
\newcommand{\lsb}{ \left[ }
\newcommand{\rsb}{ \right] }

\newcommand{\T}[1]{\tilde{#1}}
\newcommand{\Req}[1]{eq.\ (\ref{eq:#1})} 

\newcommand{\NReq}[1]{(\ref{eq:#1})}
\newcommand{\Reqs}[2]{eqs.\ (\ref{eq:#1}) and (\ref{eq:#2})}   
\newcommand{\BReqs}[2]{Eqs.\ (\ref{eq:#1}) and (\ref{eq:#2})}   
\newcommand{\NReqs}[2]{(\ref{eq:#1},\ref{eq:#2})}
\newcommand{\Reqss}[2]{eqs.\ (\ref{eq:#1}-\ref{eq:#2})}

\newcommand{\Rfig}[1]{Fig.\ \ref{fig:#1}}
\newcommand{\Rfigs}[2]{Figs.\ \ref{fig:#1} and \ref{fig:#2}}

\newcommand{\Lfig}[1]{\label{fig:#1}}
\newcommand{\Leq}[1]{\label{eq:#1}}
\newcommand{\Rsec}[1]{sec.\ \ref{sec:#1}}
\newcommand{\BRsec}[1]{Sec.\ \ref{sec:#1}}
\newcommand{\Rsecs}[2]{secs.\ \ref{sec:#1}-\ref{sec:#2}}
\newcommand{\Lsec}[1]{\label{sec:#1}}
\newcommand{\be}{\begin{eqnarray}}
\newcommand{\ee}{\end{eqnarray}}
\newcommand{\ba}{\begin{array}}
\newcommand{\ea}{\end{array}}
\newcommand{\no}{\nonumber}

\newcommand{\subbe}{\begin{subequations}}
\newcommand{\subee}{\end{subequations}}

\newcommand{\mc}[1]{\mathcal{#1}}
\newcommand{\argmax}{\mathop{\rm arg\,max}\limits}
\newcommand{\argmin}{\mathop{\rm arg\,min}\limits}

\newcommand{\bs}{\backslash}
\newcommand{\E}{\mathbb{E}}

\newcommand{\Cov}{\mathrm{Cov}}

\newcommand{\remid}[1]{\mathrel{}\middle#1\mathrel{}}
\newcommand{\ul}[1]{\underline{#1}}
\newcommand{\PPc}{n}
\newcommand{\rPPc}{\ell}
\newcommand{\LLNap}{\approx}
\newcommand{\CLTap}{\approx}
\newcommand{\CLTapd}{\overset{\text{d}}{\approx}}
\newcommand{\RI}{\textrm{I}}
\newcommand{\RII}{\textrm{II}}
\newcommand{\RIII}{\textrm{III}}
\newcommand{\RIV}{\textrm{IV}}
\newcommand{\ind}{\mathbb{I}}
\newcommand{\erf}{\mathrm{erf}}
\newcommand{\erfi}{\mathrm{erfi}}
\newcommand{\rn}{p}
\newcommand{\tm}{m}

\title{Perfect reconstruction of sparse signals using nonconvexity control and one-step RSB message passing}
\author{Xiaosi Gu$^{1,2}$ \and Ayaka Sakata$^{2,3}$  \and  Tomoyuki Obuchi$^{1,2}$} 
\date{%
   $^{1}$Graduate School of Informatics, Kyoto University, Kyoto, Japan\\%
 $^{2}$RIKEN center for AIP, Tokyo, Japan \\%
 $^{3}$Department of Information Science, Ochanomizu University, Tokyo, Japan\\%
    \today
    }
    
\begin{document} 
\maketitle

\begin{abstract}
We consider sparse signal reconstruction via minimization of the smoothly clipped absolute deviation (SCAD) penalty, and develop one-step replica-symmetry-breaking (1RSB) extensions of approximate message passing (AMP), termed 1RSB-AMP. Starting from the 1RSB formulation of belief propagation, we derive explicit update rules of 1RSB-AMP together with the corresponding state evolution (1RSB-SE) equations. A detailed comparison shows that 1RSB-AMP and 1RSB-SE agree remarkably well at the macroscopic level, even in parameter regions where replica-symmetric (RS) AMP, termed RS-AMP, diverges and where the 1RSB description itself is not expected to be thermodynamically exact. Fixed-point analysis of 1RSB-SE reveals a phase diagram consisting of success, failure, and diverging phases, as in the RS case. However, the diverging-region boundary now depends on the Parisi parameter due to the 1RSB ansatz, and we propose a new criterion---minimizing the size of the diverging region---rather than the conventional zero-complexity condition, to determine its value. Combining this criterion with the nonconvexity-control (NCC) protocol proposed in a previous RS study improves the algorithmic limit of perfect reconstruction compared with RS-AMP. Numerical solutions of 1RSB-SE and experiments with 1RSB-AMP confirm that this improved limit is achieved in practice, though the gain is modest and remains slightly inferior to the Bayes-optimal threshold. We also report the behavior of thermodynamic quantities---overlaps, free entropy, complexity, and the non-self-averaging susceptibility---that characterize the 1RSB phase in this problem.
\end{abstract}


\section{Introduction}
Compressed sensing (CS) is a signal processing framework for reconstructing a (high-dimensional) signal vector from a set of linear measurements in the condition that the number of measurement is smaller than the signal dimension~\cite{donoho2006cs,candes2006robust}. This setting leads to an underdetermined inverse problem and there are infinitely many candidate signals consistent with the observations. To make the reconstruction feasible, CS exploits the assumption that the true signal is sparse: Only a small number of its components are nonzero. The central technical issue in CS thus becomes how to estimate the positions of these nonzero elements accurately. 

Let $\V{x}_0\in\mR^N$ be the sparse signal to be estimated, $\V{A}\in\mR^{M\times N}$ be the measurement matrix, and $\V{y}\in \mR^{M}$ be the measurement result that is generated as $\V{y}=\V{A}\V{x}_0$. A natural formulation reconstructing $\V{x}_0$ from given $\V{A}$ and $\V{y}$ is to find the sparsest solution satisfying the measurement relation $\V{y}=\V{A}\V{x}$, which leads to the so-called $\ell_0$ minimization problem:
\be
\min_{\V{x}} \|\V{x} \|_0\quad \mathrm{s.t.} \quad \V{y}=\V{A}\V{x},
\ee
where $\| \V{x} \|_0$ defines the $\ell_0$ norm giving the number of nonzero components of $\V{x}$. However, this formulation involves a combinatorial optimization problem, which becomes computationally intractable in high-dimensional settings. In fact, it is known to be NP-hard in the worst case~\cite{natarajan1995sparse}. A standard strategy to overcome this computational issue is the relaxation approach, which replaces the $\ell_0$ norm with a more tractable quantity. Among these alternatives, the substitution with the $\ell_p$ norm, $\| \V{x} \|_p=\lb \sum_{i=1}^{N}|x_i|^p \rb^{1/p}$, with $p\leq 1$ has been examined in a number of works~\cite{chartrand2007exact,chartrand2008ripnonconvex,7987040}. Especially, the $p=1$ case converts the problem into a convex optimization one, enabling the use of numerous efficient algorithms while keeping the obtained solution still sparse. The $\ell_1$ relaxation has achieved success across a wide range of applications, and its theoretical properties have also been well established. As a result, it has become the most commonly used and standard choice in CS.

Despite its success, the $\ell_{1}$ relaxation carries a well-known drawback: Compared with the $\ell_{p}$ case with $0\leq p<1$, it typically exhibits worse sample complexity, i.e., it requires a larger number of samples for perfect reconstruction~\cite{chartrand2007exact,chartrand2008ripnonconvex,7987040}. On the other hand, $\ell_p$ minimization at $p < 1$ is known to have some undesirable properties: Not only does it lead to a non-convex optimization problem, but its minimizer can also be discontinuous with respect to the input, resulting in instability of both the optimization algorithm and its output solution. Consequently, although $\ell_p$ minimization with $p < 1$ are superior in terms of sample complexity, it has not been considered practical. Some alternative approach that achieves both low sample complexity and stability of the solution is thus desired.

Promising candidates for this purpose are piecewise continuous nonconvex penalty functions, such as Smoothly Clipped Absolute Deviation (SCAD)~\cite{fan2001scad} and Minimax Concave Penalty (MCP)~\cite{zhang2010mcp}. SCAD and MCP are designed to simultaneously provide continuity, unbiasedness, and sparsity to the estimated signal. Their nonconvexities are controlled by two hyperparameters intrinsic to each penalty, allowing flexible adjustment of their influence on both the estimator and the optimization process, and they include the $\ell_1$ relaxation as a specific limiting case, meaning that these methods can surpass the $\ell_1$ relaxation under appropriately chosen hyperparameters. 

A precise quantitative evaluation of how much the sample complexity actually improves by these methods was clarified relatively recently by a part of the present authors~\cite{sakata2021scad}. In that work, assuming the measurement matrix $\V{A}$ to be a random matrix with entries identically independently distributed (i.i.d.) from zero-mean Gaussian, the authors derived the phase boundary separating the possible and impossible regions of reconstruction and also proposed an algorithm that aligns precisely with it, using the replica method and the approximate message passing (AMP), both at the replica symmetric (RS) level. The obtained phase boundary was clearly improved compared to that of the $\ell_1$ relaxation, and was quantitatively close to the phase boundary based on the Bayes-optimal (BO) method, both in the SCAD and MCP cases. This implies that SCAD and MCP can achieve nearly optimal performance in terms of sample complexity while keeping the computational cost sufficiently low.

Yet, some nontrivial observations, whose consequences were not understood well at that time, were made in the above work. One was that the phenomenology determining the reconstruction limit differs from the standard typology of phase transitions~\cite{Zdeborova02092016}: One of the typical failure scenarios of reconstruction is that some metastable state emerges and the basin of attraction (BoA) that contains the true solution becomes unreachable from reasonable initial conditions due to the free energy barrier originating from the metastable state; in contrast, in the SCAD/MCP minimization, no such metastable state appears but instead the AMP dynamics diverges in some parameter regions, which prevents the recovery of the true signal. Another nontrivial observation, which is related to the above, was that the replica symmetry breaking (RSB) always happens except in the phase where the perfect reconstruction is possible. Correspondingly, in~\cite{sakata2021scad}, the authors proposed a nonconvexity-control (NCC) protocol, which starts the algorithm with reasonable initial conditions which are in the failure phase exhibiting RSB and gradually varies the parameter controlling nonconvexity during the iterations. Resultantly, the AMP trajectory is navigated from the failure phase (RSB) to the success phase (RS) if the measurement rate $\alpha=M/N$ is large enough compared to the sparsity $\rho=\|\V{x}_0\|_0/N$; the boundary, at which this NCC fails to navigate the trajectory into the success phase due to the divergence region, defines the algorithmic limit of the SCAD/MCP minimization. Numerical experiments demonstrated that this strategy indeed works effectively and aligns well with the theory.

The above algorithmic limit was derived under the RS ansatz. While the RS analysis---more precisely, the state evolution (SE) analysis---accurately describes the macroscopic behavior of AMP, it does not necessarily capture the true performance of SCAD/MCP minimization itself. This is because RS is invalid within the parameter regions traversed during the NCC process. This suggests that incorporating RSB may improve the algorithmic limit, since the diverging region that defines the previous limit lies entirely within the RSB phase and is thus likely to be affected by RSB.

Accordingly, in this paper, we apply the one-step RSB (1RSB) version of AMP---also known as Approximate Survey Propagation (ASP)---to the SCAD minimization problem. We also introduce the corresponding macroscopic theory, namely the 1RSB version of SE, to analyze the macroscopic dynamical behavior of 1RSB version of AMP. For brevity, we hereafter denote AMP at the 1RSB level as 1RSB-AMP, while that at the RS level is denoted as RS-AMP if necessary to emphasize RS. The same notational convention applies to the corresponding SE and belief propagation (BP). 

The remainder of this paper is organized as follows. After reviewing earlier literatures about 1RSB message passing in the subsection immediately below, we summarize the formulation necessary to the present analysis in \Rsec{Formulation}. In \Rsec{Review}, the RS results obtained in~\cite{sakata2021scad} are briefly reviewed, to appropriately motivate our 1RSB analysis. \BRsec{Derivation} provides the explicit algorithms of 1RSB-AMP/SE, a derivation outline of 1RSB-BP being the basis of 1RSB-AMP/SE, and the results obtained from them; the detailed derivation of the algorithms are deferred to \Rsec{Message}. The last section is devoted to the conclusion.

\subsection{Related work on 1RSB message passing}\Lsec{Related}
Applications of the 1RSB message-passing method began with the analysis of random instances of constraint satisfaction problems (CSPs), including the well-known \emph{satisfiability (SAT) problem}, and the method was named \emph{Survey Propagation} (SP)~\cite{mezard-parisi-zecchina-science02,braunstein-mezard-zecchina-rsa05,mezard-montanari-book09}. SP propagates distributions (``surveys'') of BP messages, rather than single messages, enabling effective search even when the solution space shatters into exponentially many isolated clusters and standard BP from uninformed initializations ceases to converge. In practice, combining SP with decimation pushes the algorithmic threshold well past the standard BP threshold, and adding controlled backtracking further extends the solvable regime: In the case of random SAT with three variables in each clause, it is shown to essentially reach the satisfiability threshold~\cite{marino-parisi-riccitersenghi-natcomm16}. 

More recently, these ideas have been revisited in inference problems represented by densely connected graphs~\cite{antenucci2019asp,barbier2023l0asp,Barbier2025JSTAT}. In~\cite{antenucci2019asp}, the authors introduced the term \emph{ASP} and applied it to the $Z_2$ synchronisation problem which is equivalent to the ferromagnetically biased Sherrington--Kirkpatrick model in spin glasses, to find that ASP stabilises convergence even in the regime where AMP fails, and that the choice of the Parisi parameter has an impact on the algorithm performance; based on some empirical observations it has been suggested that tuning the Parisi parameter so as to satisfy the zero-complexity condition is beneficial. Building on this, Lucibello et al. proposed \emph{generalised ASP} (GASP) for a fairly wide class of estimation problems based on generalised linear models~\cite{saglietti2019gasp}; applied to phase retrieval, it was shown that the algorithmic threshold achieved by GASP is far better than that of generalized AMP (GAMP) and is comparable with the BO one. Besides, in~\cite{barbier2023l0asp} ASP was applied to $\ell_0$-regularised CS, to find that its perfect reconstruction limit is much improved from that of $\ell_1$ minimization and even close to the BO one. Its extended version was provided in~\cite{Barbier2025JSTAT}, deriving a complete phase diagram with detailed information; the choice of the Parisi parameter was also examined and the zero-complexity condition was shown to give the ideal reconstruction limit; in the zero-regularization limit under the zero-complexity condition, a simplified ASP with identical reconstruction limit, from which the Parisi parameter dependence is removed, was also proposed and demonstrated.

\section{Formulation}\Lsec{Formulation}

\subsection{Problem setting}\Lsec{Problem setting}
As above, the true signal, measurement matrix, and measurement result are denoted as $\V{x}_0\in\mR^N, \V{A}\in\mR^{M\times N},$ and $\V{y}=\V{A}\V{x}_0\in \mR^{M}$, respectively. Given $(\V{A}, \V{y})$, we aim to obtain the estimator for $\V{x}_0$ by solving the following minimization problem:
\be
\V{x}^{\star}
= \argmin_{\V{x} \in \mathbb{R}^{N}} J(\V{x};a,\lambda) \quad \mathrm{s.t.} \quad \V{y} = \V{A} \V{x},
  \Leq{problem}
\ee
where $J(\V{x};a,\lambda)=\sum_{i=1}^{N} J(x_i;a,\lambda)$ denotes the SCAD penalty whose explicit form is
\be
J(x; a,\lambda) =
\left\{
\begin{array}{cc}
\lambda |x|,     & |x|\leq \lambda   \\
-\frac{x^2-2a\lambda|x|+\lambda^2 }{2(a-1)},     & \lambda< |x| \leq a\lambda  \\
\frac{1}{2}\lambda^2(a+1),     & a\lambda< |x|  
\end{array}
\right..
\Leq{SCAD}
\ee
The parameters $a,\lambda$ are hereafter called nonconvexity parameters. 
The shape of the SCAD penalty can be understood as an interpolation between the $\ell_1$ and $\ell_0$ penalties. For $|x|\leq \lambda$, it behaves as the $\ell_1$ penalty and promotes sparsity around the origin.
For $|x|>a\lambda$, it becomes constant, as in the $\ell_0$ penalty, and hence large components are not further penalized, avoiding shrinkage bias; the middle concave part smoothly connects these two regimes.
In the limit $a\to \infty$, the standard $\ell_1$ norm with the coefficient $\lambda$ is reproduced. 

The objective of this study is to propose an efficient algorithm using AMP combined with 1RSB, 1RSB-AMP, for solving \Req{problem}. In addition, we aim to elucidate the asymptotic properties of both the algorithm and the estimator in the limit $N \to \infty$. For appropriately treating this limit, we adopt a random generative process of the signal that is commonly used for this class of problems. Specifically, each component of the signal vector is assumed to be i.i.d. and drawn from the following Bernoulli-Gaussian distribution:
\be
&&
  P_{0}(x)\coloneqq   \rho \,\mathcal{N}(x\mid 0, \sigma_{x}^{2})
  + (1 - \rho )\,\delta(x),
\Leq{prior}
\ee
where $\mathcal{N}(\mu, \sigma^{2})$ denotes the Gaussian distribution with mean $\mu$ and variance $\sigma^2$ and $\mathcal{N}(x \mid \mu, \sigma^{2})=e^{-(x-\mu)^2/(2\sigma^2)}/\sqrt{2\pi \sigma^2}$ represents the corresponding probability density function for the variable $x$. Meanwhile, each component of the measurement matrix, $A_{\mu i}$, is also assumed to be i.i.d. from the zero-mean Gaussian with variance $1/N$: $A_{\mu i}\sim \mc{N}(0,1/N)$. Besides, the number of measurements $M$ is assumed to obey the following scaling:
\be
M=\alpha N,~\alpha=O(1).
\ee

\subsection{One-dimensional estimator and related quantities}\Lsec{One-dimensional estimator}
To clarify the basic properties of the estimator, we consider the following one-dimensional optimization problem:
\be
&&
g(x,\Gamma,h,a,\lambda)\coloneqq -\frac{1}{2}\Gamma x^2+hx -J(x;a,\lambda),
\\ &&
\hat{x}\lb 
\Gamma,h,a,\lambda\rb
\coloneqq
\argmax_{x}\lb g(x,\Gamma,h,a,\lambda) \rb
\no \\ &&
=
\left\{
\begin{array}{cc}
\frac{h}{\Gamma},  & |h| > a\lambda \Gamma  \\
\frac{1}{\Gamma-(a-1)^{-1}}\lb  h-\frac{a\lambda}{a-1}\sgn{h}\rb ,  & \lambda(1+\Gamma) < |h| \leq  a\lambda \Gamma \\
\frac{1}{\Gamma}\lb h-\lambda \sgn{h}\rb,  & \lambda < |h| \leq  \lambda(1+\Gamma) \\
0,  & |h| \leq \lambda     
\end{array}
\right..
\ee
As will be clear later, this $\hat{x}$ serves as an effective estimator directly related to \Req{problem}. This is a piecewise linear function with respect to (w.r.t.) $h$, and for brevity we express each analytic interval in $h$ as
\subbe
\be
&&
\RI(\Gamma,a,\lambda)   \coloneqq \{h \mid |h| > a\lambda \Gamma\},
\\&&
\RII(\Gamma,a,\lambda)  \coloneqq \{h \mid   \lambda(1+\Gamma) < |h| \leq  a\lambda \Gamma \},
\\&&
\RIII(\Gamma,a,\lambda) \coloneqq \{h \mid \lambda < |h| \leq  \lambda(1+\Gamma)\},
\\&&
\RIV(\Gamma,a,\lambda) \coloneqq \{h \mid  |h| \leq \lambda\}.
\ee
\subee
The arguments may be omitted for brevity if it does not cause confusion. In particular, $a$ and $\lambda$ are not important during the most of analytic computations and they are basically omitted in \Rsecs{Review}{Message} for any functions. Substituting the optimal solution $\hat{x}$, $g$ can be rewritten as follows:
\be
&&
\hat{g}(\Gamma,h,a,\lambda)\coloneqq g(\hat{x}(\Gamma,h,a,\lambda),\Gamma,h,a,\lambda)
\no \\ &&=
\left\{
\begin{array}{cc}
\frac{1}{2}\frac{h^2}{\Gamma}-\frac{1}{2}\lambda^2(a+1)
=\frac{1}{2}\Gamma \hat{x}^2-\frac{1}{2}\lambda^2(a+1)
,  & \RI  \\
\frac{1}{2}\frac{1}{\Gamma-(a-1)^{-1}}\lb h-\frac{a\lambda}{a-1} \sgn{h}\rb^2+\frac{1}{2}\frac{\lambda^2}{a-1}
=
\frac{1}{2}\lb \Gamma-(a-1)^{-1} \rb \hat{x}^2+\frac{1}{2}\frac{\lambda^2}{a-1} 
 ,  &  \RII \\
\frac{1}{2}\frac{(h-\lambda \sgn{h})^2}{\Gamma}
=\frac{1}{2}\Gamma \hat{x}^2
,  & \RIII \\
0,  & \RIV
\end{array}
\right..
\Leq{ghat-cbc}
\ee
We also introduce (rescaled) \emph{susceptibility} which is defined as the differentiation of $\hat{x}$ w.r.t. $h$:
\be
&&
\chi\lb  \Gamma,h,a,\lambda\rb
=\partial_h \hat{x}
\lb  \Gamma,h,a,\lambda\rb
=
\left\{
\begin{array}{cc}
\frac{1}{\Gamma},  & \RI  \\
\frac{1}{\Gamma-(a-1)^{-1}} ,  & \RII \\
\frac{1}{\Gamma},  &  \RIII \\
0,  &  \RIV  
\end{array}
\right..
\ee
For illustration, the behaviors of $\hat{x}$ and $\chi$ against $h$ at specific parameter values are plotted in \Rfig{one-dim estimator}.
\begin{figure}[htbp]
\begin{center}
\vspace{0mm}
\includegraphics[width=0.45\columnwidth]{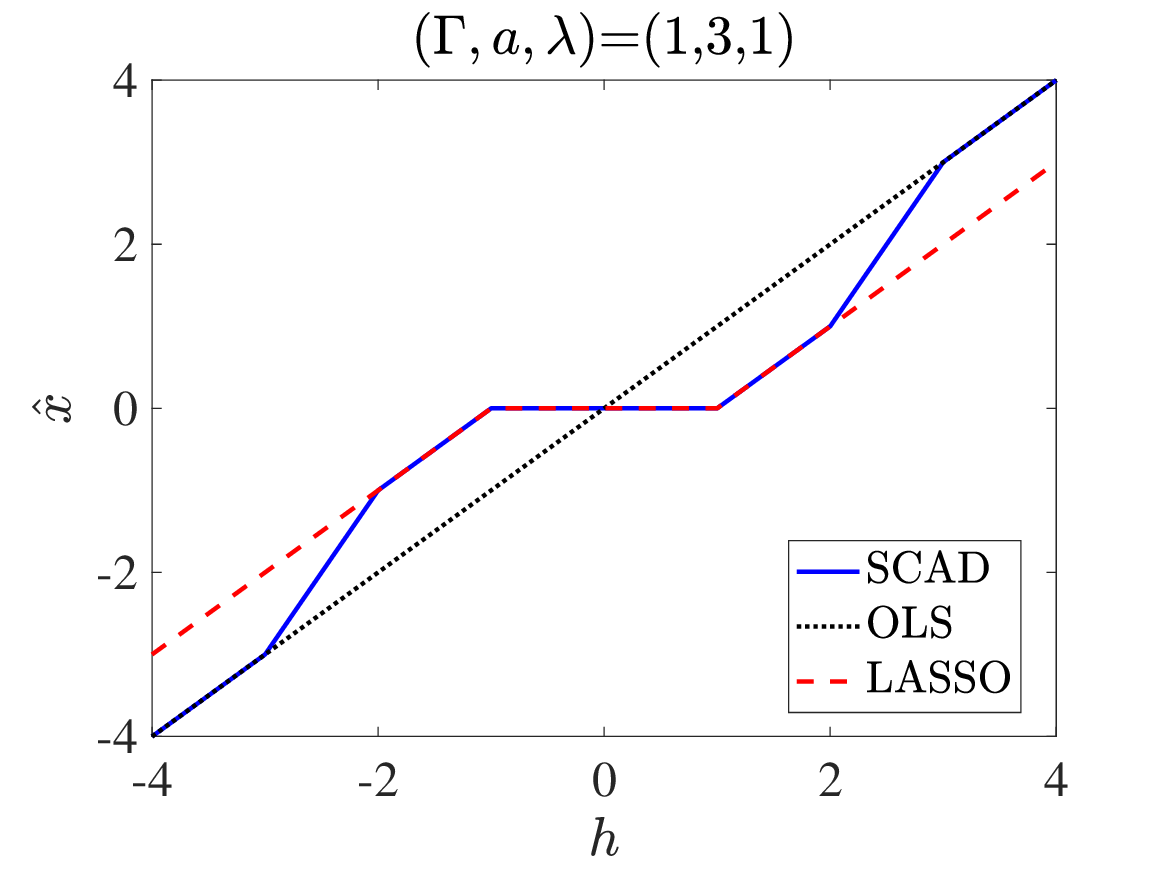}
\includegraphics[width=0.45\columnwidth]{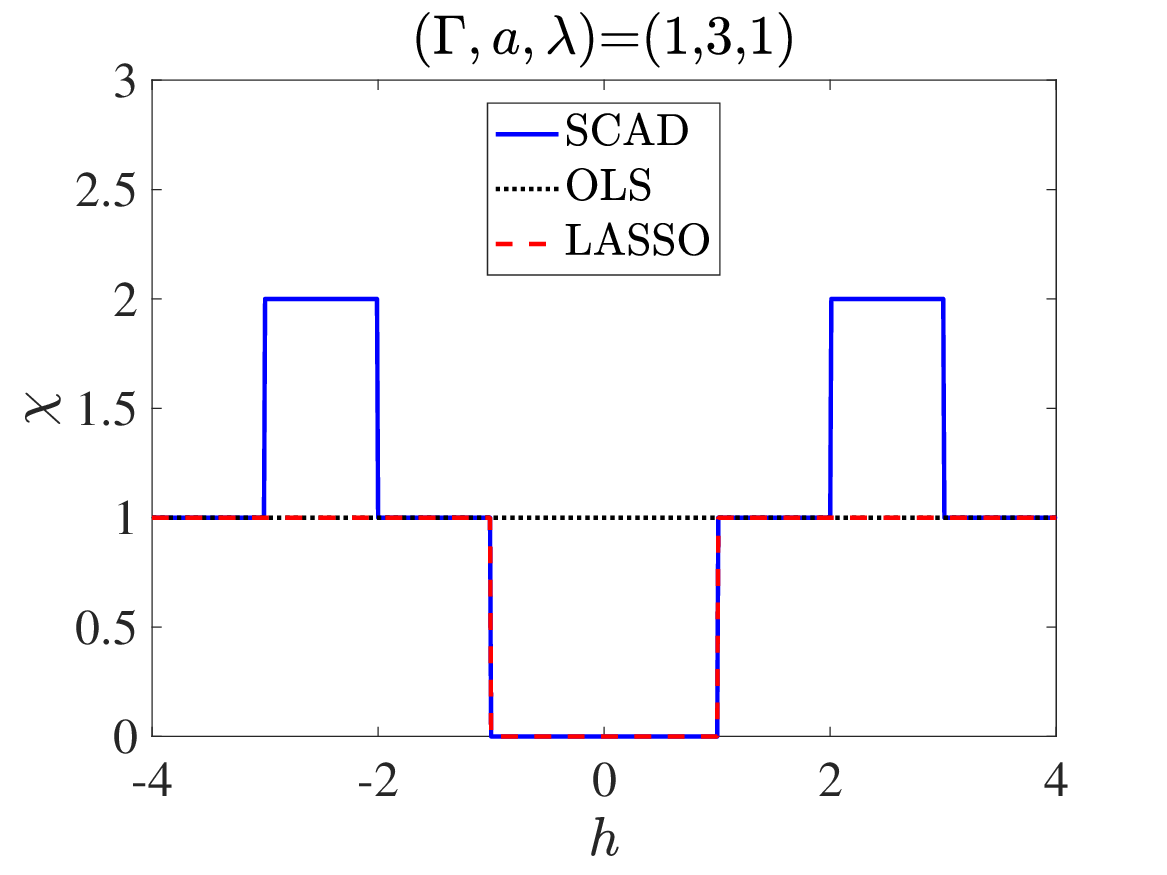}
\vspace{0mm}
\caption{The blue solid lines represent $\hat{x}$ (left) and $\chi$ (right) plotted against $h$ at $(\Gamma,a,\lambda)=(1,3,1)$. The black dotted lines are for the case of ordinary least square (OLS), and the red broken ones are for the $\ell_1$ minimization or LASSO. These cases are derived as specific limits of the SCAD estimator: OLS and LASSO correspond to the $\lambda=0$ and $a\to \infty$ cases, respectively.
}   
\Lfig{one-dim estimator}
\end{center}
\end{figure}
These one-dimensional estimator and susceptibility play important roles in later computations.

\subsection{Statistical-mechanical formulation}\Lsec{Statistical-Mechanical Formulation}
The estimator given in \Req{problem} can be analyzed within a statistical-mechanical framework by introducing an equivalent Boltzmann distribution. Specifically, we define the following Boltzmann distribution:
\be
&&
  P_{\beta, \tau}(\V{x} \mid \V{y},\V{A})
  = \frac{1}{Z_{\beta, \tau}(\V{y},\V{A})}
  \frac{1}{\sqrt{2\pi \tau}^{M}}\exp\lbb
    -\frac{1}{2\tau}\|\V{y} - \V{A} \V{x}\|_{2}^{2}
    - \beta\, J(\V{x}; a,\lambda)
\rbb,
\Leq{boltzmann_distribution}
\\ &&
 Z_{\beta, \tau}(\V{y},\V{A})
  = \int d \V{x}\;
  \frac{1}{\sqrt{2\pi \tau}^{M}}\exp\lbb
    -\frac{1}{2\tau}\|\V{y} - \V{A} \V{x}\|_{2}^{2}
    - \beta\, J(\V{x}; a,\lambda)
\rbb,
 \Leq{partition_function}
\ee
where $\beta$ denotes the inverse temperature and $\tau$ controls the noise level of the measurement constraint. The normalization constant $Z_{\beta,\tau}$ is called partition function according to the physics terminology. The Boltzmann distribution concentrates on the global minimizer of the SCAD objective in the following limit:
\begin{equation}
  \lim_{\beta \to \infty}\,\lim_{\tau \to 0}   P_{\beta, \tau}(\V{x}\mid \V{y},\V{A} )
  = \delta(\V{x} - \V{x}^{\star}).
\Leq{delta_limit}
\end{equation}
Strictly speaking, if the global minimizer is not unique, $\V{x}^{\star}$ in \Req{delta_limit} should be understood as the set of global minimizers, and the right-hand side should be interpreted as the equally weighted sum of delta functions supported on that set.
In the present random setting, however, we expect the global minimizer to be unique with high probability.

Thus, $\V{x}^{\star}$ in \Req{problem} can be computed from the average over the Boltzmann distribution in the limit. The order of the limits ensures that the hard measurement constraint $\V{y} = \V{A} \V{x}$ is imposed. 
 
Based on this probabilistic formulation, we construct the message passing algorithms below.

\subsubsection{Factors and messages}\Lsec{Factors and Messages}
For the message passing algorithms, it is convenient to express the Boltzmann distribution~\NReq{boltzmann_distribution} in a factorized form as 
\be
  P_{\beta, \tau}(\V{x}\mid \V{y},\V{A} )
  = \frac{1}{Z_{\beta, \tau}(\V{y},\V{A})}
  \Biggl( \prod_{\mu = 1}^{M} \Phi_{\mu}(\V{x}) \Biggr)
  \Biggl( \prod_{i = 1}^{N} \varphi(x_{i}) \Biggr),
  \Leq{factorized_form}
\ee
where 
\begin{align}
  \Phi_{\mu}(\V{x})
  &= \frac{1}{\sqrt{2\pi\tau}}
     \exp\lbb 
       -\frac{1}{2\tau}
       \lb y_{\mu} - \sum_{j=1}^{N} A_{\mu j} x_{j}\rb^{2}
     \rbb,
     \Leq{measurement_factor}\\
  \varphi(x_{i})
  &= \exp\lbb -\beta\, J(x_{i}; \lambda, a)\rbb.
     \Leq{prior_factor}
\end{align}
In graphical modeling using bipartite factor graphs, the factors $\Phi_{\mu}(\V{x})$ and $\varphi(x_i)$ are associated with function node and variable node, respectively, and BP provides a recursive scheme for (approximately) computing the marginal distribution of each variable node from the local interactions of these factors based on that modeling. Concretely, the BP messages are determined by the following recursive relation:
\subbe
\Leq{BP}
\be
&&
\tilde{\phi}_{\mu \to i}(x_i)
\propto
\int \! d\V{x}^{\bs i}\,
\Phi_\mu(\V{x})
\prod_{j (\neq i)} \phi_{j \to \mu}(x_j), 
\Leq{BP1}\\
&&
\phi_{i \to \mu}(x_i)
\propto
\varphi(x_i)
\prod_{\nu (\neq \mu)} \tilde{\phi}_{\nu \to i}(x_i)
\Leq{BP2},
\ee
\subee
where $d\V{x}^{\bs i} = \prod_{j(\neq i)}dx_j$. The message $\tilde{\phi}_{\mu \to i}(x_i)$ expresses the ``belief’’ sent from the factor node $\mu$ to the variable node $i$, representing the effective likelihood of $x_i$ given the measurement $y_\mu$. Conversely, $\phi_{i \to \mu}(x_i)$ is the message from variable node $i$ to factor node $\mu$, encapsulating the prior knowledge and influences from all other connected factors except $\mu$ and describing the marginal distribution on $x_i$ when the $\mu$th function node is deleted from the original graph. The true marginal distribution for each variable node is obtained using the messages as
\be
\phi_i(x_i) \propto \varphi(x_i) \prod_{\mu=1}^M \tilde{\phi}_{\mu \to i}(x_i).
\Leq{full marginal}
\ee
The algorithms at the 1RSB level are also derived based on these basic BP iterations.

\section{Review of the RS result}\Lsec{Review}
In this section, we briefly review the RS result obtained in the previous work~\cite{sakata2021scad}, in order to clarify the position of the present study and the need for RSB analysis.

The RS-AMP is given by
 \subbe
 \Leq{AMP_RS}
 \be
 &&
 \chi^{(t)}=\frac{1}{M}\sum_{\mu=1}^{M}\sum_{j}A_{\mu j}^2\chi_{j}^{(t)},
 \Leq{chi_AMP_RS_avg}
  \\ &&
 r_{\mu}^{(t)}=\frac{y_{\mu}-\sum_{j}A_{\mu j}\hat{x}_{j}^{(t)}}{\chi^{(t)}}+r_{\mu}^{(t-1)},
\Leq{r_AMP_RS}
  \\ &&
\Gamma^{(t+1)}=\frac{\alpha}{\chi^{(t)}},
  \\ &&
 h_{i}^{(t+1)}
 =
  \sum_{\mu } A_{\mu i}r_{\mu}^{(t)}+
  \frac{\alpha}{\chi^{(t)}}\hat{x}_{i}^{(t)},
 \\ &&
 \hat{x}_{i}^{(t+1)}=\hat{x}(\Gamma^{(t+1)},h_{i}^{(t+1)}),
\Leq{xhat_AMP_RS}
 \\ &&
 \chi_{i}^{(t+1)}=\partial_{h}\hat{x}(\Gamma^{(t+1)},h_{i}^{(t+1)}).
\Leq{chi_AMP_RS}
\ee
\subee
Here, $\hat{x}^{(t+1)}_i$ in \Req{xhat_AMP_RS} is the estimate of $x_{0i}$ at the $(t+1)$th AMP update, and on the righthand side (rhs) it is computed using the one-dimensional estimator introduced in \Rsec{One-dimensional estimator}; the susceptibility $\chi_i$ in \Req{chi_AMP_RS} is computed in a similar way; for brevity, the arguments $a$ and $\lambda$ are omitted. Besides, $\chi$ in the lefthand side (lhs) of \Req{chi_AMP_RS_avg} gives the averaged susceptibility of each variable's susceptibility $\chi_i$, and $r_{\mu}$ in \Req{r_AMP_RS} is defined as the ``cavity'' residual $(y_{\mu}-\sum_{i}A_{\mu i}\hat{x}_{i\to \mu})$ weighted by $1/\chi$ and that with a one-step time shift in the rhs corresponds to the so-called Onsager reaction term which arises when rewriting the ``cavity'' estimator $\hat{x}_{i\to \mu}$ with the full estimator $\hat{x}_i$. The macroscopic behavior of AMP is characterized by the following SE equations with two scalar variables:
\subbe
\Leq{SE_RS}
\be
&&
\chi^{(t+1)}
=\int dx_0 P_0(x_0) \int Dz
 \partial_h \hat{x}
\lb \frac{\alpha}{\chi^{(t)}},\frac{\alpha}{\chi^{(t)}}x_{0}+\frac{1}{ \chi^{(t)} }\sqrt{\alpha\epsilon^{(t)}}z
\rb,
\\ &&
\epsilon^{(t+1)}
=\int dx_0 P_0(x_0) \int Dz
\lb
x_0- \hat{x}\lb \frac{\alpha}{\chi^{(t)}},\frac{\alpha}{\chi^{(t)}}x_{0}+\frac{1}{ \chi^{(t)} }\sqrt{\alpha\epsilon^{(t)}}z \rb
\rb^2,
\ee
\subee
where $\chi$ is again the averaged susceptibility and $\epsilon$ is the mean square error (MSE) defined by
\be
\epsilon^{(t)}
=\frac{1}{N}\left\| \V{x}_0-\hat{\V{x}}^{(t)}\right\|_2^2
=\frac{1}{N}\sum_i^{N} \lb x_{0i}-\hat{x}_i^{(t)} \rb^2,
\ee
and 
\be
\int Dz(\cdots)=\int \frac{dz}{\sqrt{2\pi}}e^{-\frac{1}{2}z^2}(\cdots).
\ee
The fixed points $(\chi^*,\epsilon^*)$ of the SE equations characterize the macroscopic stationary state of the system. The stability condition of the fixed points against the RSB perturbation, which is usually called Almeida-Thouless (AT) condition and is empirically known to be equivalent of the dynamical stability of BP/AMP~\cite{doi:10.1143/JPSJ.72.1645,Kabashima_2003}, is given by
\be
\frac{\alpha}{(\chi^*)^2}
\int dx_0 P_0(x_0) \int Dz 
\lbb 
 \partial_h \hat{x}
\lb \frac{\alpha}{\chi^*},\frac{\alpha}{\chi^*}x_{0}+\frac{1}{\chi^*}\sqrt{\alpha\epsilon^*}z
\rb
\rbb^2 \leq 1.
\Leq{AT_RS}
\ee
These constitute the central tools of the analysis.

Based on \Reqs{SE_RS}{AT_RS}, phase transitions and the achievable limit of perfect reconstruction can be analyzed: The resultant phase diagrams are displayed in \Rfig{PD_RS1}.
\begin{figure}[htbp]
\begin{center}
\vspace{0mm}
\includegraphics[width=0.45\columnwidth]{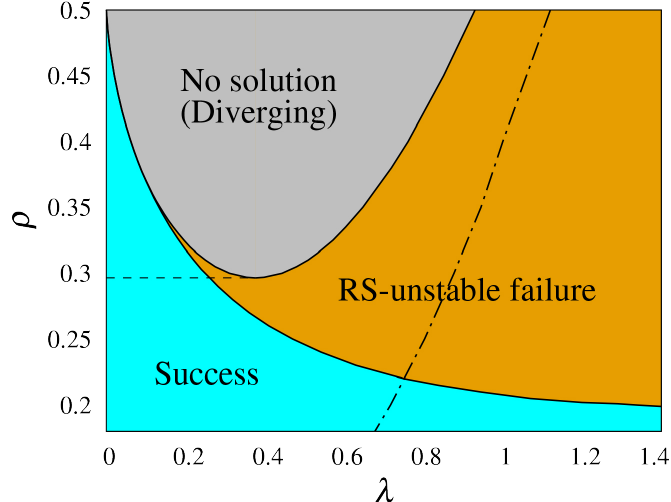}
\includegraphics[width=0.45\columnwidth]{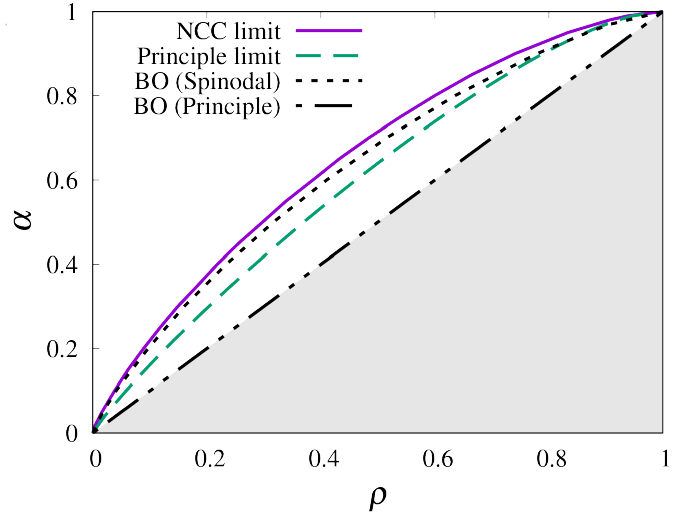}
\vspace{0mm}
\caption{
(Left) Phase diagram on the $\lambda$--$\rho$ plane at $a=3,\alpha=0.5,\sigma_x^2=1$. The horizontal dashed line indicates the limit achievable by NCC, while the dotted–dashed line marks the boundary of the region where SE flow divergence occurs: On the left side of this line, a diverging state appears in the $\chi$--$\epsilon$ plane, whereas it does not exist on the right side.
\\
(Right) The reconstruction limit on $\rho$--$\alpha$ plane by NCC (purple solid line) and the principle limit at $(a,\lambda)=(3,0.1)$ (green dashed line) that describes the boundary below which the perfect reconstruction solution becomes unstable. For comparison, the algorithmic (BO, spinodal) and principle (BO, principle) limits of the BO method are also shown.
\\  
Both panels are reproductions of the corresponding figures from~\cite{sakata2021scad}. 
}   
\Lfig{PD_RS1}
\end{center}
\end{figure}
The left panel describes the phase diagram on the $\lambda$--$\rho$ plane at $a=3,\alpha=0.5$. We can see that there are three distinct phases:
\begin{itemize}
\item{RS-unstable failure phase: This region has only one fixed point involving RSB. The corresponding MSE $\epsilon^*$ is greater than zero, indicating a failure in reconstruction.}
\item{Success phase: This phase also has a single fixed point, but RS remains unbroken. The corresponding MSE $\epsilon^*$ (and also $\chi^*$) becomes zero, indicating perfect reconstruction.}
\item{No solution (Diverging) phase: In this region, no fixed point exists, and the SE flow diverges.}
\end{itemize}
This phase structure seemingly implies that we can choose a small value of $\lambda$ for any $\rho$ to achieve the perfect reconstruction, but unfortunately this does not work. This is because, although the perfect reconstruction solution indeed exists as a fixed point in the small $\lambda$ region, its BoA is extremely small: Unless the initial condition has an exceptionally small MSE, the dynamics cannot reach the perfect reconstruction fixed point; when starting from reasonable initial conditions, the SE flow and the AMP trajectory diverge. Such diverging states always exist and occupy a part of the $\epsilon$--$\chi$ plane on the left of the dotted–dashed line. For this reason, in~\cite{sakata2021scad} a strategy called NCC was adopted. In that strategy, the AMP algorithm starts from a reasonable initial condition and the nonconvexity parameters are gradually varied, mainly decreasing $\lambda$ which is initialized at a large value. Through this approach, it was revealed that for given $\alpha$, the limiting value of $\rho$ for which the perfect reconstruction solution is reachable corresponds to the borderline that just avoids entering the No solution phase---that is, the horizontal dashed line on this phase diagram. This limit is referred to as the NCC limit. 

The NCC limit for the whole $\alpha$ region is plotted in the right panel of \Rfig{PD_RS1}. 
Here, the principle limit means the local-stability limit of the perfect-reconstruction fixed point, following the terminology of Ref.~\cite{sakata2021scad}.
This limit is quite different from the NCC limit. This implies that the algorithmic limit under reasonable conditions, corresponding to the NCC limit in this case, is much different from the globally optimal solution (principle limit) of the SCAD minimization. For comparison, the algorithmic and principle limits of the Bayes Optimal (BO) method are also plotted. 

This NCC limit is indeed realizable by AMP. To demonstrate this, \Rfig{nonconvexity_control} shows the trajectories of $\chi$ and $\epsilon$ computed from AMP and SE under NCC.
\begin{figure}[htbp]
\begin{center}
\vspace{0mm}
\includegraphics[width=0.45\columnwidth]{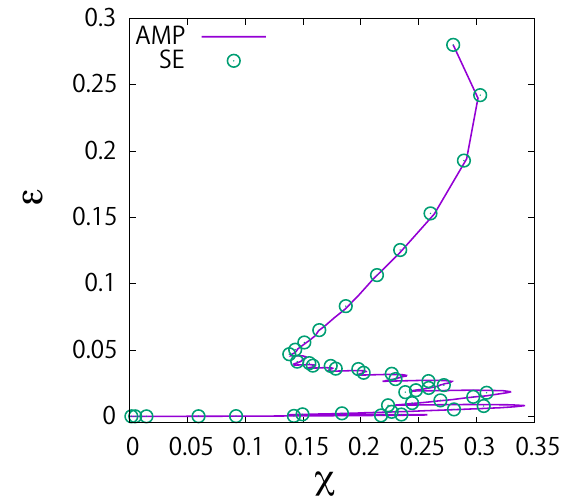}
\vspace{0mm}
\caption{
The AMP trajectory and the SE snapshots on $\chi$--$\epsilon$ plane during NCC. This trajectory is for one realization of $\V{x}_0$ and $\V{A}$, and the parameters are $(N,a,\sigma_x^2,\alpha,\rho)=(10^5,3,1,0.5,0.28)$. The initial condition is set to be $\V{x}^{(0)}=\V{0},\chi_i^{(0)}=\rho \sigma_x^2$, implying that the corresponding initial condition of SE is $\chi^{(0)}=\epsilon^{(0)}=\rho \sigma_x^2$. The parameter $\lambda$ is initialized at $\lambda=1$ and gradually decreased. Adapted from~\cite{sakata2021scad} with minor modifications.
}   
\Lfig{nonconvexity_control}
\end{center}
\end{figure}
From this figure, it is evident that the AMP trajectory behaves consistently with the SE evolution and ultimately converges to zero MSE. The examined parameters are $(N,a,\sigma_x^2,\alpha,\rho)=(10^5,3,1,0.5,0.28)$. Since the NCC limit at $\alpha=0.5$ is $\rho_{\rm NCC}\approx0.297$, this setting is quite close to the limit, yet AMP successfully achieves perfect reconstruction. This clearly demonstrates that the NCC protocol indeed works and that its behavior precisely corresponds to the prediction of SE. Incidentally, the NCC protocol used here is an ``equilibrium'' one: We monitor the susceptibility $\chi^{(t)}$ and $D^{(t)} = \|\hat{\V{x}}^{(t+1)} - \hat{\V{x}}^{(t)}\|/N$, and once these quantities begin to exhibit stationary behavior, we slightly decrease the value of $\lambda$; then, for that new $\lambda$, we again wait until these quantities become stationary, and once they do, we decrease $\lambda$ a little further. Here, the decrement of $\lambda$ was set to $\Delta \lambda = 0.1$, and \cite{sakata2021scad} discussed that this choice is sufficiently safe from the perspective of SE.

The above behavior was derived under the RS assumption. However, since the AMP/SE trajectories pass through the RS-unstable failure phase, and because the diverging phase that determines the algorithmic limit lies entirely within the RSB phase, the RS results may fail to accurately capture the true thermodynamic (thus global) properties of the system in that region. This suggests that taking RSB into account could improve the NCC limit, motivating the following analysis and algorithm construction at the 1RSB level. 

It should be noted that the previous results in~\cite{sakata2021scad} are correct in the sense that RS-SE accurately describes, at the macroscopic level, the dynamics of the RS-AMP algorithm. However, since RSB occurs in the relevant parameter region, this does not necessarily imply that the RS description captures the true global structure of the SCAD minimization problem. Motivated by this observation, in the present work we consider AMP and SE incorporating RSB effects, aiming at a description that more faithfully reflects the global structure of the problem.

\section{Derivation outline and results at the 1RSB level}\Lsec{Derivation}
In this section, we present the explicit forms of AMP and SE at the 1RSB level together with an outline of the corresponding BP derivation, and then describe the results obtained from them.

\subsection{Derivation outline of 1RSB-BP}\Lsec{Derivation outline}
Computations at the 1RSB level are usually involved. However, beyond such technical aspects, it is interesting to see how 1RSB-BP differs from RS-BP conceptually. Therefore, in this subsection, we focus on this conceptual difference and provide an outline of the derivation of 1RSB-BP.

In RSB phases, it is known that many metastable states, which are exponentially many w.r.t. $N$, emerge. These metastable states possess certain properties and are called \emph{pure states}~\cite{MezardParisiVirasoro1987}. In the context of BP, each pure state is considered to be corresponding to
the fixed point solution of \Req{BP}~\cite{mezard-montanari-book09}. RSB-BP/AMP deals with these solutions as an ensemble. 

Two methodological approaches are currently known for handling this ensemble. The one is a method that directly treats the distribution of the BP messages. In this approach, one considers the Bethe free energy, a functional of the messages, as an energy and constructs a Boltzmann distribution by multiplying the Boltzmann factor defined from this energy with factors imposing the constraint that the messages satisfy \Req{BP}~\cite{mezard-montanari-book09}. The other resorts to the replica method: Multiple replicas are introduced and the Boltzmann distribution is simply extended to these replicas (by taking a product of the Boltzmann distribution over each replica), and BP is applied to that Boltzmann distribution for the replicated system. Depending on the context, this usage of replicas is referred to as the method of real replicas, the cloning method, or the replica extension method~\cite{mezard1999352,PhysRevLett.75.2847,doi:10.1143/JPSJ.74.2133,yasuda2012replica,parisi_urbani_zamponi_2020}. Although both approaches yield the same result, the computations are simpler in the replica-based method. Therefore, a derivation based on the replica approach is shown below. Detailed computations are deferred to~\Rsec{Message}.

Let us consider an $\PPc$-replica system $\ul{\V{x}}=(\V{x}_1,\V{x}_2,\ldots,\V{x}_{\PPc})$, where the underline $\ul{\cdot}$ hereafter represents the replicated system and the related quantities: For instance, $\ul{x}_{i}=(x_{1i},x_{2i},\ldots,x_{\PPc i})$. The corresponding Boltzmann distribution is 
\be
\ul{P}_{\beta, \tau}(\ul{\V{x}}\mid \V{y},\V{A})=\prod_{a=1}^{\PPc}P_{\beta, \tau}(\V{x}_a \mid \V{y},\V{A}),
\ee
and the associated factor representation is
\be
\ul{P}_{\beta, \tau}(\ul{\V{x}}\mid \V{y},\V{A})
=\frac{1}{\ul{Z}_{\beta,\tau}}\prod_{\mu=1}^{M}\ul{\Phi}_{\mu}(\ul{\V{x}})
\prod_{i=1}^{N}
\ul{\varphi}(\ul{x}_i),
\Leq{replicated Boltzmann-factor}
\ee
where
\be
&&
\ul{\Phi}_{\mu}(\ul{\V{x}})=\prod_{a=1}^{\PPc}\Phi_{\mu}(\V{x}_a),
\\ &&
\ul{\varphi}(\ul{x}_i)=\prod_{a=1}^{\PPc}\varphi(x_{ai}).
\ee
Here, $\PPc$ is introduced as the number of real replicas in the 1RSB construction, and is not the replica-trick index to be sent to zero. It corresponds to the finite-temperature Parisi parameter in the present formulation; its zero-temperature counterpart will be later introduced as $\rPPc=\lim_{\beta\to\infty}\beta\PPc$.
Since \Req{replicated Boltzmann-factor} has the same structure as \Req{factorized_form}, BP is applied in the same manner. This yields 
\subbe
\Leq{BP-1RSB}
\be
&&
\ul{\T{\phi}}_{\mu \to i}(\ul{x}_i) \propto \int d\ul{\V{x}}^{\bs i} \ul{\Phi}_{\mu}(\ul{\V{x}})\prod_{j (\neq i)}\ul{\phi}_{j \to \mu}(\ul{x}_j),
\Leq{BP1-replicated}
\\ &&
\ul{\phi}_{i \to \mu}(\ul{x}_i)\propto \ul{\varphi}(\ul{x}_i)\prod_{\nu (\neq \mu)}\ul{\T{\phi}}_{\nu \to i}(\ul{x}_i),
\Leq{BP2-replicated}
\ee
\subee
where $d\ul{\V{x}}^{\bs i}=\prod_{a=1}^{\PPc}\prod_{j(\neq i)}dx_{aj}$, and $\ul{\phi},\ul{\T{\phi}}$ are the messages for the replicated system. It has the trivial factorized solution
\be
\ul{\phi}_{i\to \mu}(\ul{x}_i)=\prod_{a=1}^{\PPc}\phi_{i \to \mu}(x_{ai}),\quad
\ul{\T{\phi}}_{\mu \to i}(\ul{x}_i)=\prod_{a=1}^{\PPc}\T{\phi}_{\mu \to i}(x_{ai}),
\ee
which is just a bunch of the BP messages for the non-replicated system and corresponds to the RS solution. To take into account the 1RSB effect, we need to include appropriate fluctuations and correlations between replicas while keeping the permutation symmetry among the replicas. According to de Finetti's theorem, any probability distributions having that symmetry, thus the messages, are expressed in the following form:
\subbe
\Leq{message_1RSB}
\be
&&
\T{\ul{\phi}}_{\mu \to i}(\ul{x}_{i})
=\int d\T{\V{\xi}}_{\mu \to i} \T{P}_{\mu \to i}(\T{\V{\xi}}_{\mu \to i}) \prod_{a=1}^{\PPc}\T{\phi}_{\mu \to i}(x_{ai}\mid \T{\V{\xi}}_{\mu \to i}),
\Leq{1RSBBP_main2tilde}
\\ &&
\ul{\phi}_{i \to \mu}(\ul{x}_{i})
=\int d\V{\xi}_{i \to \mu} P_{i \to \mu}(\V{\xi}_{i \to \mu}) 
\prod_{a=1}^{\PPc}\phi_{i \to \mu}(x_{ai} \mid \V{\xi}_{i \to \mu}),
\Leq{1RSBBP_tilde2main}
\ee
\subee
where $\V{\xi}_{i \to \mu},\T{\V{\xi}}_{\mu \to i}$ are appropriate random variables and $P_{i \to \mu},\T{P}_{\mu \to i}$ are their probability distributions: They characterize the correlations and fluctuations among replicas, while the symmetry does hold by the functional form of \Req{message_1RSB}. 

In the present case, it is appropriate to choose $\V{\xi},\T{\V{\xi}}$ as the parameters naturally parametrizing the messages for the non-replicated systems: According to~\cite{sakata2021scad}, in the $\tau \to 0$ limit, they are parametrized as
\subbe
\be
&&
\phi_{i\to \mu}(x_i)=\phi_{i\to \mu}(x_i\mid \Gamma_{i\to \mu},h_{i\to \mu})\propto e^{\beta \lb -\frac{1}{2}\Gamma_{i\to \mu}x_i^2+h_{i\to \mu}x_i-J(x_i) \rb},
\\ &&
\T{\phi}_{\mu \to i}(x_i)=\T{\phi}_{ \mu\to i}(x_i\mid \T{\Gamma}_{\mu\to i},\T{h}_{\mu\to i})\propto e^{\beta \lb -\frac{1}{2}\T{\Gamma}_{\mu\to i}x_i^2+\T{h}_{\mu\to i}x_i \rb}.
\ee
\subee
Hence we regard $\V{\xi}_{i \to \mu}=(\Gamma_{i \to \mu},h_{i \to \mu})$ and $\T{\V{\xi}}_{\mu \to i}=(\T{\Gamma}_{\mu \to i},\T{h}_{\mu \to i})$. Physically, $P_{i \to \mu}$ and $\T{P}_{\mu \to i}$ are interpreted to represent the distribution of the original BP parameters which comes from the solution's multiplicity of \Req{BP}. The insertion of \Req{message_1RSB} into \Req{BP-1RSB} yields a set of integral equations determining $P_{i \to \mu}$ and $\T{P}_{\mu \to i}$. As a particular solution of the integral equations, a set of equations explicitly determining the functional form of $\T{P}_{\mu \to i}$ from $P_{i \to \mu}$, and vice versa, are obtained as follows:
\be
&&
\hspace{-1cm}
\T{P}_{\mu \to i}(\T{\Gamma}_{\mu \to i},\T{h}_{\mu \to i})
\propto
\int \lb \prod_{j(\neq i)}d\Gamma_{j\to \mu}dh_{j\to \mu}P_{j\to \mu}(\Gamma_{j\to \mu},h_{j\to \mu}) \rb
\no \\
&& 
\hspace{-1cm}
\times
\lb 
\int d\V{x} \Phi_{\mu}(\V{x}) 
\prod_{j(\neq i)} \phi_{j \to \mu}\lb x_j\remid{|} \Gamma_{j\to \mu},h_{j\to \mu}\rb 
\rb^{\PPc}
\delta\lb 
\T{\Gamma}_{\mu \to i}-
f^{(\T{\Gamma})}_{\mu i}
\lb 
\sum_{j(\neq i)} A_{\mu j}^2 f^{(\chi)} \lb \Gamma_{j\to \mu},h_{j\to \mu} \rb 
\rb
\rb
\no \\
&& 
\hspace{-1cm}
\times
\delta\lb 
\T{h}_{\mu \to i}-
f^{(\T{h})}_{\mu i} 
\lb
\sum_{j(\neq i)} A_{\mu j} f^{(m)} \lb \Gamma_{j\to \mu},h_{j\to \mu} \rb
\rb
\rb
,
\Leq{BP_P_main2tilde}
\ee
where the functions $f^{(\T{\Gamma})}_{\mu i},f^{(\T{h})}_{\mu i},f^{(\chi)},f^{(m)}$ describe the parameter transformation rule corresponding to \Req{BP1}. The explicit formulas of these functions are not important here and are omitted; similarly, the equation for $P_{i \to \mu}(\Gamma_{i \to \mu},h_{i \to \mu})$ is also derivable but is omitted. Only the relevant property for the purpose of this subsection is the scaling of the arguments of $f^{(\T{\Gamma})}_{\mu i},f^{(\T{h})}_{\mu i}$. That is, the argument of $f^{(\T{\Gamma})}_{\mu i}$ follows a scaling for which the law of large numbers (LLN) applies ($A_{\mu j}^2 = O(1/N)$), whereas that of $f^{(\T{h})}_{\mu i}$ follows a scaling for which the central limit theorem (CLT) applies. This consideration means that $\T{\Gamma}$ takes a single deterministic value, while $\T{h}$ is distributed in a Gaussian manner. One noteworthy point is that the factor $\lb \int d\V{x} \Phi_{\mu}(\V{x}) \prod_{j(\neq i)} \phi_{j \to \mu}\lb x_j\remid{|} \Gamma_{j\to \mu},h_{j\to \mu} \rb \rb^{\PPc}$ in \Req{BP_P_main2tilde} may modulate the distribution form, but a straightforward calculation shows that this factor gives a trivial constant term and does not depend on $\{\Gamma_{j\to \mu}, h_{j\to \mu}\}_j$, and hence it has no effect on the final outcome. Hence, we can safely parameterize the distribution $\T{P}_{\mu \to i}(\T{\Gamma}_{\mu \to i},\T{h}_{\mu \to i})$ as
\be
\T{P}_{\mu \to i}(\T{\Gamma}_{\mu \to i},\T{h}_{\mu \to i})
=\delta\lb \T{\Gamma}_{\mu \to i} - \T{a} \rb
\frac{1}{\sqrt{2\pi \T{b}}}e^{-\frac{1}{2 \T{b}}\lb \T{h}_{\mu \to i} - \T{c} \rb^2},
\Leq{CBD}
\ee
with parameters $(\T{a},\T{b},\T{c})$. This distribution form leads to a simple analytic form of $\ul{\T{\phi}}_{\mu \to i}$:
\be
&&
\T{\ul{\phi}}_{\mu \to i}(\ul{x}_{i})
\propto \int d\T{\Gamma}_{\mu \to i}d\T{h}_{\mu \to i} \T{P}_{\mu \to i}(\T{\Gamma}_{\mu \to i},\T{h}_{\mu \to i}) \prod_{a=1}^{\PPc}\T{\phi}_{\mu \to i}(x_{ai}\mid \T{\Gamma}_{\mu \to i},\T{h}_{\mu \to i}),
\no \\ &&
\propto 
e^{
-\frac{1}{2}\beta \T{\Gamma}_{\mu\to i}\sum_{a=1}^{\PPc}x_{ai}^2 
+\frac{1}{2}\beta^2 \T{B}_{\mu\to i}\lb \sum_{a=1}^{\PPc}x_{ai} \rb^2
+\beta \T{C}_{\mu\to i}\sum_{a=1}^{\PPc}x_{ai} 
}.
\Leq{1RSBBP_tilde_parametrized}
\ee
The coefficients $\T{\Gamma}_{\mu\to i},\T{B}_{\mu\to i},\T{C}_{\mu\to i}$ have a dependence on $\T{a},\T{b},\T{c}$ but the explicit form is omitted here. This message form leads to a simple formula also for $\ul{\phi}_{i \to \mu}$ from \Req{BP2-replicated} as
\be
&&
\ul{\phi}_{i \to \mu}(\ul{x}_{i})\propto 
e^{
-\frac{1}{2}\beta\Gamma_{i \to \mu}\sum_{a=1}^{\PPc}x_{ai}^2 
+\frac{1}{2}\beta^2 B_{i \to \mu}\lb \sum_{a=1}^{\PPc}x_{ai}\rb^2 
+\beta C_{i \to \mu}\sum_{a=1}^{\PPc}x_{ai} 
-\beta \sum_{a=1}^{\PPc}J(x_{ai})
},
\Leq{1RSBBP_main_parametrized}
\ee
where 
\be
&&
\Gamma_{i \to \mu}=\sum_{\nu (\neq \mu)}\T{\Gamma}_{\nu \to i},
\quad
B_{i \to \mu}=\sum_{\nu (\neq \mu)}\T{B}_{\nu \to i},
\quad
C_{i \to \mu}=\sum_{\nu (\neq \mu)}\T{C}_{\nu \to i}.
\Leq{1RSBBP_tilde2main_parametrized}
\ee
\BReqs{1RSBBP_tilde_parametrized}{1RSBBP_main_parametrized} manifest how these 1RSB messages can be parametrized, and \Req{1RSBBP_tilde2main_parametrized} yields the update rule from $\{\ul{\T{\phi}}_{\mu \to i}\}$ to $\{\ul{\phi}_{i \to \mu}\}$ in the parameter level. This is the consequence we aimed to obtain in this subsection.

The opposite update rule from $\{\ul{\phi}_{i \to \mu}\}$ to $\{\ul{\T{\phi}}_{\mu \to i}\}$ requires certain intricate average operations over $\{\ul{\phi}_{i \to \mu}\}$, and the details are deferred to \Rsec{Message}. Based on the 1RSB-BP algorithm, the 1RSB-AMP algorithm and the 1RSB-SE equations are obtained from the standard linear perturbation and the standard convergence discussion in the large $N$ limit, respectively. Those details are also in \Rsec{Message}.

\subsection{Summary of the 1RSB-AMP and 1RSB-SE formulas}\Lsec{Summary of}
Here we summarize the 1RSB-AMP algorithm and the corresponding 1RSB-SE equations. The 1RSB-AMP algorithm in the limit $\beta \to \infty$ is given by
\subbe
\Leq{AMP_1RSB_simpler}
 \be
 &&
  \hspace{-1cm}
\chi^{(t)}=\frac{1}{M}\sum_{\mu}\sum_{i}A_{\mu i}^2 \chi_{i}^{(t)},\quad
Q_{}^{(t)}=\frac{1}{M}\sum_{\mu}\sum_{i}A_{\mu i}^2 Q_{i}^{(t)},\quad
q_{0}^{(t)}=\frac{1}{M}\sum_{\mu}\sum_{i}A_{\mu i}^2 \lb m_{i}^{(t)} \rb^2,
 \\ &&
  \hspace{-1cm}
R_{\mu}^{(t)}=
\frac{
y_{\mu}-\sum_{i}A_{\mu i}m_{i}^{(t)}
}{
 \chi^{(t)}+\rPPc \lb Q^{(t)}-q_{0}^{(t)} \rb
 }
+R_{\mu}^{(t-1)},
  \\ &&
  \hspace{-1cm}
 \Gamma^{(t+1)}= \frac{\alpha}{\chi^{(t)}},
  \\ &&
  \hspace{-1cm}
B^{(t+1)}
 =
\frac{\alpha}{ \chi^{(t)}}\frac{ Q^{(t)}-q_{0}^{(t)}
}{ \chi^{(t)}+\rPPc \lb Q^{(t)}-q_{0}^{(t)} \rb}
,
  \\ &&
  \hspace{-1cm}
C_{i}^{(t+1)}
 =
\sum_{\mu } 
A_{\mu i}R_{\mu}^{(t)}
+
\frac{ 
\alpha
}{ 
\chi^{(t)}+\rPPc \lb Q^{(t)}-q_{0}^{(t)} 
\rb}m_{i}^{(t)},
 \\ &&
  \hspace{-1cm}
m_{i}^{(t+1)}=\Ave{\hat{x}_{i}^{(t+1)} }_{i}^{(t+1)},\quad
Q_{i}^{(t+1)}=\Ave{\lb \hat{x}_{i}^{(t+1)}\rb^2 }_{i}^{(t+1)},\quad
\chi_{i}^{(t+1)}=\Ave{\partial_{C_{i}^{(t+1)}}\hat{x}_{i}^{(t+1)} }_{i}^{(t+1)},
 \ee
 \subee
where $t=0,1,\ldots$ describes the algorithm step and
\be
&&
\rPPc=\lim_{\beta \to \infty}\beta \PPc,
\\ &&
\hat{x}_{i}^{(t)}=\hat{x}\lb \Gamma_i^{(t)},\sqrt{B_i^{(t)}} z+C_{i}^{(t)} \rb,
\quad 
\Ave{(\cdots)}_i^{(t)}
=
\frac{
\int Dz (\cdots)e^{\rPPc \hat{g}\lb \Gamma_i^{(t)},\sqrt{B_i^{(t)}} z+C_{i}^{(t)} \rb }
}{
\int Dz e^{\rPPc \hat{g}\lb \Gamma_i^{(t)},\sqrt{B_i^{(t)}} z+C_{i}^{(t)} \rb }
}.
\ee
In this zero-temperature scaling, $\rPPc$ is kept finite while $\beta\to\infty$, and hence the finite-temperature parameter behaves as $\PPc=\rPPc/\beta\to0$ for any fixed $\rPPc$.
Thus, the algorithmic tuning parameter in the following zero-temperature equations is $\rPPc$, rather than the bare replica number $\PPc$.
In addition, the 1RSB-SE equations describing the macroscopic dynamics of 1RSB-BP/AMP are given by
\subbe
\Leq{SE_1RSB}
\be
&&
\chi^{(t)}=\E_{x_0,z_0}\Ave{\partial_{C^{(t)}} \hat{x}(\Gamma^{(t)},\sqrt{B^{(t)}}z_1+C^{(t)}(x_0,z_0))}^{(t)}_{z_1\mid x_0,z_0},
\Leq{SE_1RSB^{(t)}chi}
\\ &&
Q^{(t)}=\E_{x_0,z_0}\Ave{ \hat{x}^2(\Gamma^{(t)},\sqrt{B^{(t)}}z_1+C^{(t)}(x_0,z_0))}^{(t)}_{z_1\mid x_0,z_0},
\Leq{SE_1RSB^{(t)}Q}
\\ &&
q_{0}^{(t)}=\E_{x_0,z_0}\lb \Ave{\hat{x}(\Gamma^{(t)},\sqrt{B^{(t)}}z_1+C^{(t)}(x_0,z_0))}^{(t)}_{z_1\mid x_0,z_0}\rb^2,
\Leq{SE_1RSB^{(t)}q0}
\\ &&
\epsilon_{q_0}^{(t)}=\E_{x_0,z_0}\lb x_0-\Ave{\hat{x}(\Gamma^{(t)},\sqrt{B^{(t)}}z_1+C^{(t)}(x_0,z_0))}^{(t)}_{z_1\mid x_0,z_0} \rb^2,
\Leq{SE_1RSB_MSE}
\\ &&
\Gamma^{(t+1)}=\frac{\alpha}{\chi^{(t)}},
\Leq{SE_1RSB_Gamma}
\\ &&
B^{(t+1)}=\frac{\alpha}{\chi^{(t)}}
\frac{(Q^{(t)}-q_{0}^{(t)})}{\chi^{(t)}+\rPPc(Q^{(t)}-q_{0}^{(t)})},
\Leq{SE_1RSB_B}
\\ &&
C^{(t+1)}(x_0,z_0)
=\frac{\alpha x_0}{\chi^{(t)}+\rPPc(Q^{(t)}-q_{0}^{(t)})}
+\frac{\sqrt{\alpha \epsilon_{q_0}^{(t)}}z_0}{\chi^{(t)}+\rPPc(Q^{(t)}-q_{0}^{(t)})},
\Leq{SE_1RSB_C}
\ee
\subee
where
\be
&&
\Ave{(\cdots)}^{(t)}_{z_1\mid x_0,z_0}
=
\frac{
\int Dz_1 e^{\rPPc \hat{g} \lb \Gamma^{(t)}, \sqrt{B^{(t)}} z_1+C^{(t)}(x_0,z_0) \rb }
(\cdots)
}{
\int Dz_1 e^{\rPPc \hat{g}\lb \Gamma^{(t)},\sqrt{B^{(t)}} z_1+C^{(t)}(x_0,z_0) \rb}
},
\\ &&
\E_{x_0,z_0}(\cdots)=\int dx_0 P_0(x_0) \int Dz_0(\cdots).
\ee
These 1RSB-AMP and 1RSB-SE equations coincide, at the level of their formal structure, with those in~\cite{antenucci2019asp,barbier2023l0asp,Barbier2025JSTAT,saglietti2019gasp}. All properties specific to our problem setting appear through $g,\hat{x},\chi$ discussed in \Rsec{One-dimensional estimator}. Analytical and numerical techniques for actually evaluating these quantities are summarized in \Rsec{Computational}.

There is a noteworthy remark. Here, $m_i=\Ave{\hat{x}_i}_i$ in \Req{AMP_1RSB_simpler} is the average of $x_i$ using the full marginal, and this ``average'' corresponds, at the 1RSB description, to taking an ensemble average over the exponentially many BP fixed points. Therefore, to be honest, it is not entirely clear in what theoretical sense $m_i$ should be regarded as a suitable estimator of $x_{0i}$. However, accepting this interpretation is convenient, so in what follows we regard $m_i = \Ave{x_i}_i$ as the 1RSB estimator of $x_{0i}$, and accordingly regard $\epsilon_{q_0}$ as the associated MSE.

\subsection{Free entropy and related quantities}\Lsec{Free entropy}
The free entropy $s$ of the replicated system \NReq{replicated Boltzmann-factor} is given by $s=\frac{1}{N}\log \ul{Z}_{\beta,\tau}$. According to the 1RSB description, this is connected to the complexity describing the number of pure states. Namely, denoting the number of pure states given free energy value $f=-\frac{1}{N\beta}\log Z_{\beta,\tau}$ as $\mathscr{N}(f)$, the complexity is defined by
\be
\Sigma(f)=\frac{1}{N}\log \mathscr{N}(f),
\ee
and the free entropy is the Legendre transform of this complexity:
\be
s=s(\PPc)=\max_{f}\lbb -\beta \PPc f+\Sigma(f) \rbb,
\Leq{Sigma2s-1}
\ee
where $\PPc$ plays the role of the conjugate variable against the free energy $f$. In the $\beta \to \infty$ limit, the free energy converges to the energy $f\to u$ and 
\be
s(\rPPc)=\max_{u}\lbb -\rPPc u+\Sigma(u) \rbb.
\Leq{Sigma2s-2}
\ee
Here, $\rPPc$ behaves as the conjugate variable against the energy $u$. These quantities are essential for characterizing the thermodynamic behavior of the system and are also of interest in this paper. 

Unfortunately, their computations are rather cumbersome and we defer the details to \Rsec{Computations of} and here only quote the result. The free entropy is computed by
\be
s=
\frac{\alpha}{2}
\lbb
\log \frac{\chi}{Y}-\frac{\rPPc \epsilon_{q_0}}{Y}
\rbb
- \frac{\alpha}{2}\rPPc \frac{ \rho  \sigma_x^2 }{Y}
+
\E_{x_0,z_0}
\log \int Dz_1 e^{\rPPc \hat{g}\lb \Gamma,\sqrt{B}z_1+C(x_0,z_0)\rb},
\Leq{s-1RSB}
\ee
where we put 
\be
Y=\chi+\rPPc(Q-q_0).
\ee
For the macroscopic parameters $(\epsilon_{q_0},\chi,\ldots)$, the fixed point values of 1RSB-SE are inserted. Then, from the generic property of the Legendre transform, the optimizer $u^*$ in the maximization problem \NReq{Sigma2s-2} is computed from the derivative of $s$ as
\be
u^*(\rPPc)=\argmax_u\lb -\rPPc u+\Sigma(u) \rb =-\Part{s}{\rPPc}{}.
\Leq{u^*-1RSBSE}
\ee
An easy prescription to compute $u^*$ is to take the numerical differentiation of $s$ w.r.t. $\rPPc$, though, as shown in \Rsec{Replica}, in the replica formulation one can obtain a more concise expression evaluated without numerical differentiation. Once $u^*$ is obtained, the complexity value is computed as
\be
\Sigma(\rPPc)=\Sigma(u^*(\rPPc))=s+\rPPc u^*(\rPPc). 
\Leq{complexity}
\ee
The multiplicity of the pure states can be quantified by this. 

\subsection{Results} \Lsec{Results}
Here we discuss consequences derived from \Reqs{AMP_1RSB_simpler}{SE_1RSB}. In what follows, we set $\sigma_x^2 = 1$ unless otherwise stated.

\subsubsection{Consistency check of 1RSB-AMP and 1RSB-SE}
Let us first examine the consistency between 1RSB-SE and 1RSB-AMP. \Rfigs{se_amp_case1}{se_amp_case2} compare at $(a,\lambda)=(3,1)$ the 1RSB-SE predictions with the 1RSB-AMP behaviors for two representative cases $(\alpha,\rho)=(0.5,0.1)$ and $(0.5,0.4)$, respectively. 
The value $\rPPc=1$ is chosen just as a representative example; we stress that this condition $\rPPc=1$ does not correspond to the finite-temperature condition $\PPc=1$, under which the 1RSB formula reduces to the RS one.
\footnote{Note that no finite value of $\rPPc$ reduces the zero-temperature 1RSB formula to the RS one in the zero-temperature limit $\rPPc=\lim_{\beta\to\infty}\beta\PPc$.}
In each panel, we plot MSE $\epsilon_{q_0}$, susceptibility $\chi$, and overlaps $Q,q_0$ against the algorithm iteration. A noteworthy point is that one of the two cases is in the success phase (\Rfig{se_amp_case1}) sustaining RS while the other is in the failure phase (\Rfig{se_amp_case2}) exhibiting RSB. 
\begin{figure}[htbp]
\begin{center}
\vspace{0mm}
\includegraphics[width=0.45\columnwidth]{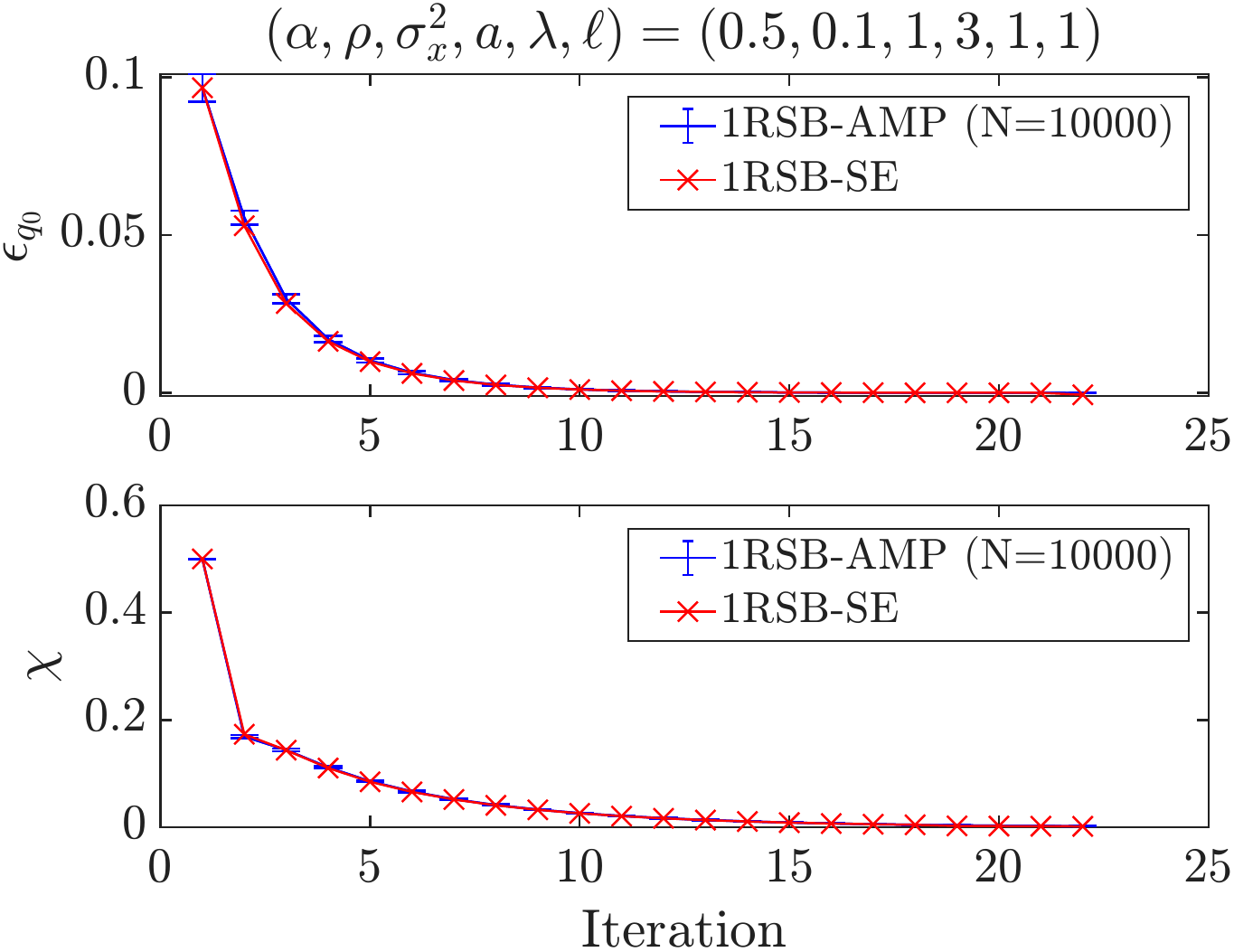}
\includegraphics[width=0.45\columnwidth]{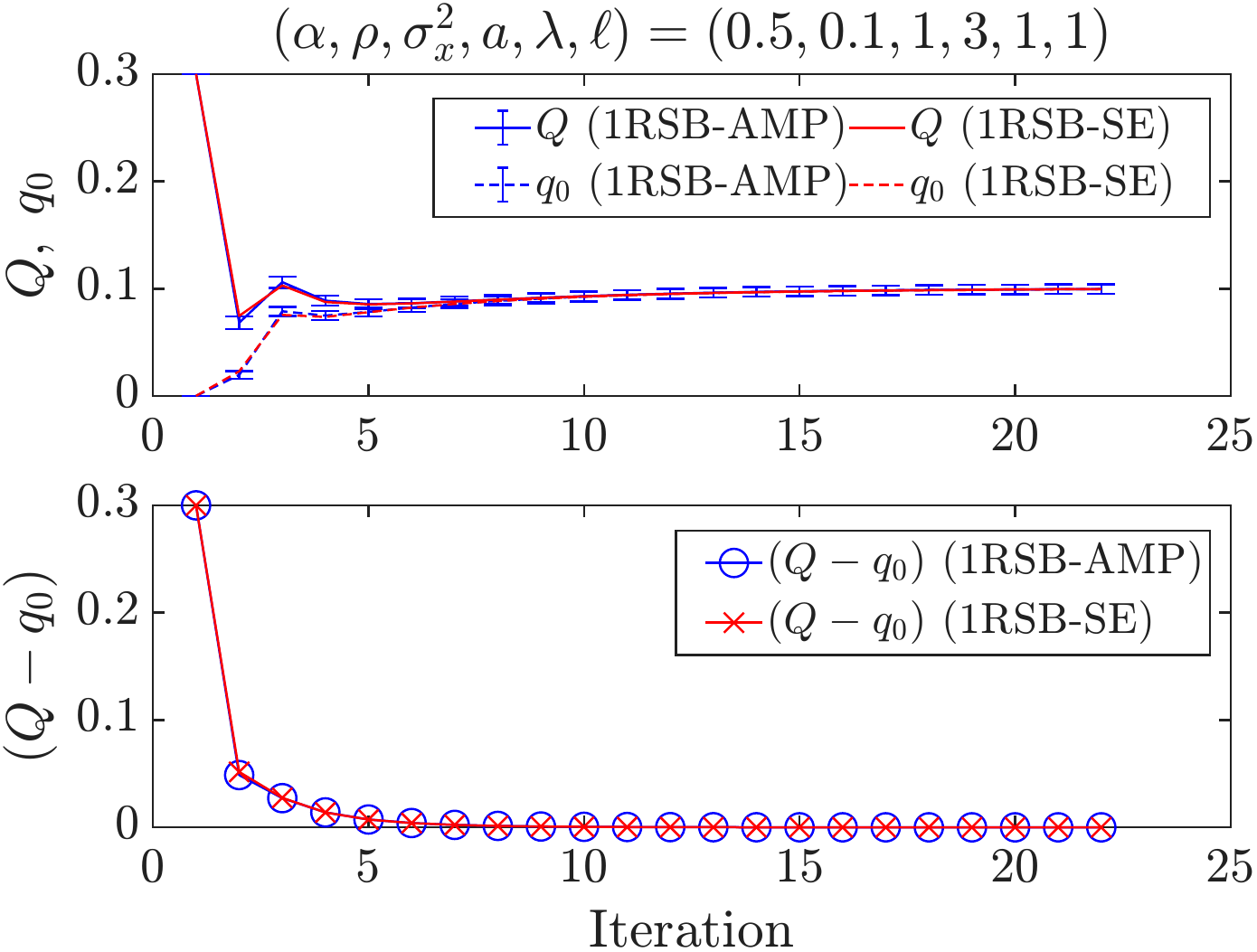}
\vspace{0mm}
\caption{
Consistency check between 1RSB-SE and 1RSB-AMP for $(\alpha,\rho)=(0.5,0.1)$ at $(a,\lambda,\rPPc)=(3,1,1)$ (success phase). The curves show the iteration dependence of $\epsilon_{q_0}$  and $\chi$ (left) and the overlaps (right). The AMP result (plus marker) is obtained by the average over 100 different realizations of $(\V{A},\V{x}_0)$, with standard error bars included. The agreement demonstrates that 1RSB-SE correctly describes the macroscopic dynamics of 1RSB-AMP in this phase.
}
\Lfig{se_amp_case1}
\end{center}
\end{figure}
\begin{figure}[htbp]
\begin{center}
\vspace{0mm}
\includegraphics[width=0.45\columnwidth]{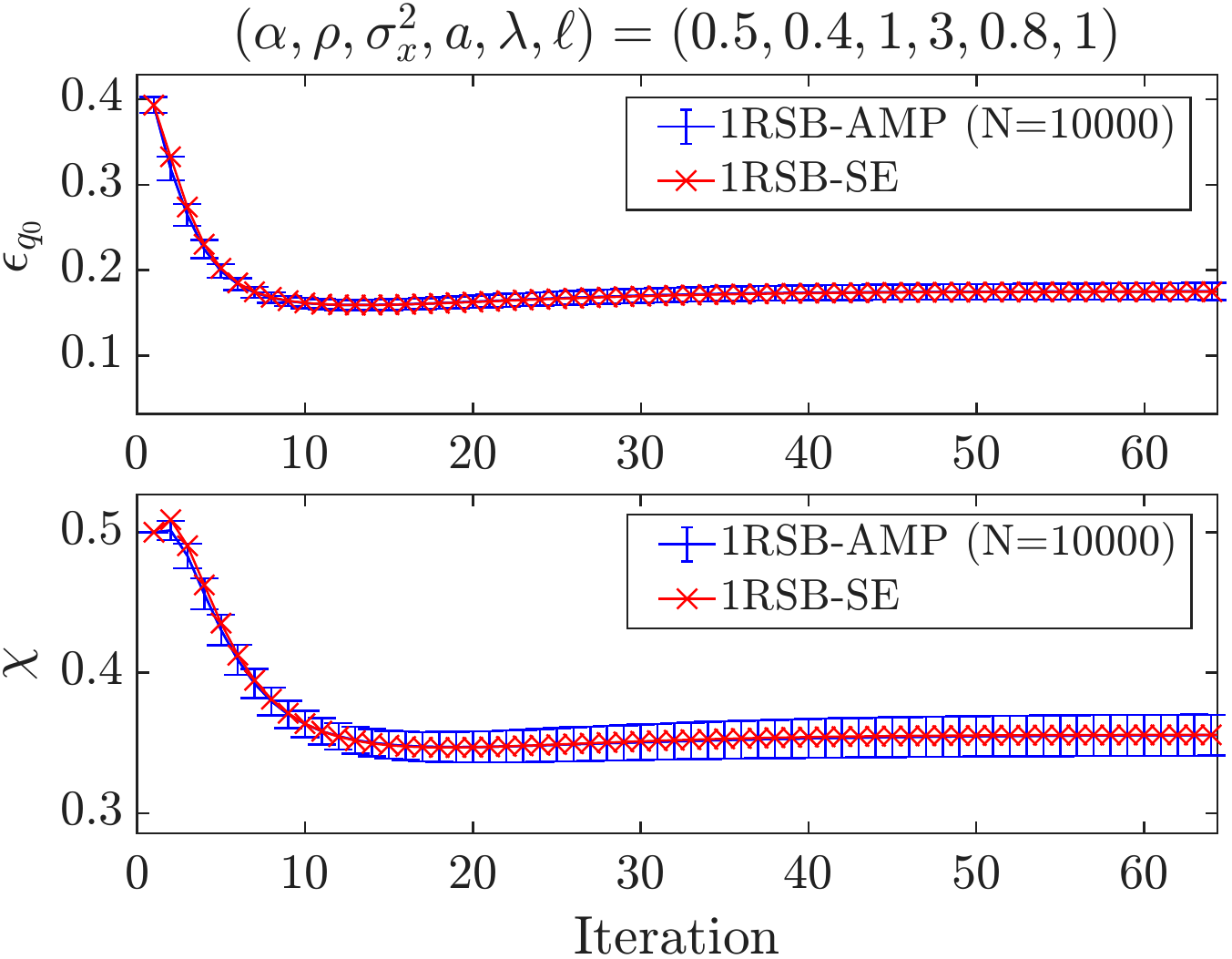}
\includegraphics[width=0.45\columnwidth]{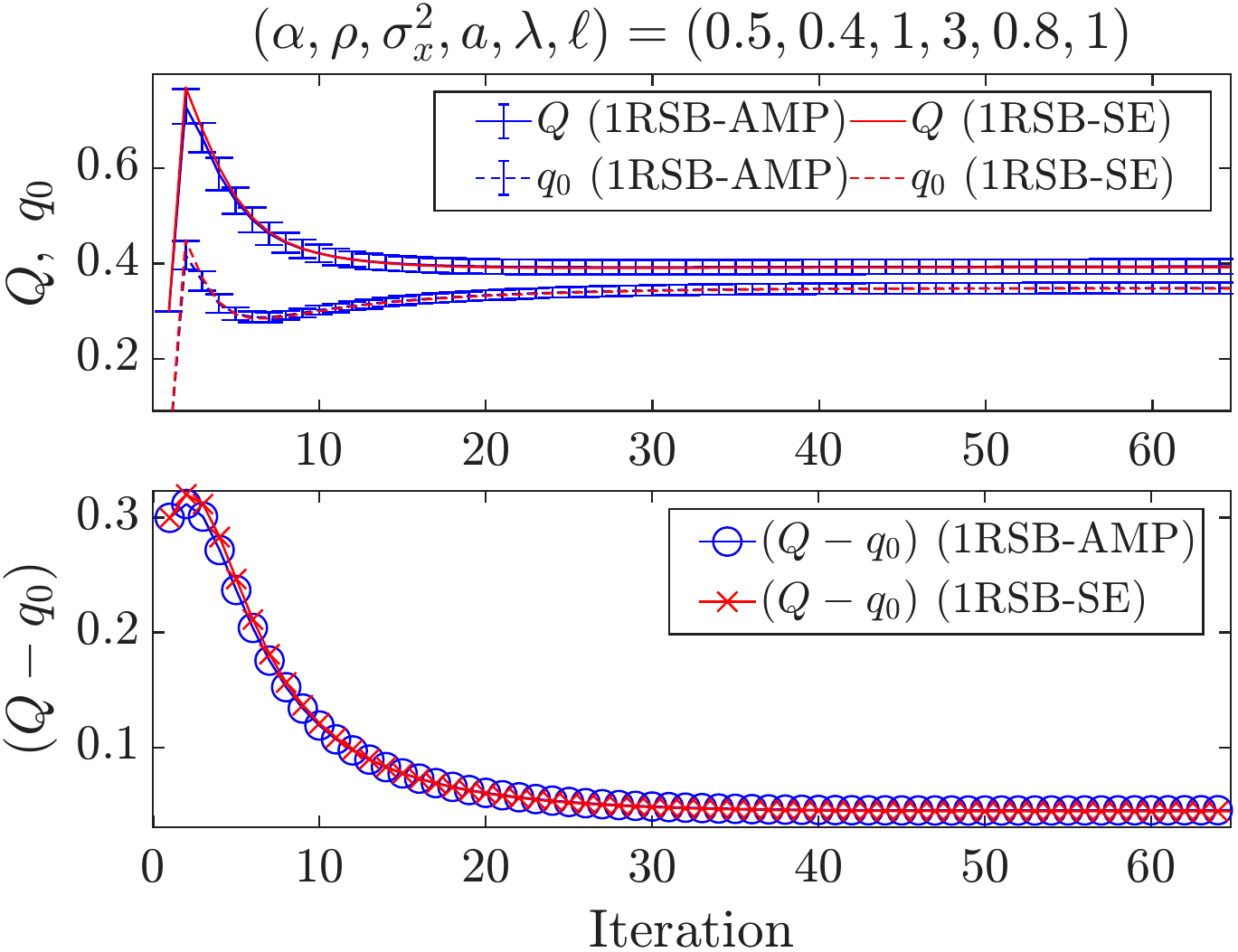}
\vspace{0mm}
\caption{
The counterpart of \Rfig{se_amp_case1} for $(\alpha,\rho)=(0.5,0.4)$ and $(a,\lambda,\rPPc)=(3,0.8,1)$ (failure phase). Again, the 1RSB-AMP trajectories agree well with that by 1RSB-SE, demonstrating the consistency between 1RSB-AMP and 1RSB-SE. 
}
\Lfig{se_amp_case2}
\end{center}
\end{figure}
In both cases, the 1RSB-SE curves fairly agree with those of the 1RSB-AMP, demonstrating the consistency between them. As expected, in the former case (\Rfig{se_amp_case1}) $(Q-q_0)$ tends to zero and comes back to RS, while in the latter case (\Rfig{se_amp_case2}) it remains finite, showing the difference between the RS and RSB phases.

Another interesting observation in the RSB case (\Rfig{se_amp_case2}) is that the susceptibility $\chi$ has relatively larger error bars compared to the other quantities: This is actually natural because the susceptibility does not have the self-averaging property and may be different from sample to sample in RSB phases~\cite{MezardParisiVirasoro1987}. To demonstrate this, in \Rfig{susceptibility comparison}, we show the simultaneous plot of $\chi$ for 20 different realizations of $(\V{A},\V{x}_0)$ against iteration in the RS (left) and RSB (right) cases.  
\begin{figure}[htbp]
\begin{center}
\vspace{0mm}
\includegraphics[width=0.45\columnwidth]{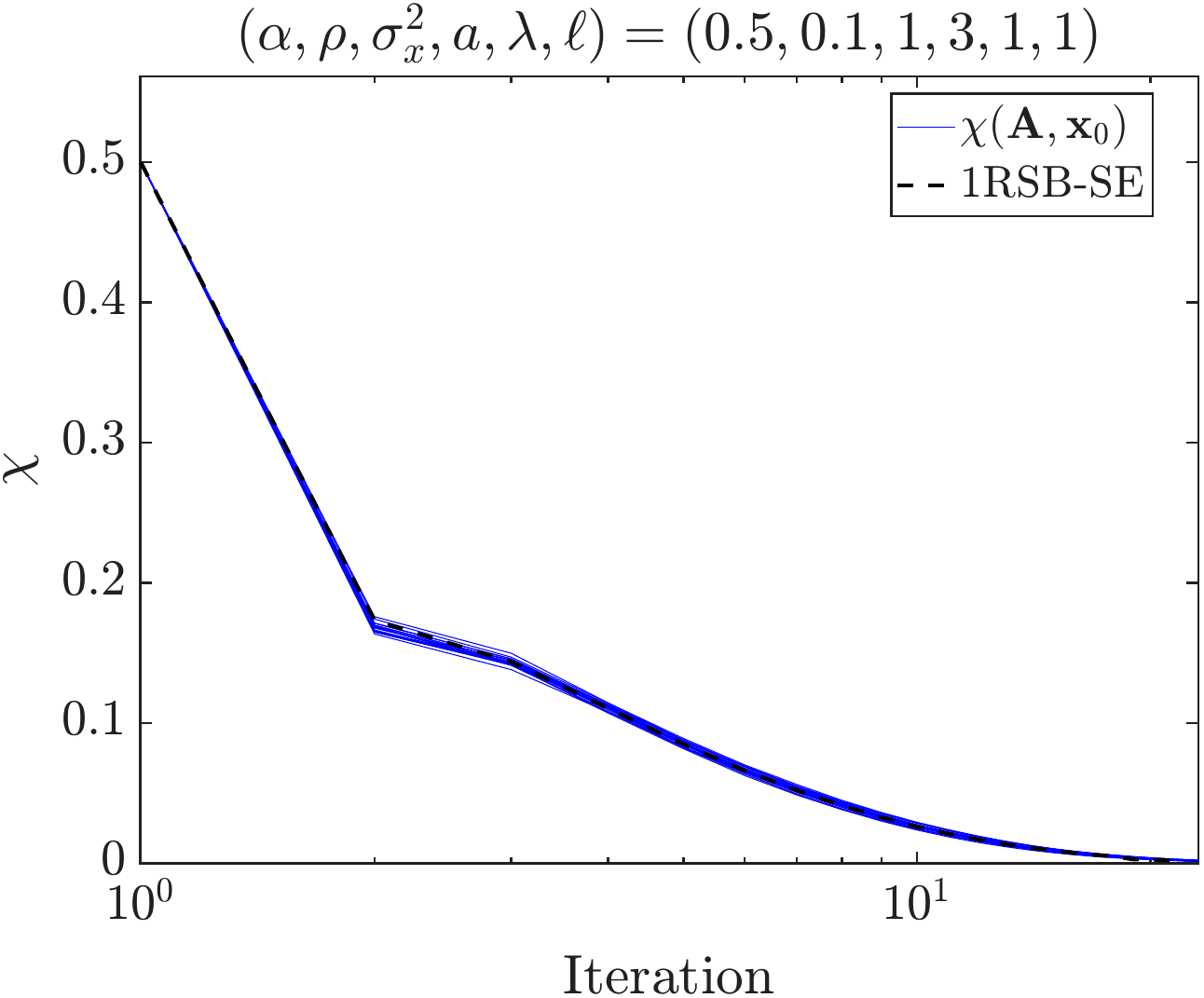}
\includegraphics[width=0.45\columnwidth]{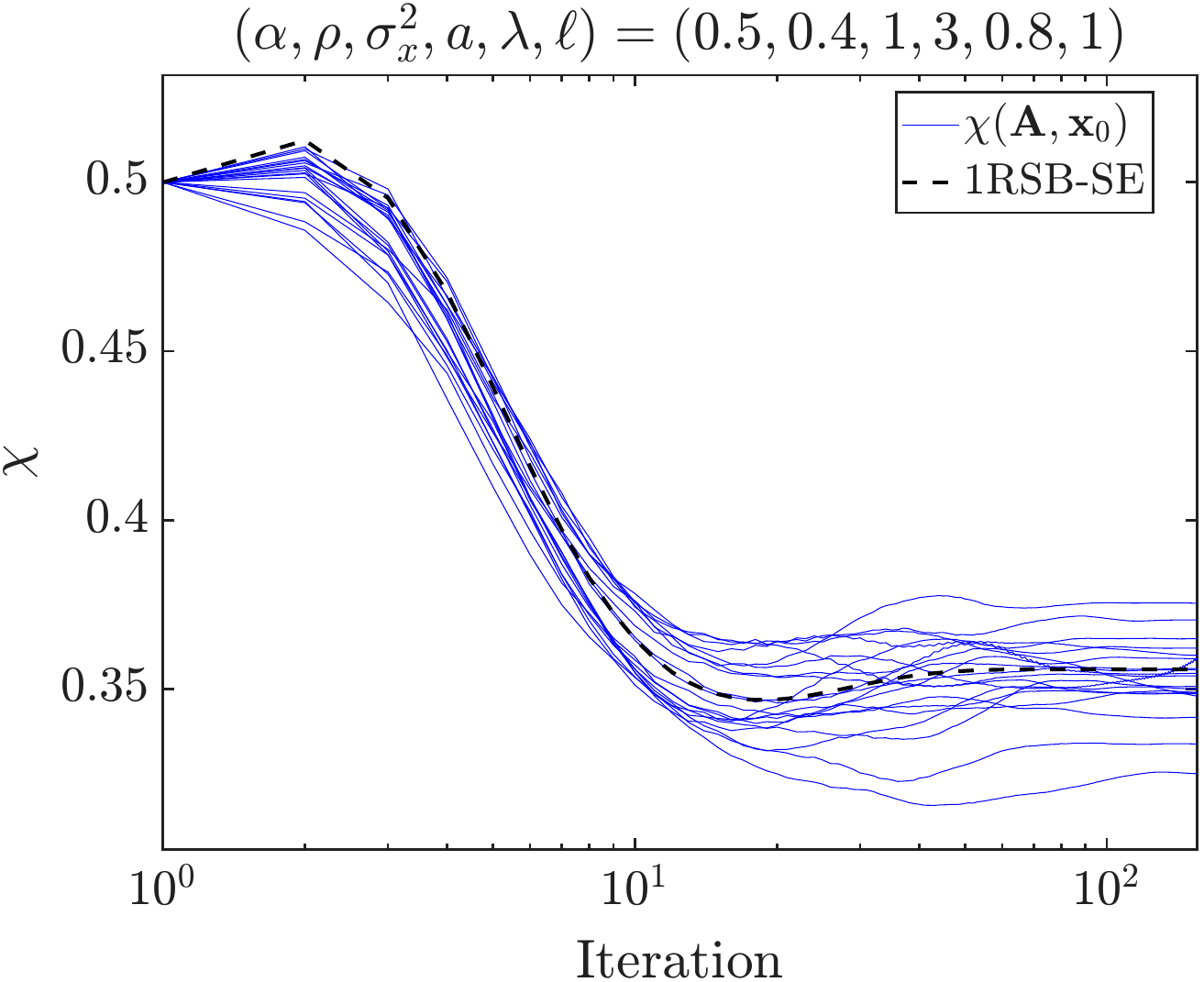}
\vspace{0mm}
\caption{
Simultaneous plot of $\chi$ against iteration for 20 realizations of $(\V{A},\V{x}_0)$. The parameters of the left and right panels are identical to those of \Rfig{se_amp_case1} and \Rfig{se_amp_case2}, respectively. The difference among samples is tiny for the RS case (left) but is substantial for the RSB case (right). The black dashed lines are the 1RSB-SE predictions and are given for reference.
}
\Lfig{susceptibility comparison}
\end{center}
\end{figure}
This figure shows that, in the RSB case, the behavior of the susceptibility $\chi$ varies significantly from sample to sample. Nevertheless, as seen in \Rfig{se_amp_case2}, the sample average of $\chi$ remains in good agreement with the prediction of 1RSB-SE. This suggests that although $\chi$ itself is non-self-averaging, its sample average maintains a stable macroscopic behavior that is accurately captured by 1RSB-SE. A similar phenomenon has been reported in a more general theoretical framework of dynamics in~\cite{Opper2016TAPGeneralInvariant}, where time-correlation functions at very large system sizes show a systematic deviation from the macroscopic predictions but their sample average remains consistent with it.

Although we have not performed a macroscopic analysis of the 1RSB instability (which can be done either via the replica method or via a macroscopic evaluation of the dynamical instability of 1RSB-AMP), we think it is better to emphasize that the RS-unstable phase considered here is probably also 1RSB-unstable. In fact, within the RS-unstable phase, we have checked with varying parameter values and initial conditions that 1RSB-AMP never converged microscopically, supporting the instability of the 1RSB description in that phase. This is further supported from the fact that the AT condition given by \Req{AT_RS} is derived from a local stability analysis of the RS solution and is actually violated in the RS-unstable failure phase; it is empirically known that once such local stability is lost, full-step RSB is required to obtain the exact solution in most cases. These suggest that our 1RSB-AMP and 1RSB-SE are unlikely to be exact for describing the true thermodynamic behavior of the system.

Nevertheless, as these figures show, 1RSB-AMP and 1RSB-SE agree very well at the macroscopic level, and 1RSB-AMP in fact achieves perfect reconstruction in certain parameter regions. This indicates that, even if they do not provide an exact description of the thermodynamic behavior, they still offer a useful algorithm associated with a precise macroscopic characterization. The advantage of this point will be further pursued later when we investigate the perfect-reconstruction limit.

Before moving on, let us note that the 1RSB theoretical considerations also show that the success phase must necessarily be RS. In the success phase, MSE must converge to zero and it is defined as the difference between $x_0$ and $\Ave{\hat{x}}^{(t)}$ as shown in \Req{SE_1RSB_MSE} in the 1RSB level. However, one can show that $\Ave{\hat{x}}^{(t)}$ does not coincide with $x_0$ unless $B \propto (Q - q_0)$ becomes zero, i.e., unless the solution return to the RS form: From the asymptotic analysis in the large $C$ limit, this averaged quantity of $\hat{x}$ behaves as
\be
&&
\Ave{\hat{x}}_{\Gamma,B,C}\approx 
\frac{C}{\Gamma}
+
\frac{\rPPc BC}{\Gamma^2(1-B\rPPc/\Gamma)}
.
\ee
The second term yields a bias significantly deviating from the signal term $C/\Gamma$ ($C$ corresponds to $h$ in \Rfig{one-dim estimator}). Hence, $\Ave{\hat{x}}$ has a strong bias as long as $B \propto (Q - q_0)$ does not vanish. This implies that the success or perfect reconstruction solution is always RS.  The detailed computation of the above asymptotics is at the end of \Rsec{Computational}.

\subsubsection{Fixed point analysis of 1RSB-SE and macroscopic quantities' behavior}
Next, we examine the behavior of macroscopic quantities obtained from the analysis of the 1RSB-SE fixed point. Those quantities are plotted against $\rPPc$ in \Rfig{orderparam-ell} where the cases $(\alpha,\rho ,\sigma_x^2,a,\lambda)=(0.5, 0.4, 1, 3, 0.8)$ and $(0.5, 0.25, 1, 3, 1)$, both are in the RS-failure phase, are treated.
\begin{figure}[htbp]
\begin{center}
\vspace{0mm}
\includegraphics[width=0.49\columnwidth]{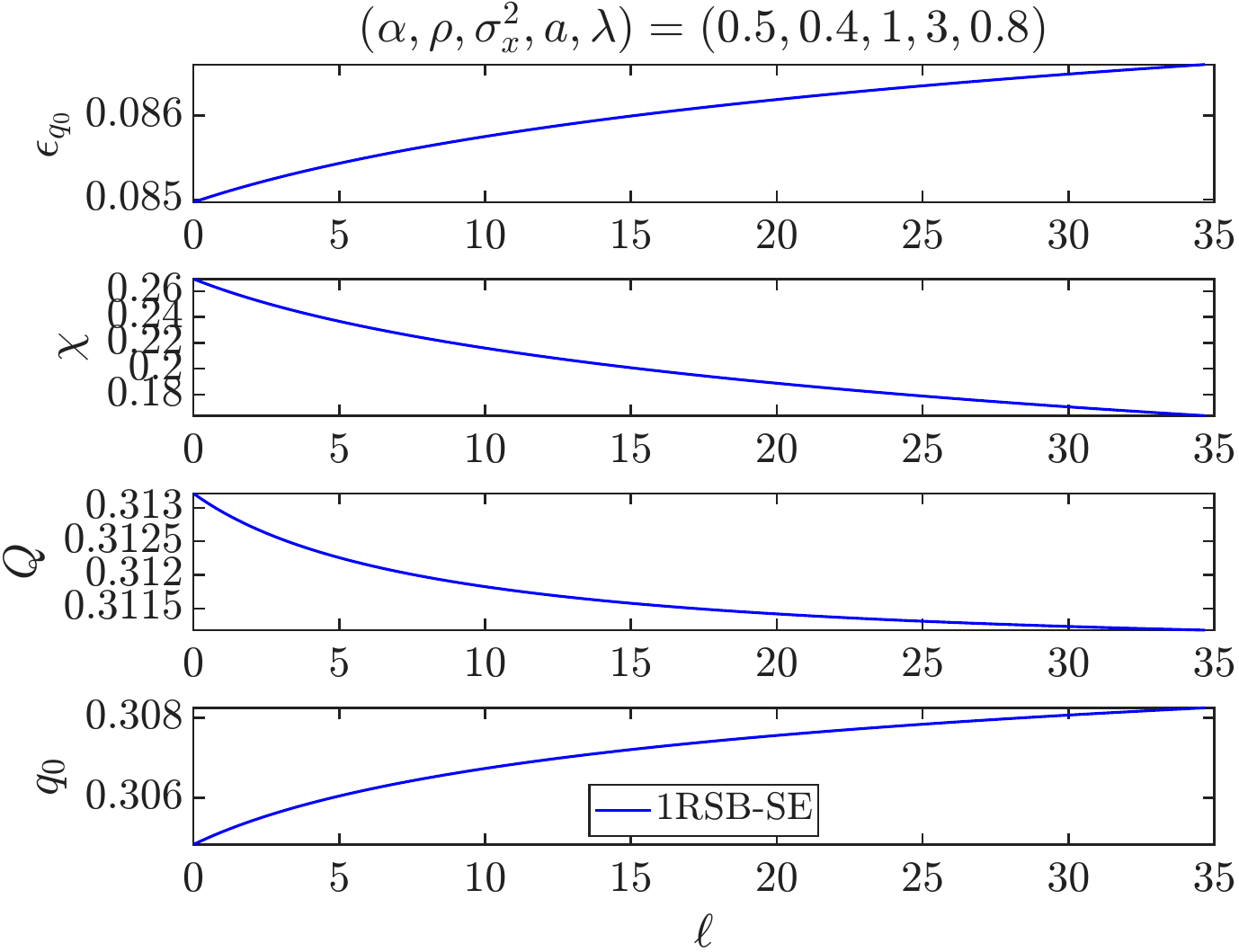}
\includegraphics[width=0.45\columnwidth]{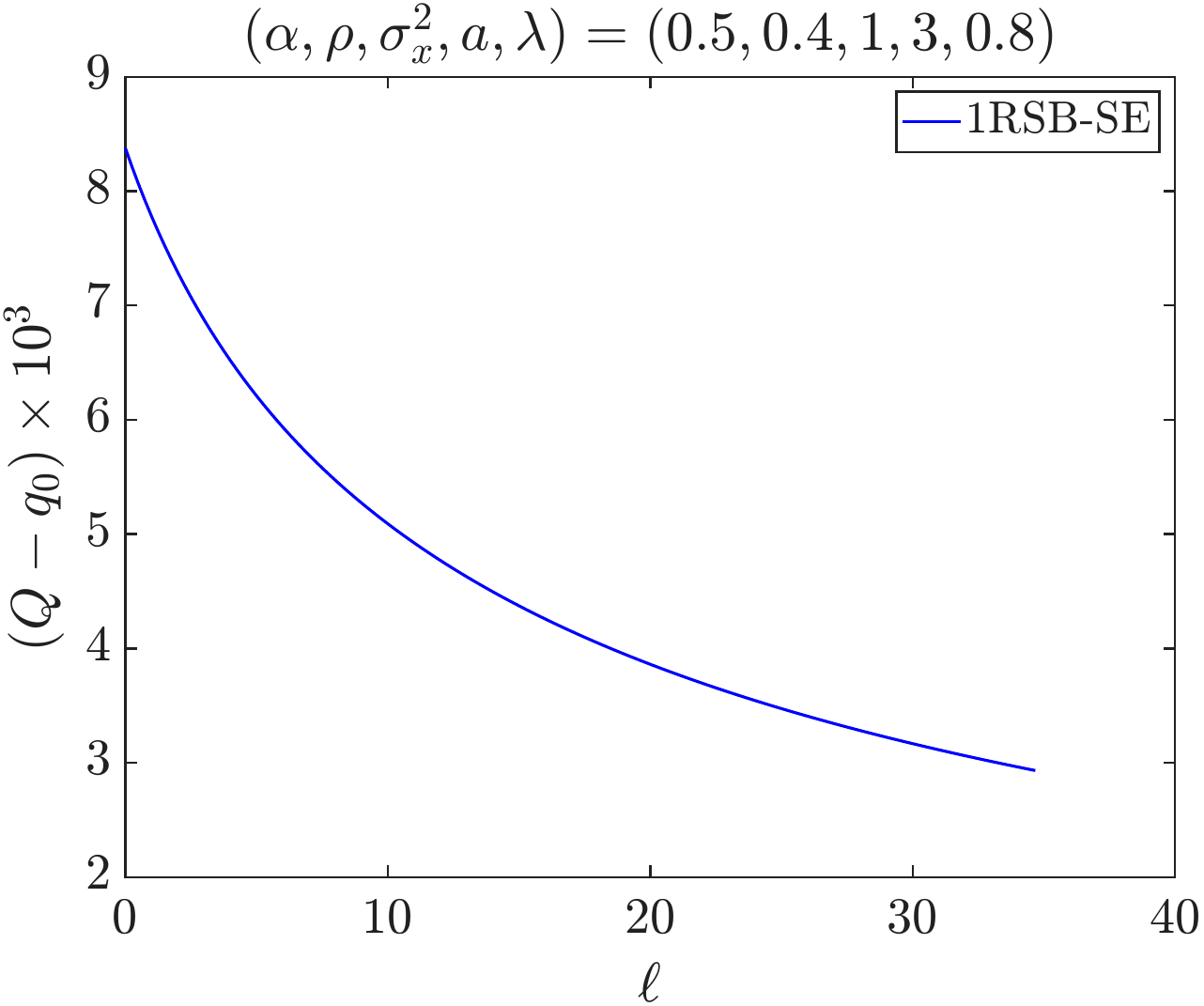}
\includegraphics[width=0.48\columnwidth]{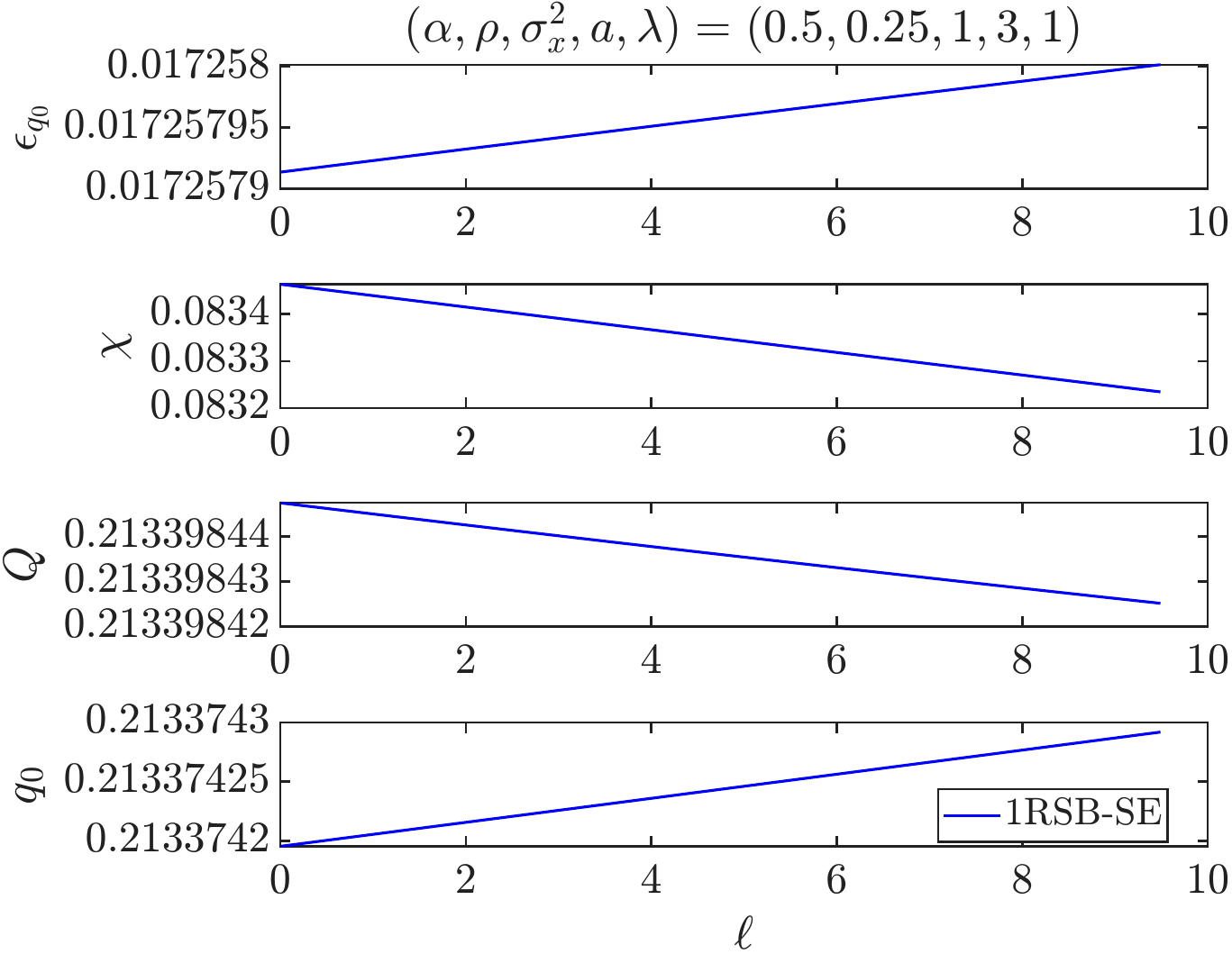}
\includegraphics[width=0.47\columnwidth]{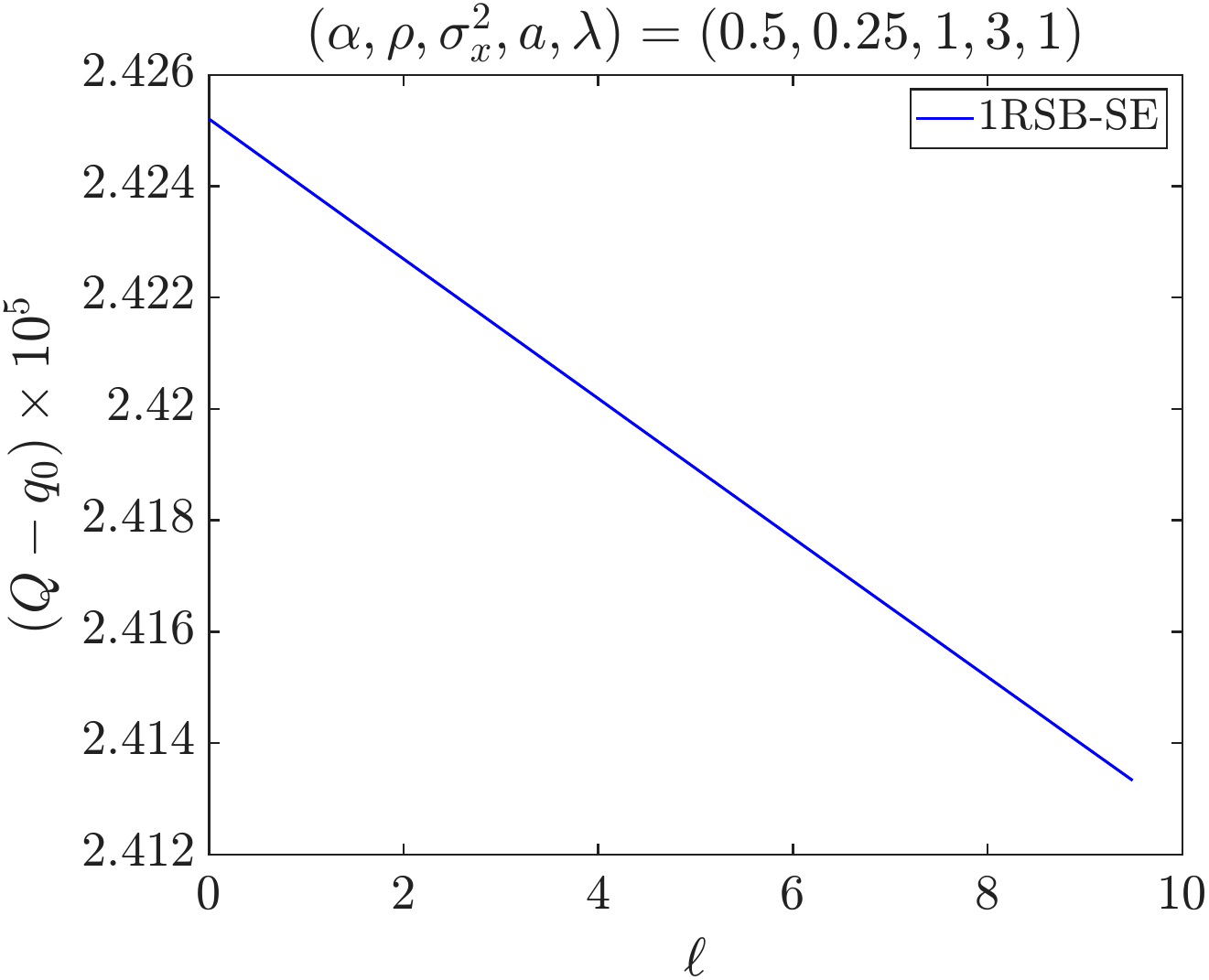}
\caption{Macroscopic quantities at the 1RSB-SE fixed point for $(\alpha,\rho ,\sigma_x^2,a,\lambda)=(0.5, 0.4, 1, 3, 0.8)$ (upper) and $(0.5, 0.25, 1, 3, 1)$ (lower) plotted against $\rPPc$. The left panels are for $\epsilon_{q_0},\chi,Q,q_0$ while the right ones are of $Q-q_0$.   
}
\Lfig{orderparam-ell}
\end{center}
\end{figure}
A characteristic feature observed in the both cases is the opposite $\rPPc$-dependence of the two overlaps $Q$ and $q_0$: As $\rPPc$ increases, the norm $Q$ monotonically decreases while the inter-state overlap $q_0$ increases. As a consequence, the quantity $Q - q_0$, which quantifies the degree of RSB, tends to decrease with $\rPPc$. Within the range of parameters we investigated, this behavior appears to be
generic in the RS-unstable failure phase: Increasing $\rPPc$ systematically reduces the magnitude of RSB. However, due to numerical limitations, we were unable to explore sufficiently large values of $\rPPc$, and thus cannot determine whether a transition from RSB to RS occurs at finite $\rPPc$ or not. 

\Rfig{freeentropy_related} displays the $\rPPc$-dependence of the free entropy $s$, internal energy $u^*$, and complexity $\Sigma$ evaluated at the same parameter as \Rfig{orderparam-ell}. 
\begin{figure}[htbp]
\begin{center}
\includegraphics[width=0.48\columnwidth]{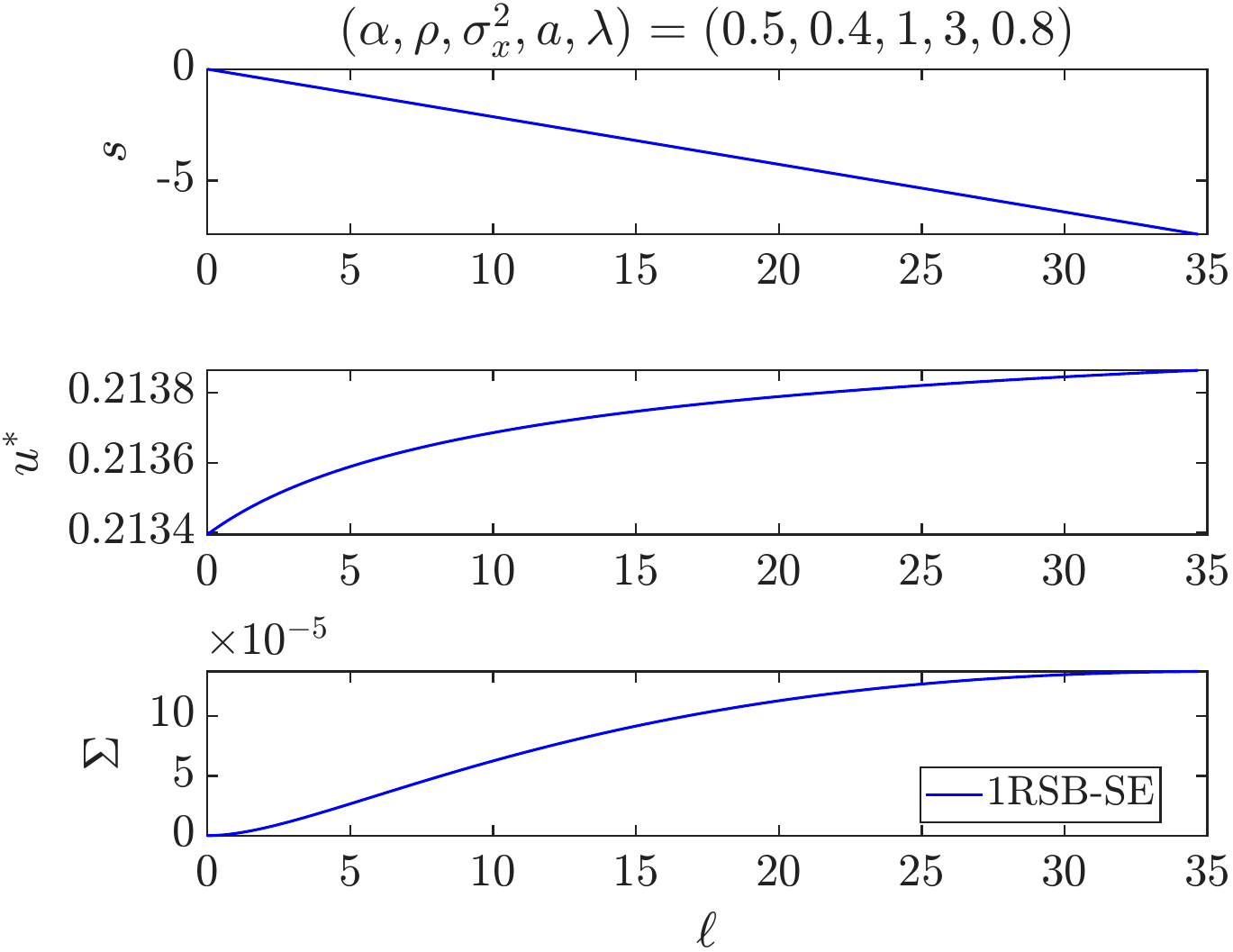}
\includegraphics[width=0.45\columnwidth]{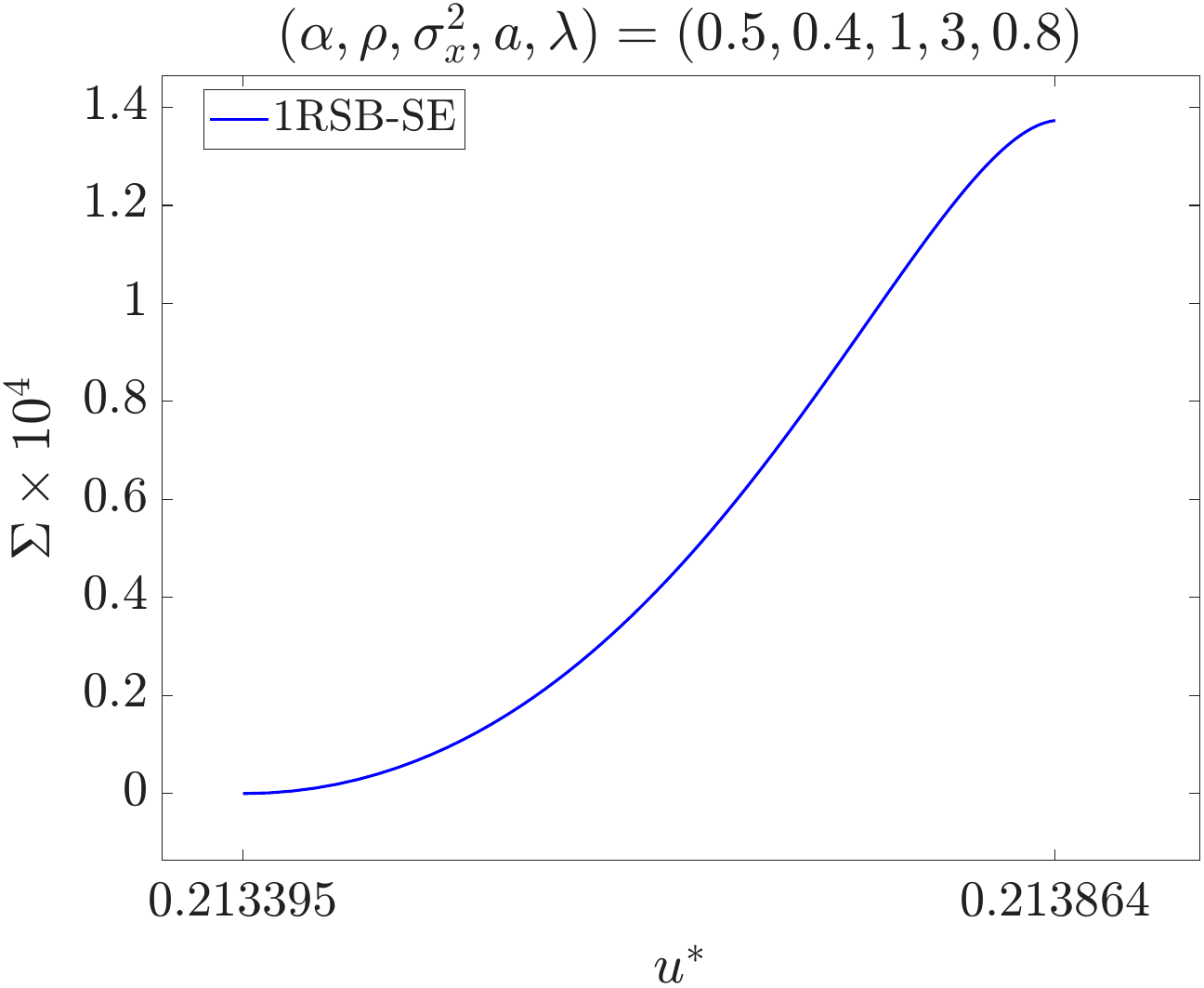}
\includegraphics[width=0.49\columnwidth]{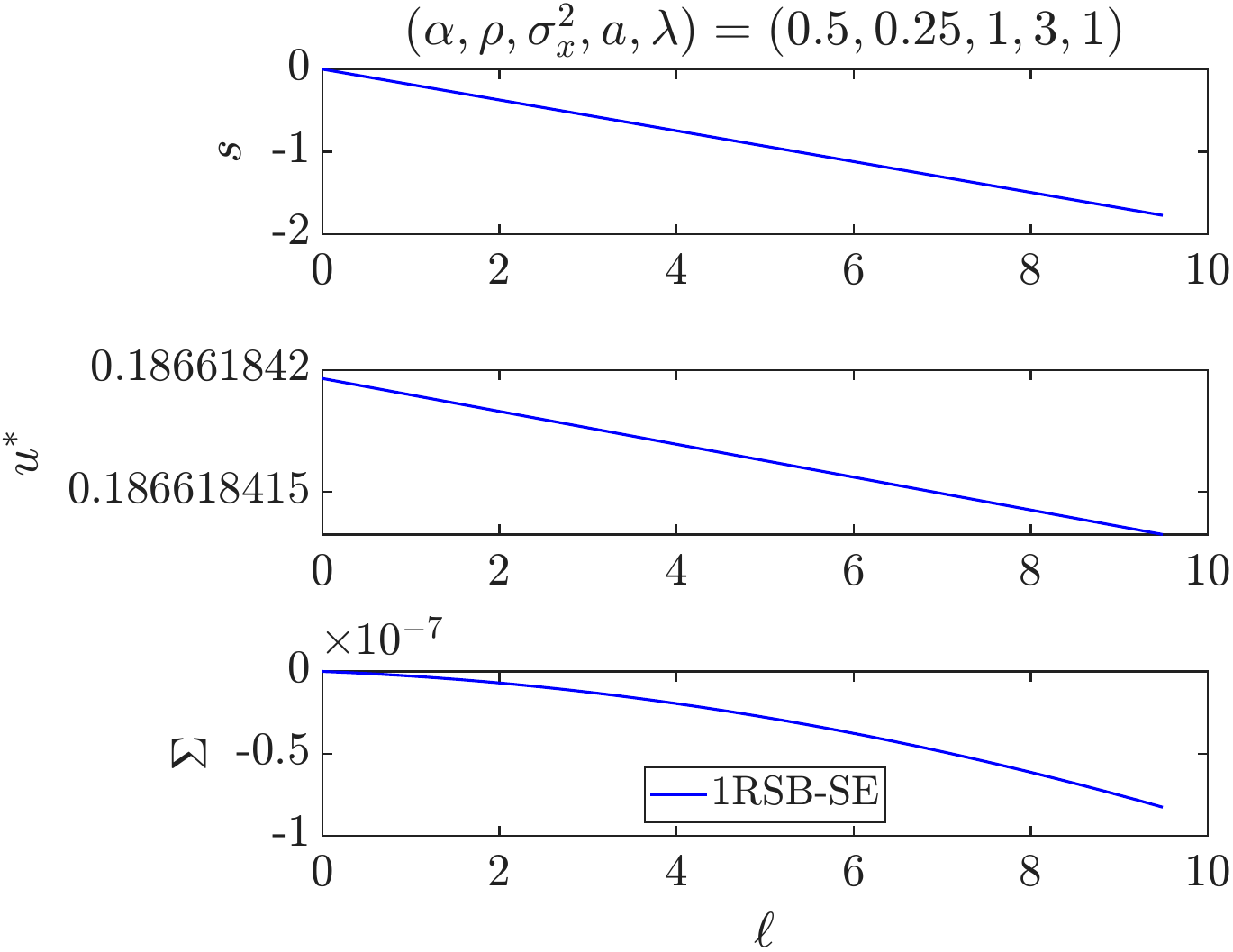}
\includegraphics[width=0.49\columnwidth]{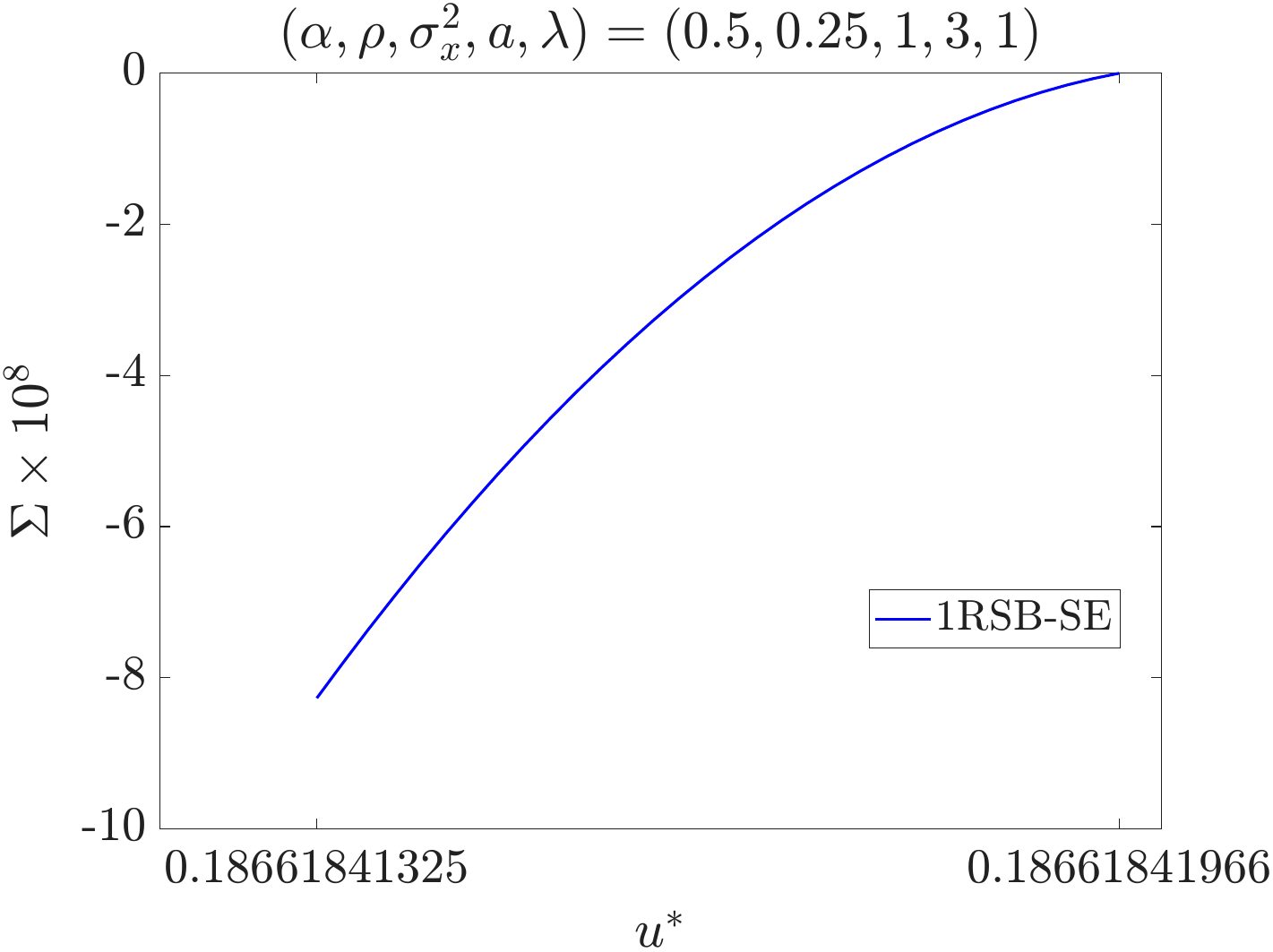}
\vspace{0mm}
\caption{The $\rPPc$-dependence of the free entropy $s(\rPPc)$, energy $u^*(\rPPc)$, and complexity $\Sigma(\rPPc)$ (left) and the energy--complexity curve $\Sigma(u^*)$ (right) at the same parameter as \Rfig{orderparam-ell}. The behavior of $\Sigma(u^*)$ is rather different between the two cases for $(\alpha,\rho ,\sigma_x^2,a,\lambda)=(0.5, 0.4, 1, 3, 0.8)$ (upper) and $(0.5, 0.25, 1, 3, 1)$ (lower).
}
\Lfig{freeentropy_related}
\end{center}
\end{figure}
A delicate point seen in these plots is that $u^*$ and $\Sigma$ vary only slightly over the explored range of $\rPPc$. Taking into account the numerical precision of our current 1RSB-SE solver, it is delicate---especially in the region of large $\rPPc$---to judge whether these behaviors truly reflect thermodynamic properties. 

Furthermore, the qualitative differences in the behavior of the complexity $\Sigma(u^*)$ (right panels) are also concerning. In the upper right panel with $(\alpha,\rho,\sigma_x^2,a,\lambda)=(0.5, 0.4, 1, 3, 0.8)$, it exhibits a nonnegative yet nonconvex behavior, whereas in the lower right one with $(\alpha,\rho,\sigma_x^2,a,\lambda)=(0.5, 0.25, 1, 3, 1)$, it is convex but negative. In general, $\Sigma(u^*)$ should be convex and nonnegative, and therefore neither of these is a thermodynamically legitimate behavior. Whether these behaviors arise because the 1RSB calculation is approximate in the present setting, or whether they are artifacts caused by the numerical precision, cannot be definitively determined, even though we have checked that the results are robust against variations of the parameters controlling the accuracy of the solver. Hence, due to this delicate technical issue in the complexity computation, we avoid using the zero-complexity condition as a criterion for determining the Parisi parameter in this paper~\cite{antenucci2019asp,Barbier2025JSTAT}\footnote{In the present case, $\rPPc=0$ gives the zero complexity, but this trivial condition does not provide any improvement in the reconstruction limit compared to the RS case. Hence, this trivial zero-complexity solution is not suitable. }. Instead, an alternative criterion will be derived below when discussing the phase diagrams.

\Rfig{PD_1RSB} presents the 1RSB version of the phase diagram in the $\lambda$--$\rho$ plane and the reconstruction limit in the $\alpha$--$\rho$ plane, with accompanying some auxiliary figures.
\begin{figure}[htbp]
\begin{center}
\vspace{0mm}
\includegraphics[width=0.45\columnwidth]{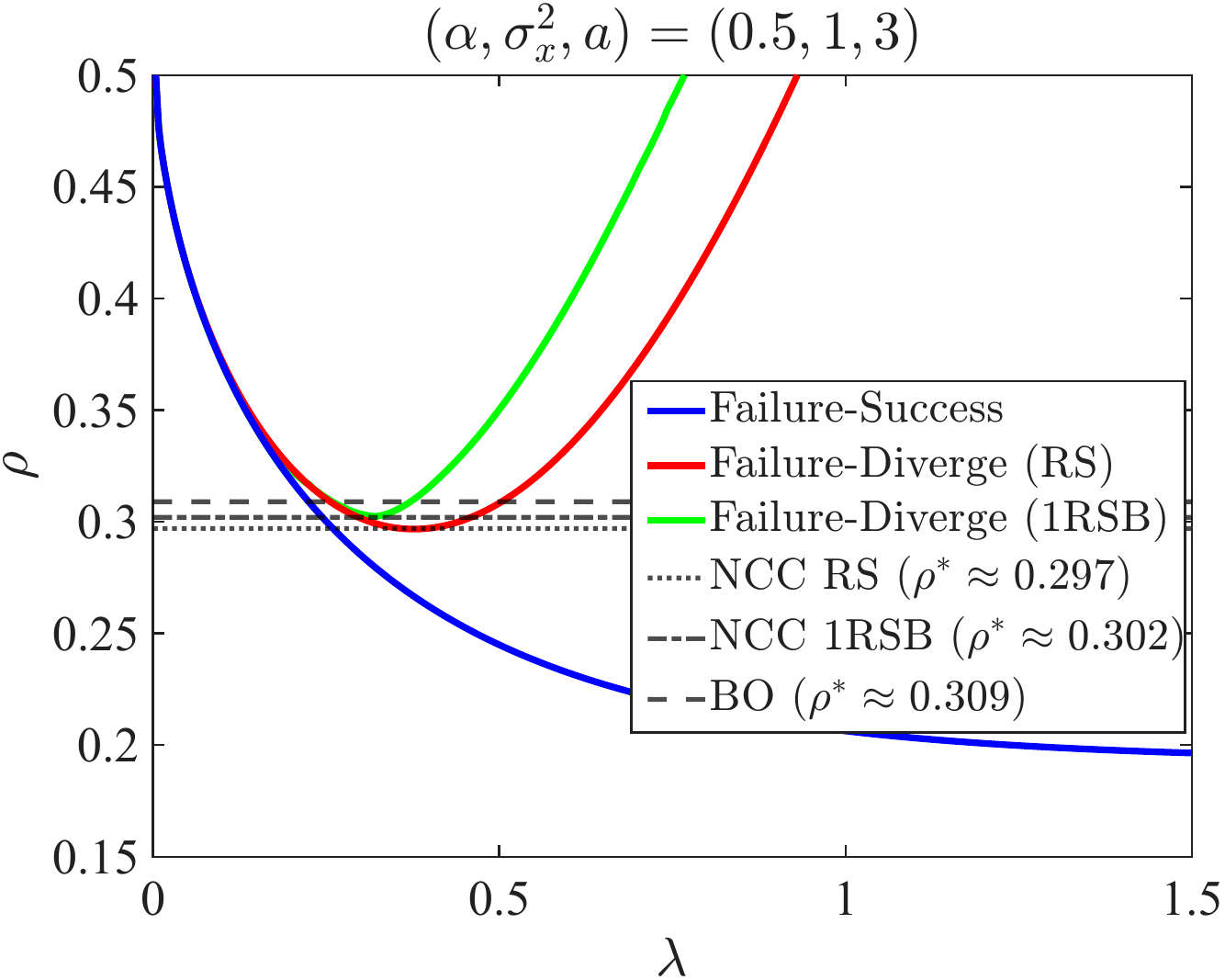}
\includegraphics[width=0.45\columnwidth]{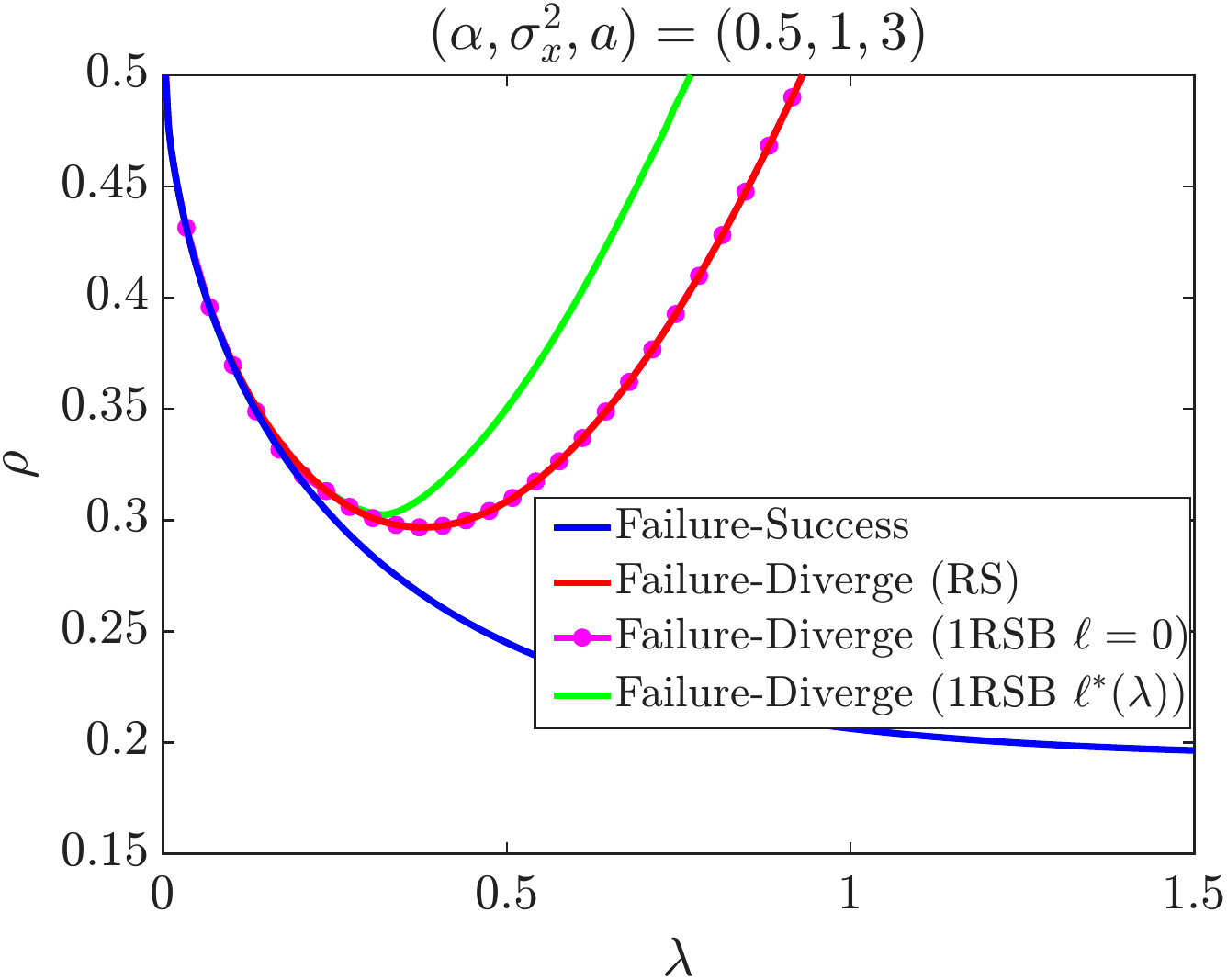}
\includegraphics[width=0.45\columnwidth]{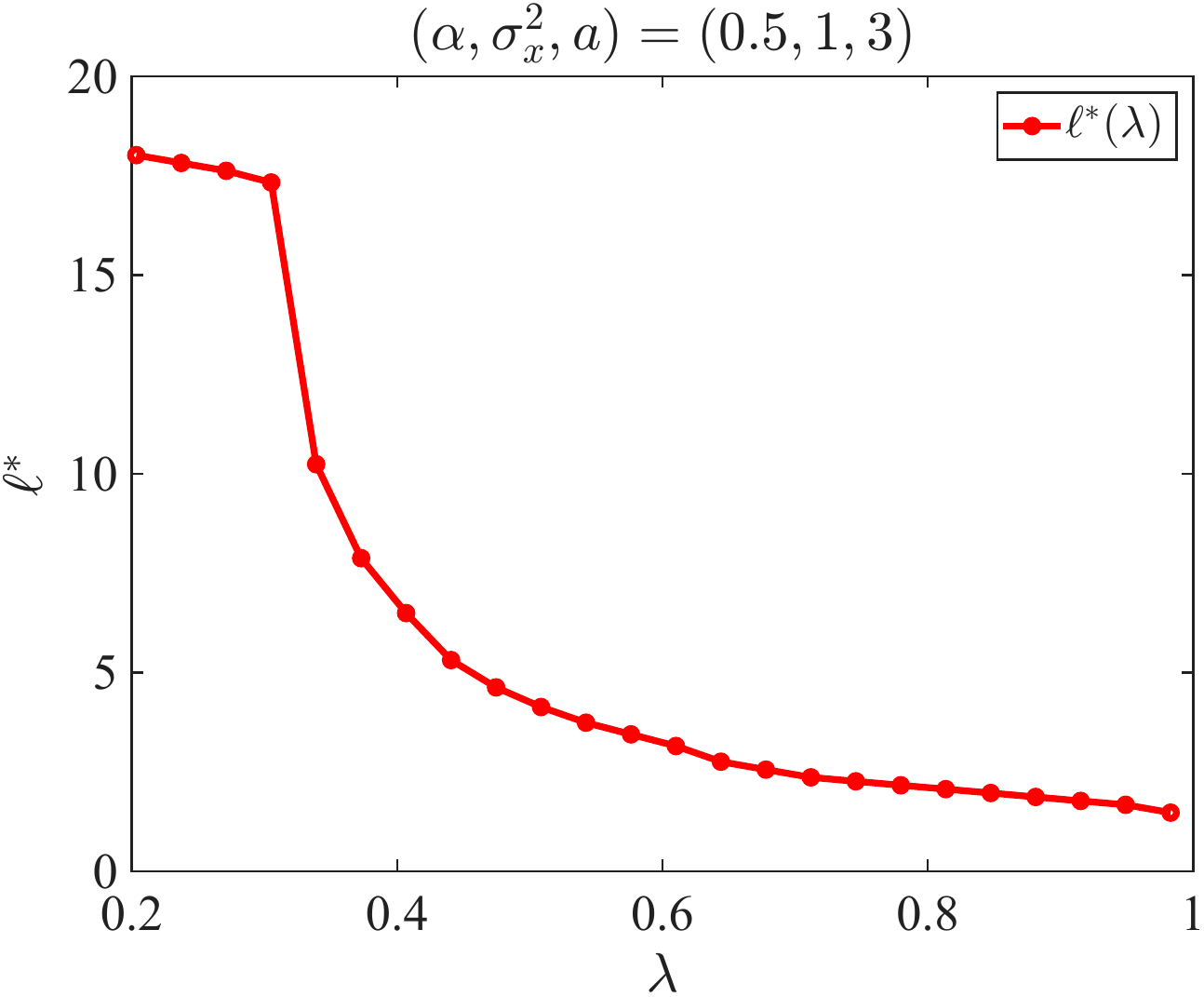}
\includegraphics[width=0.45\columnwidth]{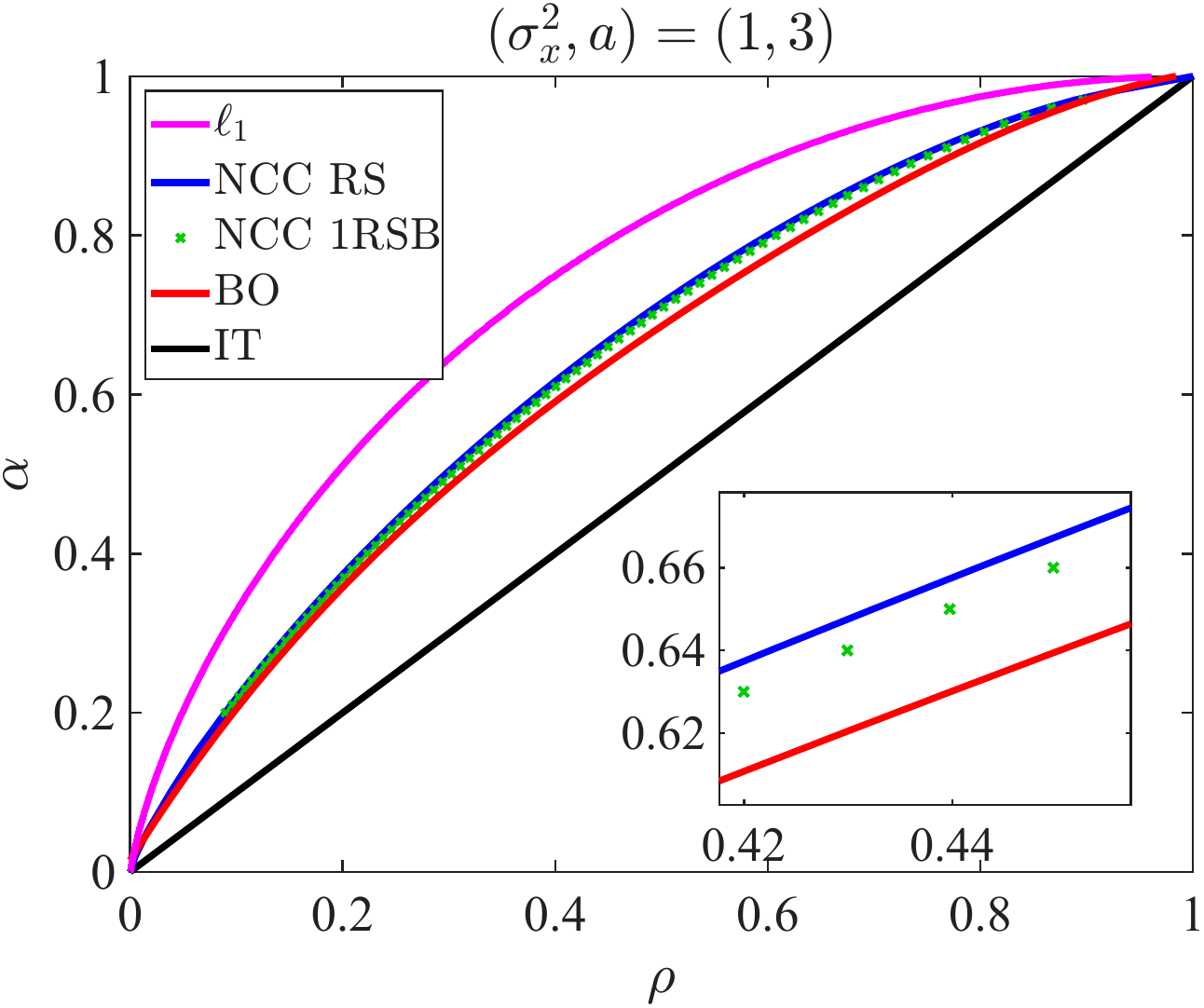}
\vspace{0mm}
\caption{
(Upper left and right) The 1RSB phase diagrams in $\lambda$--$\rho$ plane at $(\alpha,a)=(0.5,3)$, which are the counterpart of the left panel of \Rfig{PD_RS1}. The blue solid line represents the boundary between the success and (RS-unstable) failure phases, while the red and green solid lines depict the boundaries between the failure and diverging phases at the RS and 1RSB levels, respectively. In computing this green boundary, the Parisi parameter $\rPPc$ was optimized so as to minimize the existence region of the diverging phase. The phase boundary when fixing $\rPPc=0$ is depicted in the right panel (magenta line with circles): It almost perfectly coincides with the RS boundary. The NCC limits for the RS and 1RSB cases are displayed as the dotted and dotted-dashed lines ($\rho^*\approx 0.297$ and $0.302$), respectively; for comparison, the BO algorithmic limit is written as the dashed line ($\rho^*\approx 0.309$).     
\\
(Lower left)  The optimized Parisi parameter $\rPPc^*$ to obtain the green phase boundary in the upper panels plotted against $\lambda$. 
\\
(Lower right) The perfect reconstruction limit in the $(\alpha,\rho)$ plane, which is the counterpart of the right panel of \Rfig{PD_RS1}. The NCC RS limit is depicted by the blue solid curve, while the NCC 1RSB
limit is shown by green points. The $\ell_1$ limit (magenta curve), the BO algorithmic limit (red curve) and the information theoretical limit (black straight line), which accords with the BO principle limit, are also shown for reference. The difference among the NCC RS, NCC 1RSB, and BO algorithmic limits is tiny, and the inset giving a magnified view of those three curves is shown to highlight the difference.
}
\Lfig{PD_1RSB}
\end{center}
\end{figure}
The difference from the RS case (\Rfig{PD_RS1}) lies in the location of the boundary between the diverging phase and failure phases. This boundary $\rho_{\rm DF}$ is the function of $\lambda$ and $\rPPc$, and we optimize $\rPPc$ so as to maximize this $\rho_{\rm DF}$. To locate $\rho_{\rm DF}$, we gradually increase $\rho$ while tracking the fixed point of 1RSB-SE and identify the point at which the SE flow diverges. This procedure is carried out with sweeping $\rPPc$ over a certain range ($\lsb 0,20 \rsb \text{in practice})$, and we then select the value of $\rPPc$ that yields the largest $\rho_{\rm DF}$. The green curves in the upper panels of \Rfig{PD_1RSB} are obtained in this way, and the corresponding optimized values of $\rPPc$ are plotted against $\lambda$ in the lower left panel. Compared to the RS boundary (red curve), the diverging phase clearly shrinks, implying that the NCC limit when decreasing $\lambda$ is improved by 1RSB. In the specific choice of the parameters ($(\alpha,a)=(0.5,3)$) in the upper panels, the NCC 1RSB limit is given as $\rho^* \approx 0.302$, which is better than the NCC RS limit ($\rho^*\approx 0.297$). By changing $\alpha$, the overall NCC limit can be plotted in the $\alpha$--$\rho$ plane, which is given as the lower right panel. As anticipated, we observe that the NCC limit is improved by adopting the 1RSB scheme in the overall parameter region. However, the improvement is unfortunately modest, and the resulting limit does not surpass the BO algorithmic limit. This constitutes one of the main outcomes of the present paper.

The above results are for $a = 3$, but we have confirmed that varying $a$ has only a very minor effect on the outcome, as long as $a$ remains within a reasonable range (in practice, we tested $a = 4,6,8$). Therefore, the reconstruction limits shown in \Rfig{PD_1RSB} can be considered robust.

\subsubsection{1RSB-AMP with NCC}
Finally, we examine if the above 1RSB-SE prediction about the reconstruction limit can be realized by 1RSB-AMP combined with the NCC protocol. The actual protocol adapted here is as follows:
\begin{enumerate}
\item{Set the initial condition $m_i^{(0)}=0.0,Q_i^{(0)}=0.3,\chi_{i}^{(0)}=0.3,R^{(-1)}=0.0$ with a sufficiently large $\lambda$ (typically we set $\lambda=1.5$, and the Parisi parameter $\rPPc$ is set to the optimized value $\rPPc^*(\lambda)$ computed from the 1RSB-SE (see the lower left panel of \Rfig{PD_1RSB})).}
\item{Run the 1RSB-AMP algorithm with monitoring $\chi^{(t)},Q^{(t)},q_{0}^{(t)}$ and $D^{(t)}=\|\V{m}^{(t+1)}-\V{m}^{(t)}\|_2/N$.}
\item{If those quantities tend to become stationary, decrease $\lambda$ by $\Delta \lambda=0.1$.  The Parisi parameter $\rPPc$ is also reset to the corresponding $\rPPc^*(\lambda)$ accordingly.  } 
\item{
Recur 2.--3. until converging to the prefect reconstruction solution or diverging.
}   
\end{enumerate}

\Rfig{lambda_anneal} shows the result based on the above protocol at $(\alpha,N,a)=(0.5,3000,3)$. 
\begin{figure}[htbp]
\begin{center}
\vspace{0mm}
\includegraphics[width=0.45\columnwidth]{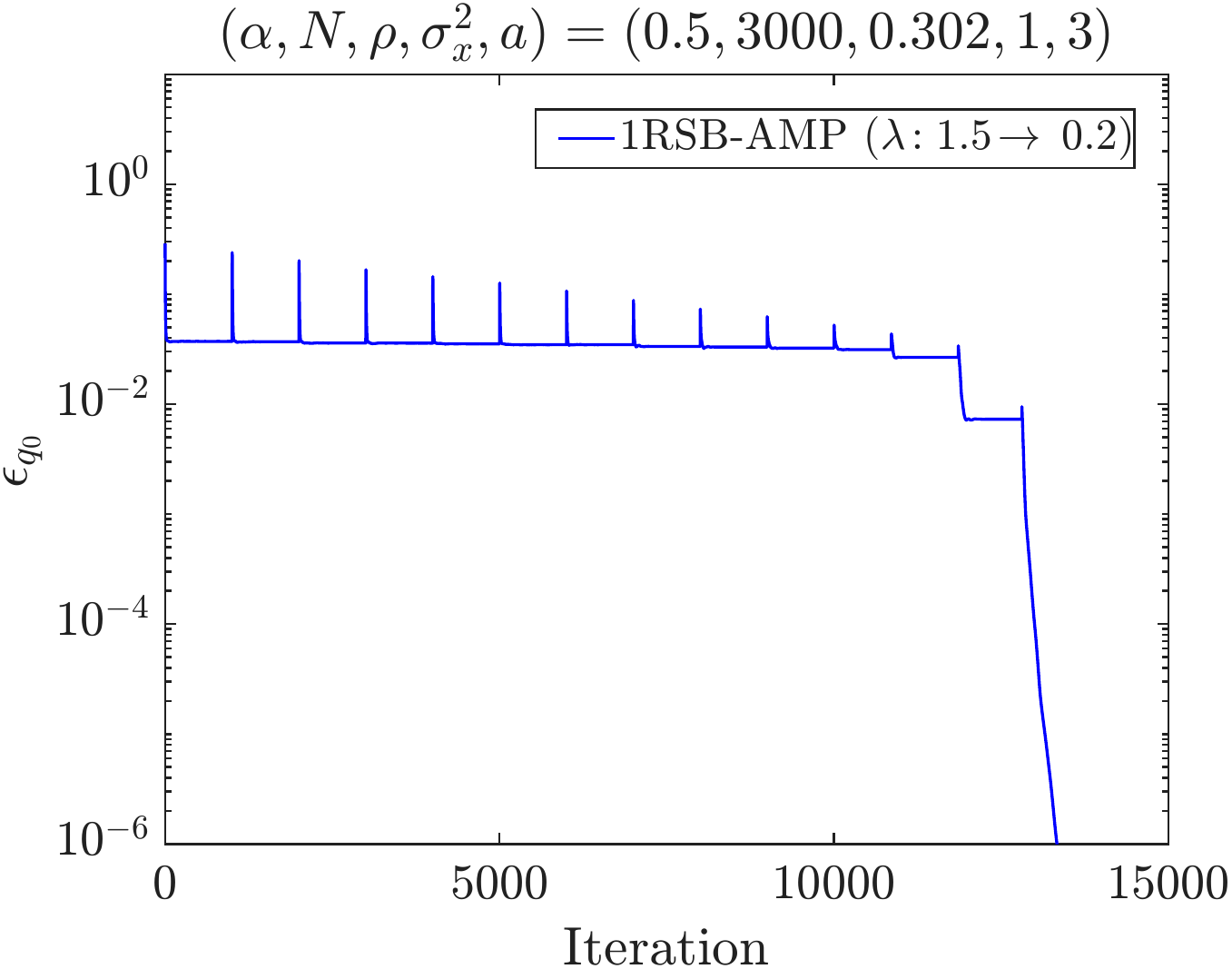}
\includegraphics[width=0.45\columnwidth]{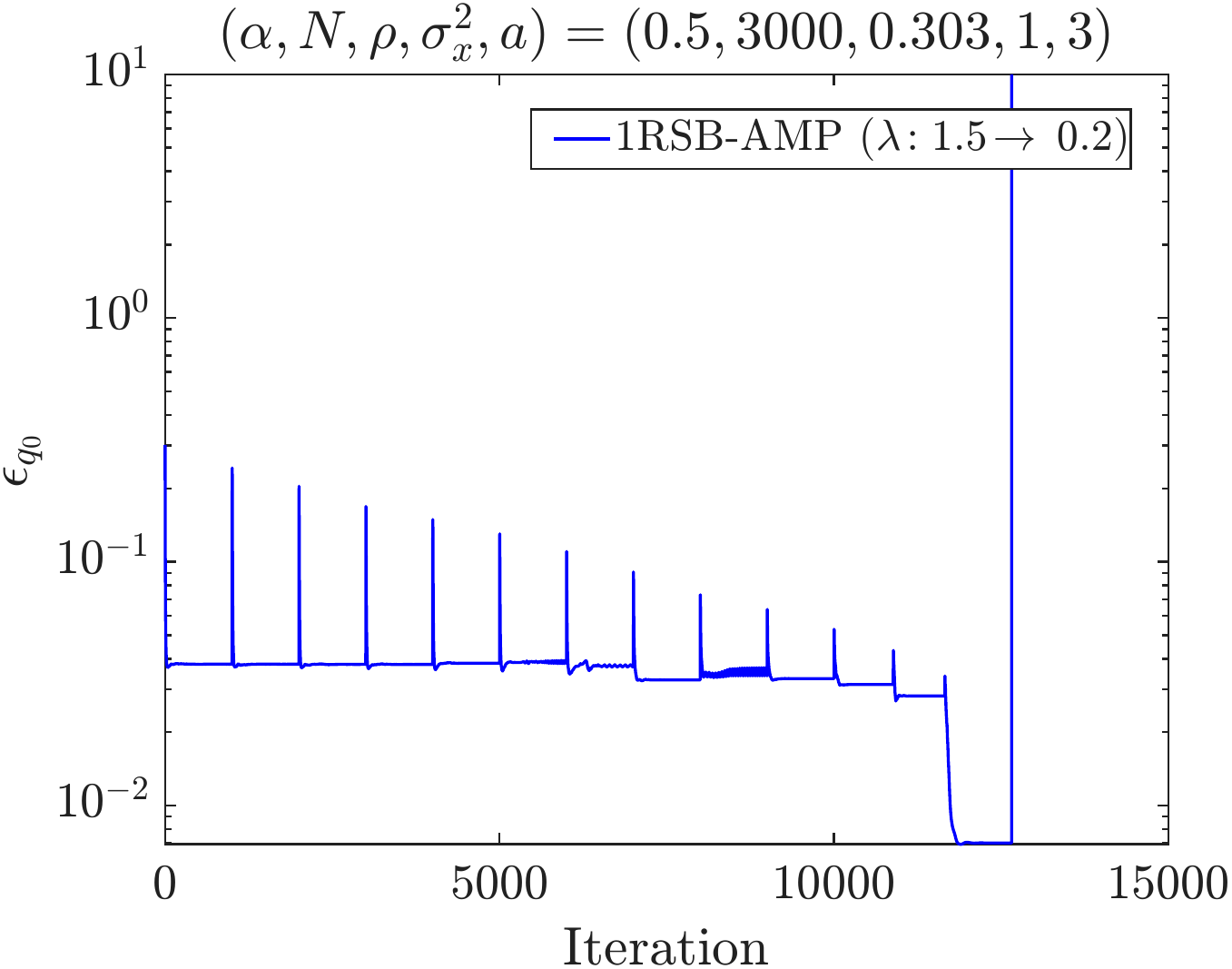}
\includegraphics[width=0.45\columnwidth]{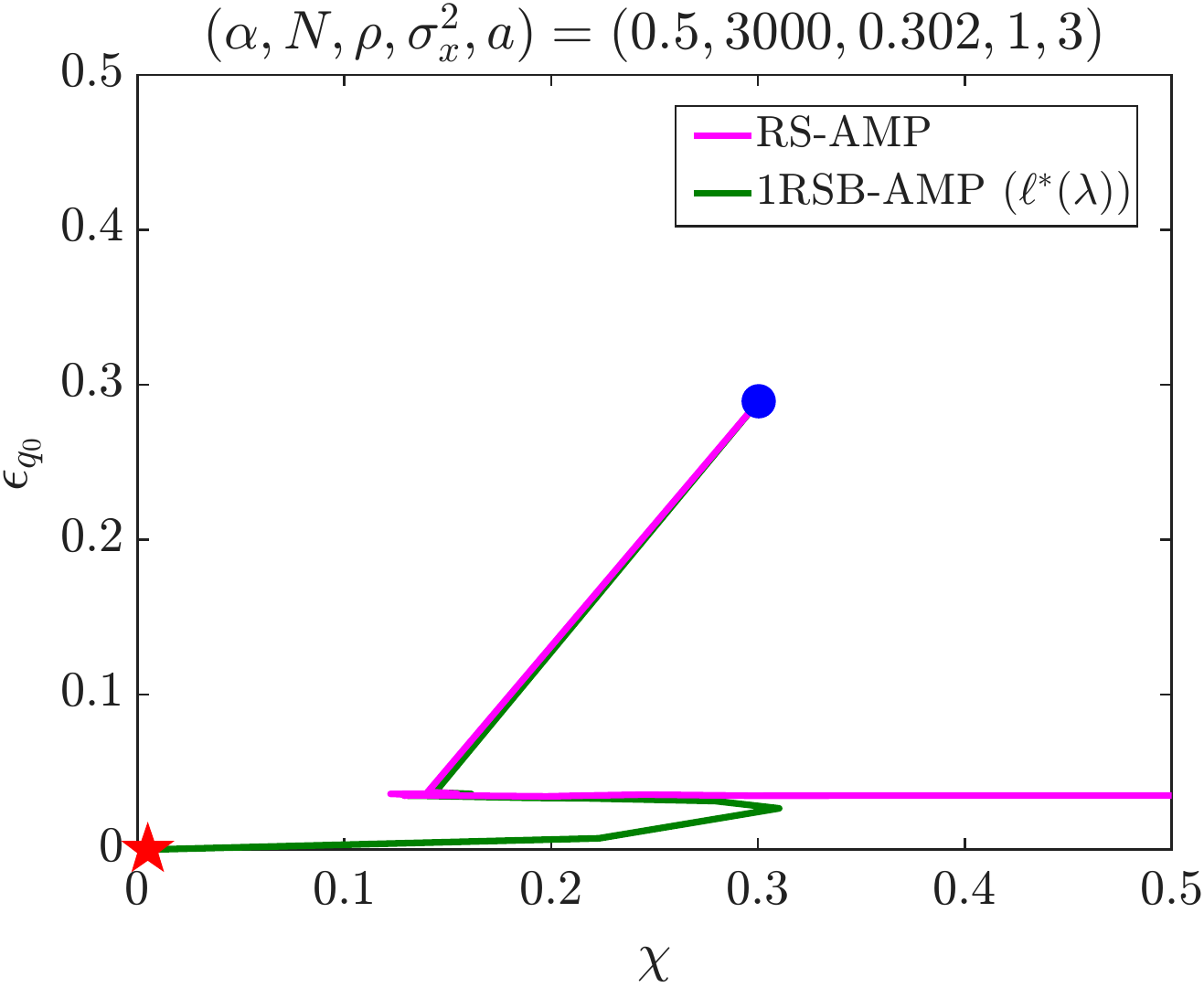}
\includegraphics[width=0.45\columnwidth]{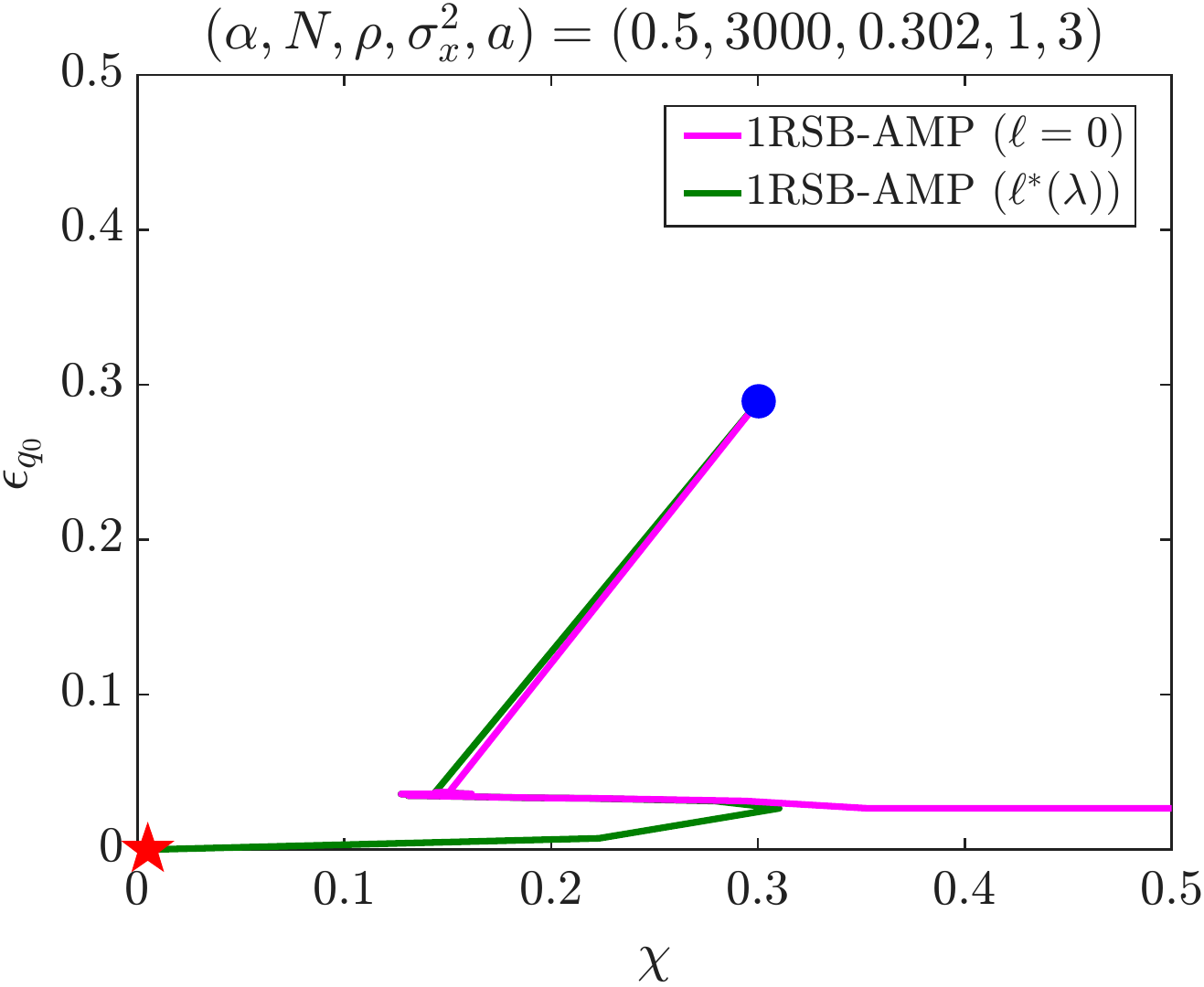}
\vspace{0mm}
\caption{
The 1RSB-AMP dynamics with the NCC protocol for specific samples. 
\\
(Upper left and right) The evolution of MSE against the iteration at $\rho=0.302$ (left) and $\rho=0.303$ (right), which are below and above the NCC 1RSB limit; the other parameters are $(\alpha,N,a)=(0.5,3000,3)$. For the former case, the algorithm achieves the perfect reconstruction while for the other case it diverges, aligning with the 1RSB-SE prediction. Spikes in MSE are observed at the iteration steps where $\lambda$ is varied.
\\ 
(Lower left and right) Comparison of the trajectories of RS-AMP, 1RSB-AMP with the optimized Parisi parameter $\rPPc^*$, and 1RSB-AMP with $\rPPc=0$ for the identical realization of $(\V{A},\V{x}_0)$ under the same nonconvexity protocol. The parameters are $\rho=0.302$ and $(\alpha,N,a)=(0.5,3000,3)$, which is in the divergence phase for RS but is below the NCC limit for 1RSB. The left panel is for RS-AMP and 1RSB-AMP with $\rPPc^*$ and the right one is of 1RSB-AMP with $\rPPc^*$ and $\rPPc=0$. While the optimized 1RSB-AMP achieves the perfect reconstruction, the other two methods eventually diverge.
}
\Lfig{lambda_anneal}
\end{center}
\end{figure}
The upper panels show the evolution of MSE $\epsilon_{q_0}$ against the iteration at $\rho=0.302$ (left) and $\rho=0.303$ (right): For the former case, the algorithm achieves the perfect reconstruction while for the other case it diverges, demonstrating that the 1RSB-SE prediction is tight. The lower panels compare the trajectories of RS-AMP, 1RSB-AMP with the optimized Parisi parameter $\rPPc^*$, and 1RSB-AMP with $\rPPc=0$ for the same realization of $(\V{A},\V{x}_0)$ under the same nonconvexity protocol. While the optimized 1RSB-AMP achieves the perfect reconstruction, the other two methods eventually diverge. Interestingly, these trajectories look quite similar up to a certain point, but beyond that point their behaviors clearly change: One converges to perfect reconstruction and the others tend to rapidly diverge. Non-monotonic trajectories are also observed in these panels, and we find it is a common phenomenon as far as we experimentally examined. 

To further examine the finite-size behavior and sample-to-sample robustness of the NCC protocol, we also compute the success probabilities by performing repeated experiments over independent realizations while varying the system size $N$. The resulting success probabilities are summarized in \Rfig{SP_1RSB_RS_AMP}. For RS-AMP, once $\rho$ exceeds the vertical dashed line describing the RS-NCC limit, the success probability tends to decrease as $N$ increases. In contrast, for 1RSB-AMP, the success probability increases with $N$ beyond the RS-NCC limit and appears to remain enhanced up to the 1RSB-NCC limit indicated by the vertical solid line. This finite-size trend supports the accuracy of the SE prediction and also suggests that the NCC protocol is robust against sample-to-sample fluctuations.
\begin{figure}[H]
\begin{center}
\vspace{0mm}
\includegraphics[width=0.5\columnwidth]{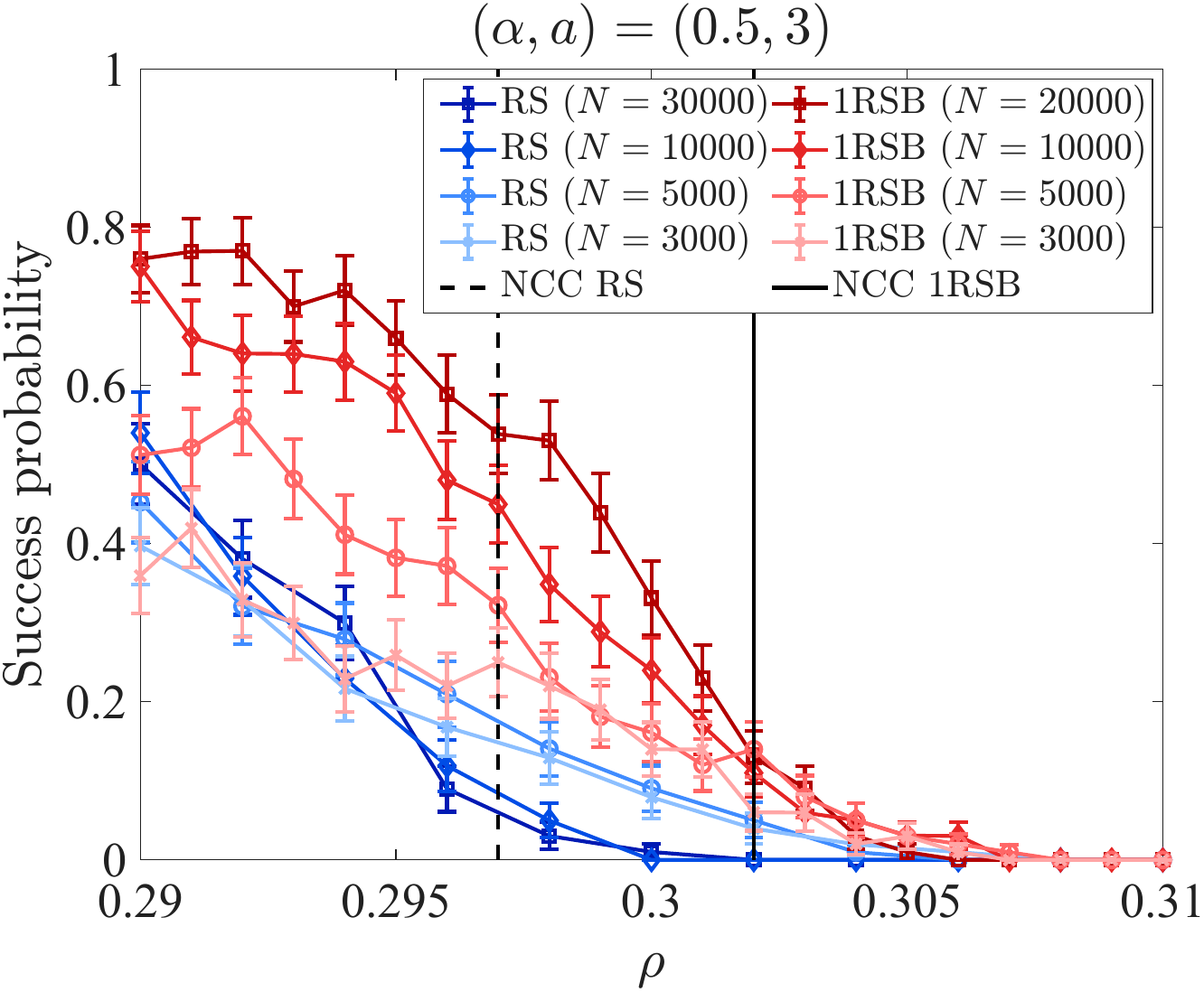}
\vspace{0mm}
\caption{
Finite-size comparison of the success probabilities of RS-AMP and 1RSB-AMP under the NCC protocol at $(\alpha,a)=(0.5,3)$. For each value of $\rho$ and $N$, the success probability is estimated over 100 independent realizations of $(\V{A},\V{x}_0)$, and the bars describe the standard deviation estimated using the Bootstrap method. Blue and red curves represent RS-AMP and 1RSB-AMP, respectively, and different shades/markers correspond to different system sizes. The vertical dashed and solid lines indicate the NCC limits predicted by RS-SE and 1RSB-SE, respectively. }
\Lfig{SP_1RSB_RS_AMP}
\end{center}
\end{figure}

These results clearly demonstrate that the predictions of 1RSB-SE are indeed realized by 1RSB-AMP.

\section{Conclusion} 
In this work, we have developed and analyzed 1RSB-AMP and 1RSB-SE for CS based on the SCAD minimization. Starting from the 1RSB formulation of BP, we derived explicit update rules of 1RSB-AMP and the corresponding 1RSB-SE equations. A detailed comparison between 1RSB-AMP and 1RSB-SE showed that they are in excellent agreement at the macroscopic level, even in a part of the phase where RS-AMP diverges. This confirms that the 1RSB extension provides a more stable and reliable algorithm.

Through the detailed analysis of the 1RSB-SE fixed points, the phase diagrams and the perfect reconstruction limit were also derived. The Parisi parameter was chosen to minimize the existence region of the diverging phase, which is a new criterion of optimizing the Parisi parameter and fits well with the properties of the current problem. As a result, the NCC limit is improved from the RS one, though the improvement is not large and the resultant NCC 1RSB limit is still slightly inferior to the BO algorithmic limit. These 1RSB-SE predictions were confirmed through the NCC experiments using 1RSB-AMP, and the result clearly demonstrated the tightness of the 1RSB-SE predictions. 

As discussed, the 1RSB description is not expected to be thermodynamically exact in the failure phase of the present problem. Nevertheless, the fact that 1RSB-AMP and 1RSB-SE remain macroscopically accurate even in regions where the 1RSB formulation itself is not exact suggests 
that thermodynamic exactness is not a necessary requirement from the viewpoint of algorithm design. Clarifying this perspective through alternative macroscopic formulations may lead to improved nonconvex optimization algorithms as well as a deeper understanding of related high-dimensional inference problems.

Finally, although the present work is limited to the 1RSB description, it is natural to ask whether the framework can be extended toward algorithms inspired by higher-step RSB. In this direction, recent developments such as $K$-step ASP~\cite{Chen2024KstepVASP} and the incremental AMP algorithm for optimizing the Sherrington--Kirkpatrick Hamiltonian~\cite{Montanari2021OptimizationSK} suggest that suitably designed message-passing dynamics, guided by RSB-type macroscopic functionals, can effectively track complex landscapes beyond the 1RSB regime. Extending and applying such constructions to high-dimensional inference problems with nonconvex regularization is an interesting direction for future research.

\section*{Acknowledgements} 
This work was supported by JST SPRING, Grant Number JPMJSP2110 (XG), JST PRESTO, Grant Number JPMJPR23J4 (AS), JST CREST, Grant Number JPMJCF1862 (TO), JSPS KAKENHI, Grant Numbers JP22K12179 (TO) and JP22H05117 (AS,TO).

\appendix

\section{Message passing at the 1RSB level}\Lsec{Message}

\subsection{1RSB-BP}\Lsec{1RSB-BP}
To derive the update rule from $\{\ul{\phi}_{i \to \mu}\}$ to $\{\ul{\T{\phi}}_{\mu \to i}\}$, let us define the following quantities and notations:
\be
&&
\Ave{(\cdots)}_{\ul{\phi}_{i \to \mu}}
\coloneqq
\int d\ul{x}_i\ul{\phi}_{i \to \mu}(\ul{x}_i)(\cdots),
\\ &&
\Cov_{\ul{\phi}_{i \to \mu}}(f,g)
\coloneqq
\Ave{fg}_{\ul{\phi}_{i \to \mu}}-\Ave{f}_{\ul{\phi}_{i \to \mu}}\Ave{g}_{\ul{\phi}_{i \to \mu}},
\\ &&
\ul{\mc{Z}}(\Gamma,B,C)\coloneqq
\int Dz
\lb
\int dx
e^{
\beta \lb
-\frac{1}{2}\Gamma x^2 
+ \lb \sqrt{B}z_1+ C \rb x 
- J(x)
\rb
}
\rb^{\PPc},
\\ &&
\ul{\mc{Z}}_{i\to \mu} \coloneqq \ul{\mc{Z}}(\Gamma_{i\to \mu},B_{i\to \mu},C_{i\to \mu}),
\\ &&
m_{i \to \mu}\coloneqq \Ave{x_{ai}}_{\ul{\phi}_{i \to \mu}}
=
\frac{1}{\beta \PPc} \partial_{C_{i\to \mu}} \log \ul{\mc{Z}}_{i\to \mu} ,
\\ &&
Q_{i \to \mu}\coloneqq \Ave{x^2_{ai}}_{\ul{\phi}_{i \to \mu}}
=
-\frac{2}{\beta \PPc}  \partial_{\Gamma_{i\to \mu}} \log \ul{\mc{Z}}_{i\to \mu} ,
\\ &&
q_{i \to \mu}\coloneqq \Ave{x_{ai}x_{bi}}_{\ul{\phi}_{i \to \mu}}
=
\frac{1}{\PPc(\PPc-1)}\lbb \frac{2}{\beta^2 }  \partial_{B_{i\to \mu}} \log \ul{\mc{Z}}_{i\to \mu} +\frac{2}{\beta}\partial_{\Gamma_{i\to \mu}} \log \ul{\mc{Z}}_{i\to \mu} 
\rbb
~(a\neq b).
\ee
Furthermore, we introduce additional average notations describing the average over the cavity system:
\be
&&
\Ave{(\cdots)}_{\prod_{j(\neq i)}\ul{\phi}_{j \to \mu}}
\coloneqq
\int \lb \prod_{j(\neq i)} d\ul{x}_j\ul{\phi}_{j \to \mu}(\ul{x}_j) \rb (\cdots),
\\ &&
\Cov_{ \prod_{j(\neq i)}\ul{\phi}_{j \to \mu} }(f,g)
\coloneqq
\Ave{fg}_{\prod_{j(\neq i)}\ul{\phi}_{j \to \mu}}
-
\Ave{f}_{\prod_{j(\neq i)}\ul{\phi}_{j \to \mu}}
\Ave{g}_{\prod_{j(\neq i)}\ul{\phi}_{j \to \mu}}.
\ee
Given $\{\ul{\phi}_{j\to \mu}\}_{j(\neq i)}$,  we can compute $\ul{\T{\phi}}_{\mu \to i}$ from \Req{BP1-replicated}. The CLT implies $\sum_{j(\neq i)}A_{\mu j}x_{aj}$ becomes a Gaussian random variable: The mean and variance are
\be
&&
\Ave{\sum_{j(\neq i)}A_{\mu j}x_{aj}}_{\prod_{j(\neq i)}\ul{\phi}_{j \to \mu}}
=\sum_{j(\neq i)}A_{\mu j}m_{j\to \mu},
\\ &&
\Cov_{\prod_{j(\neq i)}\ul{\phi}_{j \to \mu}}
\lb {\sum_{j(\neq i)}A_{\mu j}x_{aj},\sum_{j(\neq i)}A_{\mu j}x_{bj}} \rb
=\sum_{j(\neq i)}\sum_{k(\neq i)}A_{\mu j}A_{\mu k}
\Cov_{\prod_{j(\neq i)}\ul{\phi}_{j \to \mu}}(x_{aj},x_{bk})
\no \\ &&
=\sum_{j(\neq i)}A_{\mu j}^2\lb Q_{j\to \mu}\delta_{ab}+ q_{j\to \mu}(1-\delta_{ab})-m_{j\to \mu}^2 \rb.
\ee
Let us denote
\be
&&
Q_{\mu \to i} \coloneqq \sum_{j(\neq i)}A_{\mu j}^2 Q_{j\to \mu},
\quad
q_{1,\mu \to i} \coloneqq \sum_{j(\neq i)}A_{\mu j}^2 q_{j\to \mu},
\quad
q_{0,\mu \to i} \coloneqq \sum_{j(\neq i)}A_{\mu j}^2 m^2_{j\to \mu}.
\ee
Hence, a nice representation of $\sum_{j(\neq i)}A_{\mu j}x_{aj}$ using independent zero-mean unit-variance normal variables $\{z_0,(z_{1a})_{a=1}^{\PPc}\}$ is given by
\be
\sum_{j(\neq i)}A_{\mu j}x_{aj} \overset{\textrm{d}}{=} \sum_{j(\neq i)}A_{\mu j} m_{j\to \mu}+\sqrt{Q_{\mu \to i}-q_{1,\mu \to i}}z_{1a}+\sqrt{q_{1,\mu \to i}-q_{0,\mu \to i}}z_0,
\ee
where $\overset{\textrm{d}}{=}$ denotes the equality in distribution. Using this representation,  \Req{BP1-replicated} becomes
\be
&&
\ul{\T{\phi}}_{\mu \to i}(\ul{x}_i)
\propto
\int d\ul{\V{x}}^{\bs i} \ul{\Phi}_{\mu}(\ul{\V{x}})\prod_{j (\neq i)}\ul{\phi}_{j \to \mu}(\ul{x}_j)
\no \\ &&
\CLTap
\int Dz_0 \prod_{a=1}^{\PPc }
\int Dz_{1a} 
\frac{1}{\sqrt{2\pi \tau}}
e^{-\frac{1}{2\tau}\lb y_{\mu}-
\lb 
\sum_{j(\neq i)}A_{\mu j} m_{j\to \mu}
+
\sqrt{Q_{\mu \to i}-q_{1,\mu \to i}}z_{1a}+\sqrt{q_{1,\mu \to i}-q_{0,\mu \to i}}z_0
\rb
-A_{\mu i }x_{ai} \rb^2}
\no \\ &&
=
\int Dz_0 
\prod_{a=1}^{\PPc }
\sqrt{\frac{1}{2\pi (\tau+Q_{\mu \to i}-q_{1,\mu \to i})}}
e^{-\frac{1}{2}\frac{A_{\mu i}^2}{\tau+Q_{\mu \to i}-q_{1,\mu \to i}}\lb x_{ai}-\frac{R_{\mu i}}{A_{\mu i}}\rb^2}
\no \\ &&
\overset{\tau \to 0}{\propto}
e^{
-\frac{1}{2}\beta \T{\Gamma}_{\mu \to i}\sum_{a=1}^{\PPc}x_{ai}^2
+\frac{1}{2}\beta^2 \T{B}_{\mu \to i} \lb \sum_{a=1}^{\PPc}x_{ai}\rb^2
+\beta \T{C}_{\mu \to i}  \sum_{a=1}^{\PPc}x_{ai}
},
\ee
where
\subbe
\be
&&
R_{\mu i}=y_{\mu}-\sum_{j(\neq i)}A_{\mu j} m_{j\to \mu}-\sqrt{q_{1,\mu \to i}-q_{0,\mu \to i}}z_0,
\\ &&
\chi_{\mu \to i}=\beta \lb Q_{\mu \to i} - q_{1,\mu \to i}\rb,
\\ &&
\T{\Gamma}_{\mu \to i}=\frac{A_{\mu i}^2}{\chi_{\mu \to i}},
\\ &&
\T{B}_{\mu \to i}=
\frac{A_{\mu i}^2}{\chi_{\mu \to i}}
\frac{(q_{1,\mu \to i}-q_{0,\mu \to i})}{\chi_{\mu \to i}+ \beta \PPc (q_{1,\mu \to i}-q_{0,\mu \to i})},
\\ &&
\T{C}_{\mu \to i}=
\frac{A_{\mu i}(y_{\mu}-\sum_{j(\neq i)}A_{\mu j} m_{j\to \mu})}{\chi_{\mu \to i}+ \beta \PPc (q_{1,\mu \to i}-q_{0,\mu \to i})}.
\ee
\subee
This in conjunction with \Req{1RSBBP_tilde2main_parametrized} provides the set of closed-form equations. 

\paragraph{Summary of 1RSB-BP}
\subbe
\be
&&
\chi^{(t)}_{\mu \to i}=\sum_{j(\neq i)}A_{\mu j}^2\chi^{(t)}_{j \to \mu},
\\ &&
q^{(t)}_{1,\mu \to i}=\sum_{j(\neq i)}A_{\mu j}^2 q_{j \to \mu}^{(t)},
\\ &&
q^{(t)}_{0,\mu \to i}=\sum_{j(\neq i)}A_{\mu j}^2 (m_{j \to \mu}^{(t)})^2,
\\ &&
\T{\Gamma}_{\mu \to i}^{(t)}=\frac{A_{\mu i}^{2}}{\chi_{\mu \to i}^{(t)}},
\\ &&
\T{B}_{\mu \to i}^{(t)}=
\frac{A_{\mu i}^2}{\chi^{(t)}_{\mu \to i}}
\frac{(q^{(t)}_{1,\mu \to i}-q^{(t)}_{0,\mu \to i})}{\chi^{(t)}_{\mu \to i}+ \beta \PPc (q^{(t)}_{1,\mu \to i}-q^{(t)}_{0,\mu \to i})},
\\ &&
\T{C}^{(t)}_{\mu \to i}=
\frac{A_{\mu i}(y_{\mu}-\sum_{j(\neq i)}A_{\mu j}m_{j \to \mu}^{(t)})}{\chi^{(t)}_{\mu \to i}+ \beta \PPc (q^{(t)}_{1,\mu \to i}-q^{(t)}_{0,\mu \to i})},
\\ &&
\Gamma_{i \to \mu}^{(t+1)}=\sum_{\nu (\neq \mu)}\T{\Gamma}_{\nu \to i}^{(t)},
\quad
B_{i \to \mu}^{(t+1)}=\sum_{\nu (\neq \mu)}\T{B}_{\nu \to i}^{(t)},
\quad
C_{i \to \mu}^{(t+1)}=\sum_{\nu (\neq \mu)}\T{C}_{\nu \to i}^{(t)},
\\ &&
\ul{\mc{Z}}_{i \to \mu}^{(t+1)}=\ul{\mc{Z}}\lb \Gamma_{i \to \mu}^{(t+1)}, B_{i \to \mu}^{(t+1)}, C_{i \to \mu}^{(t+1)}\rb,
\\ &&
m^{(t+1)}_{i \to \mu}=\Ave{x_{ai}}_{\ul{\phi}^{(t+1)}_{i \to \mu}}
=\frac{1}{\beta \PPc} \partial_{C_{i\to \mu}^{(t+1)}} \log \ul{\mc{Z}}^{(t+1)}_{i\to \mu} ,
\\ &&
Q^{(t+1)}_{i \to \mu}=\Ave{x^2_{ai}}_{\ul{\phi}^{(t+1)}_{i \to \mu}}
=
-\frac{2}{\beta \PPc}  \partial_{\Gamma_{i\to \mu}^{(t+1)}} \log \ul{\mc{Z}}^{(t+1)}_{i\to \mu} ,
\\ &&
q^{(t+1)}_{i \to \mu}=\Ave{x_{ai}x_{bi}}_{\ul{\phi}^{(t+1)}_{i \to \mu}}
=
\frac{1}{\PPc(\PPc-1)}
\lbb 
\frac{2}{\beta^2 }  \partial_{B_{i\to \mu}^{(t+1)}} \log \ul{\mc{Z}}^{(t+1)}_{i\to \mu} +\frac{2}{\beta}\partial_{\Gamma_{i\to \mu}^{(t+1)}} \log \ul{\mc{Z}}^{(t+1)}_{i\to \mu}
\rbb,
\\ &&
\chi_{j\to \mu}^{(t+1)}=\beta \lb Q_{j \to \mu}^{(t+1)}- q_{j\to \mu}^{(t+1)}\rb.
\ee
\subee

\paragraph{Zero-temperature limit}
Let us consider the $\beta \to \infty $ limit. To this end, we define
\be
&&
\Ave{(\cdots)}_{\Gamma,B,C}
\coloneqq
\frac{
\int Dz \lb \int dx e^{\beta g(x,\Gamma,\sqrt{B}z+C)} \rb^{\PPc}
	\frac{
	\int dx (\cdots) e^{\beta g(x,\Gamma,\sqrt{B}z+C)}
	}{
	\int dx e^{\beta g(x,\Gamma,\sqrt{B}z+C)}
	}
}
{
\ul{\mc{Z}}(\Gamma,B,C)
},
\\ &&
\Ave{(\cdots)}_{i\to \mu}^{(t)}
\coloneqq
\Ave{(\cdots)}_{\Gamma_{i\to \mu}^{(t)},B_{i\to \mu}^{(t)},C_{i\to \mu}^{(t)}},
\\ &&
g_{i\to \mu}^{(t)}(x,z)=g(x,\Gamma_{i\to \mu}^{(t)},B_{i\to \mu}^{(t)}z+C_{i\to \mu}^{(t)}),
\ee
From simple algebras, we get
\be
&&
\partial_{C}\log \ul{\mc{Z}}=\beta\PPc \Ave{x}_{\Gamma,B,C},
\\&&
\partial_{\Gamma}\log \ul{\mc{Z}}=-\frac{1}{2}\beta \PPc \Ave{x^2}_{\Gamma,B,C},
\\ &&
\partial_{B}\log \ul{\mc{Z}}=
\frac{1}{2}\beta^2 \PPc(\PPc-1) 
\frac{
\int Dz \lb \int dx e^{\beta g(x,\Gamma,\sqrt{B}z+C)} \rb^{\PPc}
	\lb 
	\frac{
	\int dx x e^{\beta g(x,\Gamma,\sqrt{B}z+C)}
	}{
	\int dx e^{\beta g(x,\Gamma,\sqrt{B}z+C)}
	}
	\rb^2
}
{
\ul{\mc{Z}}(\Gamma,B,C)
}
+
\frac{1}{2}\beta^2 \PPc \beta \Ave{x^2}_{\Gamma,B,C}.
\ee
Hence
\be
&&
m_{i\to \mu}^{(t+1)}=\Ave{x}_{i\to \mu}^{(t+1)},
\\ &&
Q_{i\to \mu}^{(t+1)}=\Ave{x^2}_{i\to \mu}^{(t+1)},
\\ &&
q_{i\to \mu}^{(t+1)}=\frac{
\int Dz 
\lb \int dx e^{\beta g_{i\to \mu}^{(t+1)}(x,z)} \rb^{\PPc}
	\lb 
	\frac{
	\int dx x e^{\beta g_{i\to \mu}^{(t+1)}(x,z)}
	}{
	\int dx e^{\beta g_{i\to \mu}^{(t+1)}(x,z)}
	}
	\rb^2
}
{
\ul{\mc{Z}}_{i\to \mu}^{(t+1)}
}
,
\\ &&
\chi_{i\to \mu}^{(t+1)}
=
\frac{1}{\ul{\mc{Z}}_{i\to \mu}^{(t+1)}}
\int Dz
\lb \int dx e^{\beta g_{i\to \mu}^{(t+1)}(x,z)} \rb^{\PPc}
\no \\ &&
\times
\beta
\lbb 
\frac{
\int dx x^2 e^{\beta g_{i\to \mu}^{(t+1)}\lb x,z\rb}
}{
\int dx e^{\beta g_{i\to \mu}^{(t+1)}\lb x,z \rb}
}
-
\lb 
\frac{
\int dx x e^{\beta g_{i\to \mu}^{(t+1)}\lb x,z \rb}
}{
\int dx e^{\beta g_{i\to \mu}^{(t+1)}\lb x,z \rb}
}
\rb^2
\rbb.
\Leq{chi-1RSB}
\ee
In the $\beta \to \infty$ limit, the saddle-point method can be applied to the integration w.r.t. $x$ in the above equations; in the $\beta \to \infty $ limit keeping $\beta \PPc \to \rPPc=O(1)$, the average safely converges to
\be
&&
\ul{\mc{Z}}(\Gamma,B,C)
\to
\int Dz e^{\rPPc \hat{g}(z,\Gamma,\sqrt{B}z+C)},
\\ && 
\Ave{(\cdots)}_{\Gamma,B,C}
\to
\frac{1}{
\int Dz e^{\rPPc \hat{g}(z,\Gamma,\sqrt{B}z+C)} 
}
{\int Dz
(\cdots)
e^{\rPPc \hat{g}(z,\Gamma,\sqrt{B}z+C)} 
}.
\ee
Hence, for these symbols, we do not distinguish between the zero- and finite-temperature cases. The rhs of \Req{chi-1RSB} is slightly complicated in taking this limit, but a careful computation yields  
\be
\chi_{i \to \mu}^{(t+1)}
\to
\frac{1}{\ul{\mc{Z}}_{i\to \mu}^{(t+1)}}
\int Dz e^{\rPPc \hat{g}_{i\to \mu}^{(t+1)}(z)}
\partial_{C_{i\to \mu}^{(t+1)}} \hat{x}_{i \to \mu}^{(t+1)}(z)
=
\Ave{\partial_{C_{i\to \mu}^{(t+1)}} \hat{x}_{i \to \mu}^{(t+1)}(z)}_{i \to \mu}^{(t+1)},
\ee
where we introduced 
\be
&&
\hat{x}^{(t)}_{i\to \mu}(z)=\hat{x} \lb \Gamma^{(t)}_{i\to \mu},\sqrt{B^{(t)}_{i\to \mu} }z+C^{(t)}_{i\to \mu} \rb,
\\ &&
\hat{g}^{(t)}_{i\to \mu}(z)
=
g^{(t)}_{i\to \mu}(\hat{x}^{(t)}_{i\to \mu}(z),z)
=
\hat{g} \lb \Gamma^{(t)}_{i\to \mu},B^{(t)}_{i\to \mu}z+C^{(t)}_{i\to \mu}
\rb.
\ee
Noticing that $q_{1,\mu \to i}$ converges to $Q_{\mu \to i}$, we may write the 1RSB-BP algorithm at $\beta \to \infty$ as follows:
\subbe
\be
&&
\chi^{(t)}_{\mu \to i}=\sum_{j(\neq i)}A_{\mu j}^2\chi^{(t)}_{j \to \mu},
\\ &&
Q^{(t)}_{\mu \to i}=\sum_{j(\neq i)}A_{\mu j}^2 Q_{j \to \mu}^{(t)},
\\ &&
q^{(t)}_{0,\mu \to i}=\sum_{j(\neq i)}A_{\mu j}^2 (m_{j \to \mu}^{(t)})^2,
\\ &&
\T{\Gamma}_{\mu \to i}^{(t)}=\frac{A_{\mu i}^{2}}{\chi_{\mu \to i}^{(t)}},
\\ &&
\T{B}_{\mu \to i}^{(t)}=
\frac{A_{\mu i}^2}{\chi^{(t)}_{\mu \to i}}
\frac{(Q^{(t)}_{\mu \to i}-q^{(t)}_{0,\mu \to i})}{\chi^{(t)}_{\mu \to i}+ \rPPc (Q^{(t)}_{\mu \to i}-q^{(t)}_{0,\mu \to i})},
\\ &&
\T{C}^{(t)}_{\mu \to i}=
\frac{A_{\mu i}(y_{\mu}-\sum_{j(\neq i)}A_{\mu j}m_{j \to \mu}^{(t)})}{\chi^{(t)}_{\mu \to i}+ \rPPc (Q^{(t)}_{\mu \to i}-q^{(t)}_{0,\mu \to i})},
\\ &&
\Gamma_{i \to \mu}^{(t+1)}=\sum_{\nu (\neq \mu)}\T{\Gamma}_{\nu \to i}^{(t)},
\quad
B_{i \to \mu}^{(t+1)}=\sum_{\nu (\neq \mu)}\T{B}_{\nu \to i}^{(t)},
\quad
C_{i \to \mu}^{(t+1)}=\sum_{\nu (\neq \mu)}\T{C}_{\nu \to i}^{(t)},
\\ &&
\ul{\mc{Z}}_{i \to \mu}^{(t+1)}=\ul{\mc{Z}}\lb \Gamma_{i \to \mu}^{(t+1)}, B_{i \to \mu}^{(t+1)}, C_{i \to \mu}^{(t+1)}\rb,
\\ &&
m^{(t+1)}_{i \to \mu}
=\Ave{\hat{x}^{(t+1)}_{i\to \mu}(z)}_{i \to \mu}^{(t+1)}
=\frac{1}{\rPPc} \partial_{C_{i\to \mu}^{(t+1)}} \log \ul{\mc{Z}}^{(t+1)}_{i\to \mu} ,
\\ &&
Q^{(t+1)}_{i \to \mu}
=\Ave{ \lb \hat{x}^{(t+1)}_{i\to \mu} \rb^2 }_{i \to \mu}^{(t+1)}
=-\frac{2}{\rPPc}  \partial_{\Gamma_{i\to \mu}^{(t+1)}} \log \ul{\mc{Z}}^{(t+1)}_{i\to \mu} ,
\\ &&
\chi_{j\to \mu}^{(t+1)}=\Ave{\partial_{C_{i\to \mu}^{(t+1)}} \hat{x}_{i \to \mu}^{(t+1)}(z)}_{i \to \mu}^{(t+1)}.
\ee
\subee

\subsection{1RSB-SE}\Lsec{1RSB-SE}
The following quantities
\be
\chi_{\mu \to i},Q_{\mu \to i},q_{0,\mu \to i},\Gamma_{i \to\mu},B_{i \to\mu}
\ee
lose the dependence on $\mu,i$ in the large $N$ limit since they enjoy LLN. Meanwhile, 
\be
&&
C_{i \to \mu}^{(t+1)}=\sum_{\nu (\neq \mu)}\T{C}_{\nu \to i}^{(t)}
=
\sum_{\nu (\neq \mu)}
\frac{
A_{\nu i}(y_{\nu}-\sum_{j(\neq i)}A_{\nu j}m_{j \to \nu}^{(t)})
}{
\chi^{(t)}_{\nu \to i}+ \rPPc (Q^{(t)}_{\nu \to i}-q^{(t)}_{0,\nu \to i})
},
\no \\ &&
\LLNap
\frac{\alpha
}{ 
\chi^{(t)}+ \rPPc (Q^{(t)}-q^{(t)}_{0})
}x_{0i} 
+\frac{1}{ \chi^{(t)}+ \rPPc (Q^{(t)}-q^{(t)}_{0}) }
\sum_{\nu (\neq \mu)} \sum_{j (\neq i)}A_{\nu i}A_{\nu j} (x_{0j}-m_{j \to \nu}^{(t)})
\no \\ &&
\CLTapd
\frac{\alpha }{ \chi^{(t)}+ \rPPc (Q^{(t)}-q^{(t)}_{0}) }x_{0i} 
+
\frac{
\sqrt{\alpha \epsilon_{q_0}^{(t)}}
}{
\chi^{(t)}+ \rPPc (Q^{(t)}-q^{(t)}_{0})
}
z_{0,i\to \mu},
\Leq{C-largeN}
\ee
where the first and second approximations are by LLN and CLT, respectively: The symbol $\CLTapd$ means approximate equality in distribution and $z^{(t+1)}_{0,i\to \mu} \sim \mc{N}(0,1)$. Furthermore, we define
\be
\epsilon^{(t)}_{q_0}
=\frac{1}{N}\sum_{i=1}^{N}\lb x_{0i}-m_{i}^{(t)}\rb^2
\approx \frac{1}{N}\sum_{i=1}^{N}\lb x_{0i}-m_{i\to \mu}^{(t)}\rb^2.
\ee
Here, $m_i=\Ave{\hat{x}_i}_i$ is the expectation of $x_i$ using the full marginal, which becomes necessary when deriving AMP and is introduced later in \Req{full quantities}. This $\epsilon_{q_0}$ is interpreted as an MSE by regarding $m_i=\Ave{\hat{x}_i}_i$ (also $m_{i\to \mu}=\Ave{\hat{x}_{i\to \mu}}_{i\to \mu}$) as an estimator for $x_{0i}$, as discussed in \Rsec{Summary of}. Similarly,
\be
&&
\chi^{(t)}
=\sum_{j}A_{\mu j}^2\chi^{(t)}_{j \to \mu}
\LLNap
\frac{1}{N}\sum_{j}\chi^{(t)}_{j \to \mu}
\LLNap
\frac{1}{NM}\sum_{\mu}\sum_{j}\chi^{(t)}_{j \to \mu},
\no \\ &&
=
\frac{1}{NM}
\sum_{\mu}\sum_{j}
\frac{
\int Dz_1 e^{\rPPc \hat{g}\lb \Gamma^{(t)},\sqrt{B^{(t)}}z_1 + C_{j \to \mu}^{(t)} \rb }\partial_C 
\hat{x}\lb \Gamma^{(t)},\sqrt{B^{(t)}}z_1 + C_{j \to \mu}^{(t)} \rb
}{
\int Dz_1 e^{\rPPc \hat{g}\lb \Gamma^{(t)},\sqrt{B^{(t)}}z_1 + C_{j \to \mu}^{(t)} \rb }
}
\no \\ &&
\LLNap
\int dx_0 P_0(x_0)
\int Dz_0
\frac{
\int Dz_1 e^{\rPPc \hat{g}\lb\Gamma^{(t)},\sqrt{B^{(t)}}z_1+C^{(t)}(x_0,z_0) \rb}
\partial_C 
\hat{x}\lb \Gamma^{(t)},\sqrt{B^{(t)}}z_1+C^{(t)}(x_0,z_0) \rb
}{
\int Dz_1 e^{\rPPc \hat{g} \lb \Gamma^{(t)},\sqrt{B^{(t)}}z_1+C^{(t)}(x_0,z_0) \rb}
}
\\ &&
\eqqcolon
\E_{x_0,z_0}
\Ave{
\partial_C \hat{x}\lb \Gamma^{(t)},\sqrt{B^{(t)}}z_1+C^{(t)}(x_0,z_0) \rb
}^{(t)}_{z_1\mid x_0,z_0},
\ee
where
\be
&&
\E_{x_0,z_0}(\cdots)=\int dx_0 P_0(x_0) \int Dz_0(\cdots),
\Leq{ave_single_x0z0}
\\ &&
C^{(t+1)}(x_0,z_0)
=
\frac{\alpha }{ \chi^{(t)}+ \rPPc (Q^{(t)}-q^{(t)}_{0}) }x_{0} 
+
\frac{
\sqrt{\alpha \epsilon_{q_0}^{(t)}}
}{
\chi^{(t)}+ \rPPc (Q^{(t)}-q^{(t)}_{0})
}
z_{0},
\\ &&
\Ave{
(\cdots)
}^{(t)}_{z_1\mid x_0,z_0}
=
\frac{
\int Dz_1 e^{\rPPc \hat{g}\lb \Gamma^{(t)},\sqrt{B^{(t)}}z_1+C^{(t)}(x_0,z_0) \rb}
(\cdots)
}{
\int Dz_1 e^{\rPPc \hat{g} \lb \Gamma^{(t)},\sqrt{B^{(t)}}z_1+C^{(t)}(x_0,z_0) \rb}
}.
\Leq{ave_single_t}
\ee
In the same way,
\be
&&
Q^{(t)}
\LLNap
\E_{x_0,z_0}
\Ave{
\hat{x}^2\lb \Gamma^{(t)},\sqrt{B^{(t)}}z_1+C^{(t)}(x_0,z_0) \rb
}^{(t)}_{z_1\mid x_0,z_0},
\\ &&
q_0^{(t)}
\LLNap
\E_{x_0,z_0}
\lb \Ave{
\hat{x}\lb \Gamma^{(t)},\sqrt{B^{(t)}}z_1+C^{(t)}(x_0,z_0) \rb
}^{(t)}_{z_1\mid x_0,z_0}
\rb^2,
\\ &&
\epsilon_{q_0}^{(t)}
\approx \frac{1}{NM}\sum_{\mu}\sum_{i}(x_{0i}-m_{i\to \mu}^{(t)})^2
\no \\ &&
\LLNap \frac{1}{NM}\sum_{\mu}\sum_{i}
\lb 
x_{0i}
-
\frac{
\int Dz_1 e^{\rPPc \hat{g}\lb \Gamma^{(t)},\sqrt{B^{(t)}}z_1+C_{i \to \mu}^{(t)} \rb}
\hat{x}\lb \Gamma^{(t)},\sqrt{B^{(t)}}z_1+C_{i \to \mu}^{(t)} \rb
}{
\int Dz_1 e^{\rPPc \hat{g} \lb \Gamma^{(t)},\sqrt{B^{(t)}}z_1+C_{i \to \mu}^{(t)} \rb}
}
\rb^2
\no \\ &&
\LLNap 
\E_{x_0,z_0}
\lb 
x_{0}
-
\Ave{
\hat{x}\lb \Gamma^{(t)},\sqrt{B^{(t)}}z_1+C^{(t)}(x_0,z_0) \rb
}^{(t)}_{z_1\mid x_0,z_0}
\rb^2.
\ee
By organizing the above computations, we obtain \Req{SE_1RSB}.

\subsection{1RSB-AMP at $\beta \to \infty$}\Lsec{1RSB-AMP at}
We start from defining
\subbe
\Leq{full quantities}
\be
&&
\Gamma_{i}^{(t+1)}=\sum_{\nu}\T{\Gamma}_{\nu \to i}^{(t)},
\quad
B_{i}^{(t+1)}=\sum_{\nu}\T{B}_{\nu \to i}^{(t)},
\quad
C_{i}^{(t+1)}=\sum_{\nu}\T{C}_{\nu \to i}^{(t)},
\\ &&
\hat{x}^{(t)}_{i}(z)=\hat{x}\lb \Gamma_{i}^{(t)},\sqrt{B_{i}^{(t)}}z+C_{i}^{(t)}\rb,
\quad
\Ave{(\cdots)}_{i}^{(t)}
=
\Ave{(\cdots)}_{\Gamma_{i}^{(t)},B_{i}^{(t)},C_{i}^{(t)}},
\\ &&
m_{i}^{(t)}=\Ave{\hat{x}^{(t)}_{i}(z)}_{i}^{(t)},
\quad
Q_{i}^{(t)}=\Ave{\lb \hat{x}^{(t)}_{i}(z) \rb^2}_{i}^{(t)},
\quad
\chi_{i}^{(t)}
=\Ave{\partial_{C_{i}^{(t)}} \hat{x}^{(t)}_{i}(z)}_{i}^{(t)},
\\ &&
\chi_{\mu}^{(t)}=\sum_{j}A_{\mu j}^2\chi_{j}^{(t)},
\quad
Q_{\mu}^{(t)}=\sum_{j}A_{\mu j}^2Q_{j}^{(t)},
\quad
q_{0\mu}^{(t)}=\sum_{j}A_{\mu j}^2 \lb m_{j}^{(t)} \rb^2.
\ee
\subee
The differences $\Gamma_{i}-\Gamma_{i\to \mu},B_{i}-B_{i\to \mu},\chi_{\mu}-\chi_{\mu \to i},Q_{\mu}-Q_{\mu \to i},q_{0\mu}-q_{0,\mu \to i}$ are $O(N^{-1})$ and negligible, while $C_i-C_{i\to \mu}$ should be handled appropriately. Let us denote
\be
&&
R^{(t)}_{\mu}
=\frac{1}{ \chi^{(t)}_{\mu} + \rPPc \lb Q^{(t)}_{\mu}-q^{(t)}_{0\mu} \rb }
\lb y_{\mu}-\sum_{j}A_{\mu j}m_{j \to \mu}^{(t)}\rb, 
\Leq{R-def}
\\ &&
\Delta C^{(t+1)}_{i\to \mu}=C^{(t+1)}_{i}-C^{(t+1)}_{i\to \mu}=\T{C}^{(t)}_{\mu \to i}
=\frac{A_{\mu i}}{\chi_{\mu}^{(t)}+\rPPc(Q_{\mu}^{(t)}-q_{0\mu}^{(t)})} \lb y_{\mu}-\sum_{j(\neq i)}A_{\mu j} m_{j \to \mu}^{(t)} \rb
\no \\ &&
\approx 
A_{\mu i}R_{\mu}^{(t)}
+ \frac{A_{\mu i}^2}{\chi_{\mu}^{(t)}+\rPPc(Q_{\mu}^{(t)}-q_{0\mu}^{(t)})}m_{i}^{(t)}
 .
\Leq{Ctilde2R}
\ee
The linear perturbation yields
\be
&&
m_{i \to \mu}^{(t)}
\approx
m_{i}^{(t)}-\left. \Part{m_{i}^{(t)}}{C}{}\right|_{C=C_i^{(t)} }\Delta C_{i \to \mu}^{(t)}.
\ee
From simple algebras, we have
\be
\left. \Part{m_{i}^{(t)}}{C}{}\right|_{C=C_i^{(t)} }
=
\chi_{i}^{(t)}+\rPPc \lb Q_i^{(t)}-(m_{i}^{(t)})^2\rb.
\Leq{m_diff_C}
\ee
Hence,
\be
&&
\hspace{-1cm}
m_{i \to \mu}^{(t)}
\approx
m_{i}^{(t)}
-
A_{\mu i}
\lb 
\chi_{i}^{(t)}+\rPPc \lb Q_{i}^{(t)}-\lb m_{i}^{(t)} \rb^2 \rb
 \rb
R_{\mu}^{(t-1)}
\no \\ &&
-
\frac{
A^2_{\mu i}
\lb 
\chi_{i}^{(t)}+\rPPc \lb Q_{i}^{(t)}-\lb m_{i}^{(t)} \rb^2 \rb
 \rb
 }{
 \chi_{\mu}^{(t-1)}+\rPPc \lb Q_{\mu}^{(t-1)}-q_{0\mu}^{(t-1)} \rb
 }
m_{i}^{(t-1)}
.
\ee
Inserting this into \Req{R-def} and simplifying it, we have 
\be
&&
R_{\mu}^{(t)}
\approx 
\frac{
y_{\mu}-\sum_{i}A_{\mu i}m_{i}^{(t)}
}{
 \chi_{\mu}^{(t)}+\rPPc \lb Q_{\mu}^{(t)}-q_{0\mu}^{(t)} \rb
 }
+R_{\mu}^{(t-1)}. 
\Leq{R_iteration}
\ee
\BReqs{Ctilde2R}{R_iteration} lead us to the following 1RSB-AMP algorithm:
\subbe
 \be
 &&
  \hspace{-1cm}
m_{i}^{(t)}=\Ave{\hat{x}_{i} }_{i}^{(t)},\quad
Q_{i}^{(t)}=\Ave{\lb \hat{x}_{i}^{(t)}\rb^2 }_{i}^{(t)},\quad
\chi_{i}^{(t)}=\Ave{\partial_{C_{i}^{(t)}}\hat{x}_{i}^{(t)} }_{i}^{(t)},
 \\ &&
  \hspace{-1cm}
\chi_{\mu}^{(t)}=\sum_{i}A_{\mu i}^2 \chi_{i}^{(t)},\quad
Q_{\mu}^{(t)}=\sum_{i}A_{\mu i}^2 Q_{i}^{(t)},\quad
q_{0\mu}^{(t)}=\sum_{i}A_{\mu i}^2 \lb m_{i}^{(t)} \rb^2,
 \\ &&
  \hspace{-1cm}
R_{\mu}^{(t)}=
\frac{
y_{\mu}-\sum_{i}A_{\mu i}m_{i}^{(t)}
}{
 \chi_{\mu}^{(t)}+\rPPc \lb Q_{\mu}^{(t)}-q_{0\mu}^{(t)} \rb
 }
+R_{\mu}^{(t-1)},
  \\ &&
  \hspace{-1cm}
\T{\Gamma}_{\mu \to i}^{(t)}=\frac{A_{\mu i}^2}{ \chi_{\mu}^{(t)}},
  \\ &&
  \hspace{-1cm}
\T{B}_{\mu \to i}^{(t)}=\frac{A_{\mu i}^2}{ \chi_{\mu}^{(t)}}\frac{ Q_{\mu}^{(t)}-q_{0\mu}^{(t)}
}{ \chi_{\mu}^{(t)}+\rPPc \lb Q_{\mu}^{(t)}-q_{0\mu}^{(t)} \rb},
  \\ &&
  \hspace{-1cm}
\T{C}^{(t)}_{\mu \to i}
=
A_{\mu i}R_{\mu}^{(t)}
+ \frac{A_{\mu i}^2}{\chi_{\mu}^{(t)}+\rPPc(Q_{\mu}^{(t)}-q_{0\mu}^{(t)})}m_{i}^{(t)},
  \\ &&
  \hspace{-1cm}
 \Gamma_{i}^{(t+1)}=
 \sum_{\mu }\T{\Gamma}_{\mu \to i}^{(t)},
  \\ &&
  \hspace{-1cm}
B_{i}^{(t+1)}=\sum_{\mu }\T{B}_{\mu \to i}^{(t)}
,
  \\ &&
  \hspace{-1cm}
C_{i}^{(t+1)}=\sum_{\mu }\T{C}_{\mu \to i}^{(t)}. \ee
 \subee
As the initial condition, it is appropriate to give $\{m_i^{(0)},\chi_i^{(0)},Q_{i}^{(0)}\}$ with $R_{\mu}^{(-1)}=0$. Since $\Gamma_i,B_i,\chi_{\mu},Q_{\mu},q_{0\mu}$ enjoy LLN as can be seen in the rhs and hence their index dependence is negligible,  a simpler version of 1RSB-AMP is derivable and is given in \Req{AMP_1RSB_simpler}.

\subsection{Computations of free entropy and related quantities}\Lsec{Computations of}
According to the standard BP prescription, the free entropy can be decomposed as
\be
N s=\sum_{\mu=1}^{M}\log \ul{\mc{Z}}_{\mu}+\sum_{i=1}^{N}(1-d_i)\log \ul{\mc{Z}}_i,
\Leq{replicated free energy}
\ee
where $d_i$ is the degree of $i$th variable node and in the present case $d_i=M$ holds, and $\ul{\mc{Z}}_i,\ul{\mc{Z}}_{\mu}$ are the normalization constants for the $i$th variable node marginal and for the $\mu$th function node marginal, each of which are defined in
\be
&&
\ul{P}_{i}(\ul{x}_i)=\frac{1}{\ul{\mc{Z}}_i}\ul{\varphi}(\ul{x}_i)\prod_{\mu \in \partial i}
\T{\ul{\phi}}_{\mu \to i}(\ul{x}_{i}),
\\ &&
\ul{P}_{\mu}(\ul{\V{x}}_{\mu})=\frac{1}{\ul{\mc{Z}}_{\mu}}\ul{\Phi}_{\mu}(\ul{\V{x}}_{\mu})
\prod_{i \in \partial \mu}
\lb 
\ul{\varphi}(\ul{x}_i)
\prod_{\nu \in \partial i\bs \mu}\T{\ul{\phi}}_{\nu \to i}(\ul{x}_{i})
\rb.
\ee
The normalization of $\ul{\T{\phi}}$ is irrelevant in \Req{replicated free energy} since it cancels between the first and second terms. Hence, for simplicity we ignore the normalization and have
\be
\T{\ul{\phi}}_{\mu \to i}(\ul{x}_{i})
=e^{
-\frac{1}{2}\beta \T{\Gamma}_{\mu \to i}\sum_{a=1}^{\PPc}x_{ai}^2
+\frac{1}{2}\beta^2 \T{B}_{\mu \to i}\lb \sum_{a=1}^{\PPc}x_{ai}\rb^2
+\beta \T{C}_{\mu \to i}\sum_{a=1}^{\PPc}x_{ai}\
 }.
\ee
Putting this into the definition, we have
\be
&&
\ul{\mc{Z}}_i
=\int d\ul{x}_i \varphi(\ul{x}_i)\prod_{\mu \in \partial i}\ul{\T{\phi}}_{\mu \to i}(\ul{x}_i) 
=\ul{\mc{Z}}_{i \to \mu}\int d\ul{x}_i \ul{\phi}_{i \to \mu}(\ul{x}_i)\ul{\T{\phi}}_{\mu \to i}(\ul{x}_i) 
\Leq{ulZ_i-1}
\\ &&
=\int d\ul{x}_i
\int Dz 
\prod_{a=1}^{\PPc}
e^{
\beta\lbb 
-\frac{1}{2}\lb \sum_{\mu \in \partial i}\T{\Gamma}_{\mu \to i}\rb x_{ai}^2
+ \lb  \sqrt{\sum_{\mu \in \partial i}\T{B}_{\mu \to i}}z 
+ \lb \sum_{\mu \in \partial i}\T{C}_{\mu \to i} \rb  \rb x_{ai}
-J(x_{ai})
\rbb }
\no \\ &&
=
\int Dz 
\lb 
\int dx
e^{
\beta\lbb
-\frac{1}{2}\Gamma_i  x^2
+ \lb  \sqrt{B_i}z + C_i \rb  x
-J(x)
\rbb }
\rb^{\PPc}
=\ul{\mc{Z}}(\Gamma_i,B_i,C_i),
\Leq{ulZ_i-2}
\ee 
where
\be
\Gamma_i=\sum_{\mu \in \partial i}\T{\Gamma}_{\mu \to i},\quad
B_i=\sum_{\mu \in \partial i}\T{B}_{\mu \to i},\quad
C_i=\sum_{\mu \in \partial i}\T{C}_{\mu \to i}.
\ee
Both the expressions \NReqs{ulZ_i-1}{ulZ_i-2} are useful, which will be later demonstrated. On the other hand, $\ul{\mc{Z}}_{\mu}$ becomes a little bit more complicated:
\be
&&
\ul{\mc{Z}}_{\mu}=\int d\ul{\V{x}}
\frac{1}{\sqrt{2\pi\tau}}e^{-\frac{1}{2\tau}\sum_{a=1}^{\PPc}\lb y_{\mu}-\sum_{i=1}^N A_{\mu i}x_{ai} \rb^2 }
\no \\ && \times
\prod_{i\in \partial \mu}
e^{
-\frac{1}{2}\beta \lb \sum_{\nu \in \partial i \bs \mu}\T{\Gamma}_{\mu \to i}\rb \sum_{a=1}^{\PPc}x_{ai}^2
+ \frac{1}{2}\beta^2 \lb \sum_{\nu \in \partial i \bs \mu}\T{B}_{\mu \to i}\rb  \lb \sum_{a=1}^{\PPc}x_{ai}\rb^2
+ \lb \sum_{\nu \in \partial i \bs \mu}\T{C}_{\mu \to i}\rb \sum_{a=1}^{\PPc}x_{ai}
-\sum_{a=1}^{\PPc}J(x_{ai})
 }
\no \\ && 
=
\lb \prod_{i\in \partial \mu}
\ul{\mc{Z}}_{i \to \mu}
\rb
\int d\ul{\V{x}}
\frac{1}{\sqrt{2\pi\tau}}e^{-\frac{1}{2\tau}\sum_{a=1}^{\PPc}\lb y_{\mu}-\sum_{i=1}^N A_{\mu i}x_{ai} \rb^2 }
\lb \prod_{i\in \partial \mu}
\ul{\phi}_{i \to \mu}(\ul{x}_i)
\rb
\no \\ &&
=
\lb \prod_{i\in \partial \mu}
\ul{\mc{Z}}_{i \to \mu}
\rb
\Ave{
\frac{1}{\sqrt{2\pi\tau}}e^{-\frac{1}{2\tau}\sum_{a=1}^{\PPc}\lb y_{\mu}-\sum_{i=1}^N A_{\mu i}x_{ai} \rb^2 }
}_{\prod_{i\in \partial \mu}
\ul{\phi}_{i \to \mu}(\ul{x}_i)},
\ee
where $\ul{\mc{Z}}_{i \to \mu}$ is the normalization constant of the message $\ul{\phi}_{i \to \mu}(\ul{x}_i)$. The contribution of these normalization constants thus becomes
\be
\sum_{\mu}\log \ul{\mc{Z}}_{\mu}
\propto
\sum_{\mu}\sum_{i \in \partial \mu}\log \ul{\mc{Z}}_{i \to \mu}
=
\sum_{i}\sum_{\mu \in \partial i}\log \ul{\mc{Z}}_{i \to \mu}.
\Leq{Zmu-irrelevant}
\ee
Similar contribution can be derived from $\ul{\mc{Z}}_i$ according to \Req{ulZ_i-1}. Namely,
\be
&&
\sum_i d_i \log \ul{\mc{Z}}_i
=
\sum_i d_i \lb \log \ul{Z}_{i\to \mu}+ \log \int d\ul{x}_i\ul{\phi}_{i\to \mu}(\ul{x}_i) \ul{\T{\phi}}_{\mu \to i}(\ul{x}_i) \rb
\no \\ && 
=
\sum_i \sum_{\mu \in \partial i} \lb \log \ul{Z}_{i\to \mu}+ \log \int d\ul{x}_i\ul{\phi}_{i\to \mu}(\ul{x}_i) \ul{\T{\phi}}_{\mu \to i}(\ul{x}_i) \rb.
\ee
The first term thus cancels with \Req{Zmu-irrelevant}. Resultantly, we have
\be
&&
s
=
\sum_{\mu}
\log 
\Ave{
\frac{1}{\sqrt{2\pi\tau}}e^{-\frac{1}{2\tau}\sum_{a=1}^{\PPc}\lb y_{\mu}-\sum_{i=1}^N A_{\mu i}x_{ai} \rb^2 }
}_{\prod_{i\in \partial \mu}
\ul{\phi}_{i \to \mu}(\ul{x}_i)}
+\sum_i \log \ul{\mc{Z}}(\Gamma_i,B_i,C_i) 
\no \\ &&
-
\sum_i \sum_{\mu \in \partial i} \log \int d\ul{x}_i\ul{\phi}_{i\to \mu}(\ul{x}_i) \ul{\T{\phi}}_{\mu \to i}(\ul{x}_i).
\Leq{replicated free energy-2}
\ee
As discussed in \Rsecs{1RSB-SE}{1RSB-AMP at}, $\Gamma, B$ lose their index dependence in the large $N$ limit, and $C_i$ becomes identical to \Req{C-largeN}, though $z_{0, i\to\mu}$ should be rewritten as $z_{0i}\sim \mc{N}(0,1)$. As a result, thanks to LLN, we have
\be
&&
\hspace{-1cm}
\frac{1}{N}\sum_i \log \ul{\mc{Z}}(\Gamma_i,B_i,C_i)
\LLNap
\int dx_0 P(x_0)\int Dz_0 
\log \int Dz_1 \lb \int dx e^{\beta \lb-\frac{1}{2}\Gamma x^2+(\sqrt{B}z_1+C(x_0,z_0))x-J(x) \rb} \rb^{\PPc}
\no \\ &&
\hspace{-1cm}
\xrightarrow{\beta \to \infty, \beta\PPc\to \rPPc}
\E_{x_0,z_0}
\log \int Dz_1 e^{\rPPc \hat{g}(\Gamma,\sqrt{B} z_1+C(x_0,z_0))}.
\ee
Besides, CLT implies
\be
&&
\Ave{
\frac{1}{\sqrt{2\pi\tau}}e^{-\frac{1}{2\tau}\sum_{a=1}^{\PPc}\lb y_{\mu}-\sum_{i=1}^N A_{\mu i}x_{ai} \rb^2 }
}_{\prod_{i\in \partial \mu}\ul{\phi}_{i \to \mu}}
\no \\ &&
\CLTap
\int Dz_0 \prod_{a=1}^{\PPc} \int Dz_1 \frac{1}{\sqrt{2\pi \tau}}
e^{
-\frac{1}{2\tau}
\lb
y_{\mu}-\sum_{i=1}^NA_{\mu i} m_{i \to \mu}
-\sqrt{Q_{\mu}-q_{1\mu}}z_1
-\sqrt{q_{1\mu}-q_{0\mu}}z_0
\rb^2
}
\no \\ &&
=
\int Dz_0 
\frac{1}{\sqrt{2\pi (\tau+Q_{\mu}-q_{1\mu})}^{\PPc}}
e^{
-\frac{\PPc \lb r_{\mu }-\sqrt{q_{1\mu}-q_{0\mu}}z_0 \rb^2}{2(\tau+Q_{\mu}-q_{1\mu})}
}
\no \\ &&
\xrightarrow{\tau \to 0, \chi=\beta(Q-q_1),\beta \PPc \to \rPPc}
\sqrt{\frac{\beta}{2\pi \chi_{\mu}}}^{\PPc}
\sqrt{\frac{\chi_{\mu}}{\chi_{\mu} +\rPPc(q_{1\mu}-q_{0\mu}) }}
e^{
-\frac{1}{2}\frac{\rPPc }{\chi_{\mu} +\rPPc(q_{1\mu}-q_{0\mu})}r_{\mu }^2
}
\ee
where $r_{\mu }=y_{\mu}-\sum_{i=1}^NA_{\mu i} m_{i \to \mu}$. Again thanks to LLN, $\chi_{\mu},Q_{\mu},q_{1\mu},q_{0\mu}$ lose its $\mu$-dependence in the large $N$ limit. Moreover, $\frac{1}{N}\sum_{\mu}r_{\mu }^2=\frac{1}{N}\sum_{\mu}(y_{\mu}-\sum_{i=1}^NA_{\mu i} m_{i \to \mu})^2$ converges to $\alpha \epsilon_{q_0}$. As a result, we get
\be
&&
\frac{1}{N}\sum_{\mu}
\log \Ave{
\frac{1}{\sqrt{2\pi\tau}}e^{-\frac{1}{2\tau}\sum_{a=1}^{\PPc}\lb y_{\mu}-\sum_{i=1}^N A_{\mu i}x_{ai} \rb^2 }
}_{\prod_{i\in \partial \mu}\ul{\phi}_{i \to \mu}}
\no \\ &&
\LLNap
\alpha 
\Biggl\{
\frac{\PPc}{2}\log \frac{\beta}{2\pi \chi}
+
\frac{1}{2}\log \frac{\chi}{\chi +\rPPc(q_{1}-q_{0}) }
-
\frac{1}{2}\frac{
\rPPc \epsilon_{q_0}
}{\chi +\rPPc(q_{1}-q_{0}) }
\Biggr\}
\\ &&
\xrightarrow{\beta \to \infty, q_1\to Q,  \beta\PPc\to \rPPc}
\frac{\alpha}{2} 
\Biggl\{
\log \frac{\chi}{\chi +\rPPc(Q-q_{0}) }
-
\frac{
\rPPc \epsilon_{q_0}
}{\chi +\rPPc(Q-q_{0}) }
\Biggr\}.
\ee
The remaining term can be rewritten as
\be
\int d\ul{x}_i\ul{\phi}_{i\to \mu}(\ul{x}_i) \ul{\T{\phi}}_{\mu \to i}(\ul{x}_i) 
=\frac{\ul{\mc{Z}}(\Gamma_{i},B_{i},C_{i})}{\ul{\mc{Z}}(\Gamma_{i \to \mu},B_{i \to \mu},C_{i \to \mu})}.
\ee
The differences $\T{\Gamma}_{\mu \to i}=\Gamma_i - \Gamma_{i\to \mu},\T{B}_{\mu \to i}=B_i - B_{i\to \mu}$ are $O(N^{-1})$, while $\T{C}_{\mu \to i}=C_i - C_{i\to \mu}=O(N^{-1/2})$. We should expand the denominator w.r.t. $\T{\Gamma}_{\mu \to i},\T{B}_{\mu \to i},\T{C}_{\mu \to i}$ up to the $O(N^{-1})$ terms. The result is (denoting $\ul{\mc{Z}}(\Gamma_{i},B_{i},C_{i})=\ul{\mc{Z}}_i, \hat{g}_i=\hat{g}(\Gamma_i,\sqrt{B_i}z+C_i),~\Ave{\cdots}_i=\frac{1}{\ul{\mc{Z}}_i} \int Dz (\cdots)e^{\rPPc \hat{g}_i}$)
\be
&&
\log \frac{\ul{\mc{Z}}(\Gamma_{i},B_{i},C_{i})}{\ul{\mc{Z}}(\Gamma_{i \to \mu},B_{i \to \mu},C_{i \to \mu})}
\approx
\frac{\partial_{C_i}\ul{\mc{Z}}_i}{\ul{\mc{Z}}_i}\T{C}_{\mu \to i}
+\frac{1}{2}\frac{ (\partial_{C_i}\ul{\mc{Z}}_i)^2-\ul{\mc{Z}}_i\partial_{C_i}^2\ul{\mc{Z}}_i }{\ul{\mc{Z}}_i^2}\T{C}^2_{\mu \to i}
\no \\ &&
+
\frac{\partial_{B_i}\ul{\mc{Z}}_i}{\ul{\mc{Z}}_i}\T{B}_{\mu \to i}
+
\frac{\partial_{\Gamma_i}\ul{\mc{Z}}_i}{\ul{\mc{Z}}_i}\T{\Gamma}_{\mu \to i}.
\ee
Here 
\be
&&
\frac{\partial_{C_i}\ul{\mc{Z}}_i}{\ul{\mc{Z}}_i}=\rPPc \Ave{\hat{x}_i}_i
,\quad
\frac{\partial^2_{C_i}\ul{\mc{Z}}_i}{\ul{\mc{Z}}_i}=\rPPc \lb \Ave{\partial_{C_i} \hat{x}_i} +\rPPc \Ave{\hat{x}_i^2}_i \rb
,\quad
\frac{\partial_{\Gamma_i}\ul{\mc{Z}}_i}{\ul{\mc{Z}}_i}=-\frac{1}{2}\rPPc \Ave{\hat{x}_i^2}_i 
,
\no \\ &&
\partial_{B_i}\ul{\mc{Z}}_i
=\int Dz \frac{1}{2}\rPPc \frac{z \hat{x}_i}{\sqrt{B_i}}e^{\rPPc \hat{g}_i}
=-\frac{1}{2}\rPPc  \int (Dz)' \frac{\hat{x}_i}{\sqrt{B_i}}e^{\rPPc \hat{g}_i}
=\frac{1}{2}\rPPc  \int Dz \frac{1}{\sqrt{B_i}}\partial_{z}\lb \hat{x}_ie^{\rPPc \hat{g}_i}\rb
\no \\ &&
=\frac{1}{2}\rPPc \int Dz \frac{1}{\sqrt{B_i}}\Part{h}{z}{}\partial_{h}\lb \hat{x}_ie^{\rPPc \hat{g}_i}\rb
=\frac{1}{2}\rPPc  \int Dz \lb \lb \partial_{C_i}\hat{x}_i \rb + \rPPc \hat{x}^2_i \rb e^{\rPPc \hat{g}_i}
=\frac{\ul{\mc{Z}}_i}{2}\rPPc\lb \Ave{\partial_{C_i}\hat{x}_i}+\rPPc  \Ave{\hat{x}^2_i}\rb,
\ee
where we put $h=\sqrt{B_i}z+C_i$. Thus, 
\be
&&
\sum_{\mu\in \partial i}
\log \frac{\ul{\mc{Z}}(\Gamma_{i},B_{i},C_{i})}{\ul{\mc{Z}}(\Gamma_{i \to \mu},B_{i \to \mu},C_{i \to \mu})}
\approx 
\rPPc \Ave{\hat{x}_i}_iC_i-\frac{1}{2}\rPPc \lb \Ave{\partial_{C_i} \hat{x}_i}
+\rPPc \lb \Ave{\hat{x}_i^2} - \Ave{\hat{x}_i}^2 \rb \rb
\sum_{\mu\in \partial i}\T{C}_{\mu \to i}^2
\no \\ &&
+\frac{1}{2}\rPPc \lb \Ave{\partial_{C_i} \hat{x}_i}+\rPPc \Ave{\hat{x}_i^2}  \rb B_i
-\frac{1}{2}\rPPc \Ave{\hat{x}_i^2}  \Gamma_i.
\Leq{sum-log-ZZ}
\ee
Remember the following relations:
\be
&&
\Gamma_i \LLNap \frac{\alpha}{\chi},\quad
B_i\LLNap\frac{\alpha}{\chi}\frac{(Q-q_0)}{\chi+\rPPc (Q-q_0)},\quad
 \\ &&
C_i
=
\sum_{\mu\in \partial i}
\frac{A_{\mu i}\lb y_{\mu}-\sum_{j(\neq i)}A_{\mu j}m_{j\to \mu } \rb
}{ \chi_{\mu}+\rPPc ( Q_{\mu}- q_{0\mu} ) }
\LLNap
\frac{\alpha x_{0i}+\sqrt{\alpha \epsilon_{q_0}} z_{0i}}{\chi+\rPPc (Q-q_0)}
+
\frac{\sum_{j(\neq i)}\sum_{\mu}A_{\mu i}A_{\mu j}(x_{0j}-m_{j\to \mu })}{\chi+\rPPc (Q-q_0)}
 \\ &&
\CLTapd
\frac{\alpha x_{0i}}{\chi+\rPPc (Q-q_0)}
+
\frac{\sqrt{\alpha \epsilon_{q_0}} z_{0i}}{\chi+\rPPc (Q-q_0)}
~~(z_{0i}\sim \mc{N}(0,1)),
\Leq{C_i-CLT}
 \\ &&
\sum_i \Ave{\partial_{C_i} \hat{x}_i}=N\chi ,\quad
\sum_i m_i=0,\quad
\sum_i \Ave{\hat{x}^2_i}=NQ, \quad
\sum_i m_i^2=Nq_0.\no
\ee
A seemingly nontrivial contribution in \Req{sum-log-ZZ} is $\sum_{\mu\in \partial i}\T{C}_{\mu \to i}^2$. However, from a simple scaling argument, only a part of terms showing the CLT scaling in \Req{C_i-CLT}  survives since other terms yield vanishing contribution in the large $N$ limit. Namely,
\be
\sum_{\mu\in \partial i}\T{C}_{\mu \to i}^2
=
\sum_{\mu\in \partial i}
\lb \frac{A_{\mu i}\lb y_{\mu}-\sum_{j(\neq i)}A_{\mu j}m_{j\to \mu } \rb
}{ \chi_{\mu}+\rPPc ( Q_{\mu}- q_{0\mu} ) }
\rb^2
\LLNap
\frac{\alpha \epsilon_{q_0} }{(\chi+\rPPc ( Q- q_{0} ))^2}.
\ee
Inserting all these contributions and denoting 
\be
\tm=\frac{1}{N}\sum_{i}x_{0i}m_i,
\ee
we have
\be
&&
\frac{1}{N}\sum_i \sum_{\mu\in \partial i}
\log \frac{\ul{\mc{Z}}(\Gamma_{i},B_{i},C_{i})}{\ul{\mc{Z}}(\Gamma_{i \to \mu},B_{i \to \mu},C_{i \to \mu})}
\approx
\rPPc \frac{\alpha \tm}{Y}
+
\rPPc \frac{\sqrt{\alpha \epsilon_{q_0}}}{Y}\int Dz_0 z_0\Ave{\hat{x}}
- \frac{1}{2}\rPPc\frac{\alpha \epsilon_{q_0}}{Y}
\no \\ &&
+ \frac{1}{2}\rPPc(\chi+\rPPc Q)\frac{\alpha (Q-q_0)}{\chi Y}
- \frac{1}{2}\rPPc\frac{\alpha Q}{\chi}.
\ee
where we put 
\be
&&
Y=\chi+\rPPc (Q-q_{0}),
\ee
and the following relations hold:
\be
&&
\epsilon_{q_0}=\rho  \sigma_x^2-2\tm+q_0,
\\ &&
\frac{\sqrt{\alpha \epsilon_{q_0}}}{Y}\int Dz_0 z_0\Ave{\hat{x}}
=
\frac{\sqrt{\alpha \epsilon_{q_0}}}{Y}\int Dz_0 \partial_{z_0}\Ave{\hat{x}}
=
\frac{\sqrt{\alpha \epsilon_{q_0}}}{Y}\int Dz_0 \Part{C}{z}{}\partial_{C}\Ave{\hat{x}}
\no \\ &&
=
\frac{\alpha \epsilon_{q_0}}{Y^2}\int Dz_0 \lb \Ave{\partial_C \hat{x}}+\rPPc( \Ave{\hat{x}^2}-\Ave{\hat{x}}^2 )\rb 
=
\frac{\alpha \epsilon_{q_0}}{Y}.
\ee
Inserting these and performing some simple algebras, we get
\be
&&
\frac{1}{N}\sum_i \sum_{\mu\in \partial i}
\log \frac{\ul{\mc{Z}}(\Gamma_{i},B_{i},C_{i})}{\ul{\mc{Z}}(\Gamma_{i \to \mu},B_{i \to \mu},C_{i \to \mu})}
\approx
- \frac{1}{2}\rPPc \frac{\alpha \rho  \sigma_x^2 }{Y}. 
\ee
Inserting all the computations so far into \Req{replicated free energy-2}, we obtain the free entropy as \Req{s-1RSB}.

\section{Replica approach at the 1RSB level}\Lsec{Replica}
Here, we perform the 1RSB computation using the replica method, to check the validity of our formula in the message passing approach. 

Following the standard prescription of the replica method, to compute the average of $\log Z_{\beta,\tau}(\V{y}, \V{A})$ over $\V{x}_0,\V{A}$, we assess the $\rn$th moment of the partition function. Assuming $\rn$ is positive integer, we have
\be
&&
e^{N\psi(\rn)}\coloneqq \mathbb{E}_{\V{x}_0,\V{A}}\left[Z^\rn_{\beta,\tau}(\V{y}, \V{A})\right]  
 =\frac{1}{\sqrt{2 \pi \tau}^{M\rn}} \int d \V{x}_0 d \V{A} P_0(\V{x}_0) P_A(\V{A}) 
 \int \lb \prod_{a=1}^{\rn}d \V{x}_a\rb 
 \no \\ &&
\times  \exp \lb 
 - \sum_{a=1}^{\rn} \lb \frac{1}{2 \tau}\left\| \V{A}\left(\V{x}_0-\V{x}_a\right) \right\|_2^2 +\beta J \lb \V{x}_a \rb \rb 
 \rb,
\ee
where $\mathbb{E}_{\V{x}_0,\V{A}}$ denotes the expectation over $\V{x}_0$ and $\V{A}$, and $P_A(\V{A})$ is the distribution of $\V{A}$ that is, as described in \Rsec{Problem setting}, the multivariate zero-mean Gaussian distribution with covariance $N^{-1}\V{I}_N$ where $\V{I}_N$ denotes $N\times N$ identity matrix. To proceed further, the observation that the quantity 
\be
\theta_\mu^a=\sum_{i=1}^N A_{\mu i}\left(x_{0i}-x_{ai} \right)
\ee
obeys a Gaussian distribution in the large $N$ limit thanks to CLT is imporant: Its mean is zero but its covariance can be nontrivial. In the 1RSB ansatz, the $\rn$ replicas are partitioned into non-overlapped $\rn/\PPc$ groups of the identical size $\PPc$. Denoting the group of replica $a$ as $B(a)$, the 1RSB parametrization of the overlaps among replicas $(a,b=1,\ldots,\rn)$ and the true signal are given by
\be
&&
\E_{\V{A}}\left[\Ave{\frac{1}{N}\sum_{i=1}^{N}x_{0i} x_{ai}}\right]\eqqcolon  m, 
\\ &&
\E_{\V{A}}\left[\Ave{\frac{1}{N}\sum_{i=1}^{N}x_{ai} x_{bi}}\right]\eqqcolon  
\begin{cases}
Q & \text { for } a=b,
\\ q_1 & \text { for } B(a)=B(b) \text { and } a \neq b,
\\ q_0 & \text { for } B(a) \neq B(b),
\end{cases}
\ee
where $\Ave{\cdots}$ denotes the average over the Boltzmann distribution. The covariance of $\theta$ are written using these overlaps as 
\be
\E_{\V{A}}\left[\Ave{\theta_\mu^a \theta_\nu^b}\right]= 
\delta_{\mu \nu}\times
\begin{cases}
\varepsilon_Q:=Q-2 m+\rho \sigma_x^2 & \text { for } a=b,
\\ \varepsilon_{q_1}:=q_1-2 m+\rho \sigma_x^2 & \text { for } B(a)=B(b) \text { and } a \neq b,
\\ \varepsilon_{q_0}:=q_0-2 m+\rho \sigma_x^2 & \text { for } B(a) \neq B(b).
\end{cases}
\ee
These imply that $\theta$ is expressed as a sum of independent Gaussian random variables:
\be
\theta_\mu^{a}\overset{\textrm{d}}{=} \sqrt{Q-q_1} u_\mu^{a}+\sqrt{q_1-q_0} v_\mu^{B(a)}+\sqrt{\varepsilon_{q_0}} w_\mu
\ee
where $u_\mu^{a},v_\mu^{B(a)},w_{\mu}$ are such independent Gaussian random variables with zero mean and unit variance. Hence, given the overlap values $Q(\{\V{x}_a\}_a),q_1(\{\V{x}_a\}_a),q_0(\{\V{x}_a\}_a),m(\{\V{x}_a\}_a,\V{x}_0)$, the average over $\V{A}$ can be replaced by those over $u_\mu^{a},v_\mu^{B(a)},w_{\mu}$. This yields
\be
e^{N\psi(\rn)}=\int dQ dq_1 dq_0 dm~\mc{S}(Q,q_1,q_0,m)\mc{E}(Q,q_1,q_0,m),
\ee
where $\mc{S}$ and $\mc{E}$ are entropic and energetic contributions whose explicit forms are given by
\subbe
\be
&&
\mathcal{E}^{\frac{1}{M}}=\left(\frac{1}{2 \pi \tau}\right)^{\frac{\rn}{2}} \int D w\left[\int D v\left\{\int D u \exp \left(-\frac{1}{2 \tau}\left(\sqrt{Q-q_1} u^{a}+\sqrt{q_1-q_0} v^{B(a)}+\sqrt{\varepsilon_{q_0}} w\right)^2\right)\right\}^{\PPc}\right]^{\frac{\rn}{\PPc}}
\no \\ &&
=\left(\frac{1}{2 \pi}\right)^{\frac{\rn}{2}}\left(\frac{1}{\tau+Q-q_1}\right)^{\frac{\rn}{2}}\left(\frac{1}{1+\frac{\PPc \left(q_1-q_0\right)}{\tau+Q-q_1}}\right)^{\frac{\rn}{2 \PPc}} \sqrt{\frac{1}{1+\frac{\rn \varepsilon_{q_0}}{\tau+Q-q_1+\PPc \left(q_1-q_0\right)}}},
\\ &&
\mathcal{S} 
=
\int d\V{x}_0  \lbb \prod_{a=1}^\rn d \V{x}_a\rbb  P_0(\V{x}_0) 
\lbb \prod_{a=1}^\rn \exp \lb-\beta J \lb\V{x}_a\rb  \rb  \rbb 
\lbb \prod_{a=1}^\rn 
\delta\lb Q-\frac{1}{N}\V{x}_a^{\top}\V{x}_a  \rb 
\delta\lb m-\frac{1}{N}\V{x}_0^{\top}\V{x}_a  \rb 
\rbb
\no \\ &&
\times
\lbb \prod_{a<b\cap B(a)=B(b)} \delta \lb q_1-\frac{1}{N}\V{x}_a^{\top}\V{x}_b \rb  \rbb\lbb \prod_{a<b\cap B(a)\neq B(b)} \delta\lb q_0-\frac{1}{N} \V{x}_a^{\top}\V{x}_{b} \rb \rbb 
\no \\ &&
=
\int d\tilde{Q}d\tilde{q}_1d\tilde{q}_0d\tilde{m}
e^{
N\lbb  \frac{1}{2}  \rn \tilde{Q} Q-\frac{1}{2} \tilde{q}_1 q_1 \rn(\PPc-1)-\frac{1}{2}  \tilde{q}_0 q_0 \rn(\rn-\PPc)- \rn \tilde{m} m \rbb}.
\no \\ &&
\times
\lb
\int d x_0 P_0\lb x_0 \rb  \int D z_0
\left\{
\int Dz_{1}
\lb \int d x~
e^{ \beta g \lb x,\frac{\tilde{Q}+\tilde{q}_1}{\beta},\frac{h_{\rm 1RSB}}{\beta} \rb }
\rb ^{\PPc}
\right\}^{\frac{\rn}{\PPc}}
\rb^{N}
,
\Leq{entropic}
\ee 
\subee
where
\be
h_{\rm 1RSB}=\sqrt{\tilde{q}_1-\tilde{q}_0} z_{1}+\sqrt{\tilde{q}_0} z_0+\tilde{m} x_0.
\ee
To arrive at the last line of \Req{entropic}, we used several standard computational tricks such as the decomposition of the delta function using the Fourier transform
\be
\delta\lb Q-\frac{1}{N}\V{x}_a^{\top}\V{x}_a  \rb =\int d\tilde{Q}e^{\frac{\tilde{Q}}{2}\lb Q-\frac{1}{N}\V{x}_a^{\top}\V{x}_a  \rb},
\ee
and the Hubbard-Stratonovich transformation
\be
e^{\frac{1}{2}a^2}=\int Dz e^{az }.
\ee
Summarizing these contributions and using the saddle-point method, we have the following form of the 1RSB free energy  $f_{\rm 1RSB}=-\beta^{-1}\E_{\V{x}_0,\V{A}}\frac{1}{N}\log Z_{\beta,\tau}=-\beta^{-1}\lim_{\rn \to 0}\Part{}{\rn}{}\psi(p)$ in the $\tau\to 0$ limit as 
\be
&&
-f_{\rm 1RSB}
=\beta^{-1}\Extr{\Omega,\T{\Omega}}\Biggl\{
-\frac{\alpha}{2} \ln \lb Q-q_1\rb
-\frac{\alpha}{2 \PPc} \ln \lb 1+\frac{\PPc\lb q_1-q_0\rb}{Q-q_1}\rb-\frac{\alpha}{2} \frac{\varepsilon_{q_0}}{Q-q_1+\PPc\lb q_1-q_0\rb} 
\no \\ &&
+\frac{1}{2} \tilde{Q} Q-\tilde{m} m
-\frac{1}{2} \tilde{q}_1 q_1(\PPc-1)+\frac{1}{2} \tilde{q}_0 q_0 \PPc 
\no \\ &&
+\frac{1}{\PPc} 
\E_{x_0,z_0}
\log \left\{\int D z_{1} 
\lb \int d x~
e^{ \beta g \lb x,\frac{\tilde{Q}+\tilde{q}_1}{\beta},\frac{h_{\rm 1RSB}}{\beta} \rb }
\rb ^{\PPc}
\right\}
\Biggr\},
\Leq{f_1RSB}
\ee
where $\E_{x_0,z_0}$ is already given in \Req{ave_single_x0z0}, $\Extr{x}$ denotes the extremization condition w.r.t. $x$, and $\Omega=\{Q,q_1,q_0,m\},\T{\Omega}=\{\T{Q},\T{q}_1,\T{q}_0,\T{m}\}$ summarize the order parameters of the current system.

To take the limit $\beta \to \infty$, we assume the following scaling:
\be
\chi=\beta(Q-q_1),\quad
\Gamma=\frac{\tilde{Q}+\tilde{q}_1}{\beta},\quad 
\beta^2 B+\beta^2 \bar{q}_0=\tilde{q}_1,\quad
\beta^2 \bar{q}_0=\tilde{q}_0,\quad 
\beta \bar{m}=\tilde{m},\quad
\rPPc=\beta \PPc. 
\ee
Inserting these relations into \Req{f_1RSB} and taking the limit $\beta \to \infty$, we have 
\be
&&
-f_{\rm 1RSB}
=\Extr{\chi,Q,q_0,m,\Gamma,B,\bar{q}_0,\bar{m}}
\Biggl\{
\frac{\alpha}{2 \rPPc} \ln  \frac{\chi}{\chi+\rPPc\lb Q-q_0\rb}
-\frac{\alpha}{2} \frac{\varepsilon_{q_0}}{\chi+\rPPc\lb Q-q_0\rb} 
\no \\ &&
+\frac{1}{2}(\Gamma-\rPPc (B+\bar{q}_0)) Q -\frac{1}{2}(B+\bar{q}_0)\chi
-\bar{m} m
+\frac{1}{2}\rPPc  \bar{q}_0 q_0 
\no \\ &&
+\frac{1}{\rPPc} 
\E_{x_0,z_0}
\log 
\lb \int D z_{1} 
e^{ \rPPc \hat{g} \lb \Gamma, \bar{h}_{\rm 1RSB} \rb }
\rb
\Biggr\},
\Leq{f_1RSB_T0}
\ee
where
\be
\bar{h}_{\rm 1RSB}=\sqrt{B}z_1+\sqrt{\bar{q}_0}z_0+\bar{m}x_0.
\ee
Taking the extremization condition, the equations of state (EOS) determining the order parameters are obtained as follows:
\subbe
\Leq{EOS_1RSB}
\be
&&
\Gamma  = \frac{\alpha}{\chi},
\Leq{Gamma_1RSBEOS}
 \\
&&
\bar{m}   =\frac{\alpha}{\chi+\rPPc\left(Q -q_0\right)},
\Leq{barm_1RSBEOS}
 \\
&&
 B  =\frac{\alpha}{\chi}\lb \frac{Q-q_0}{\chi+\rPPc\left(Q-q_0\right)} \rb,  
\Leq{B_1RSBEOS}
\\
&&
\bar{q}_0  =\frac{\alpha \epsilon_{q_0}}{ \lb \chi+\rPPc\left(Q- q_0\right)\rb^2 }, 
\Leq{barq0_1RSBEOS}
\\
&&
\chi
=\E_{x_0,z_0}\Ave{\partial_{\bar{h}_{\rm 1RSB }} \hat{x}(\Gamma,\bar{h}_{\rm 1RSB })  }_{z_1 \mid x_0,z_0},
\Leq{1RSB_chi}
\\ &&
m
=\E_{x_0,z_0}\Ave{x_0 \hat{x}(\Gamma,\bar{h}_{\rm 1RSB })  }_{z_1 \mid x_0,z_0},
\Leq{1RSB_m}
\\ &&
Q
=\E_{x_0,z_0}\Ave{ \hat{x}^2(\Gamma,\bar{h}_{\rm 1RSB })  }_{z_1 \mid x_0,z_0},
\Leq{1RSB_Q}
\\ &&
q_{0}
=\E_{x_0,z_0}\Ave{ \hat{x}(\Gamma,\bar{h}_{\rm 1RSB })  }_{z_1 \mid x_0,z_0}^2,
\Leq{1RSB_q0}
\ee
\subee
and the above average is defined as  
\be
\Ave{
(\cdots)
}_{z_1 \mid x_0,z_0}
=
\frac{
\int Dz_1 e^{\rPPc \hat{g}\lb \Gamma,\bar{h}_{\rm 1RSB} \rb}
(\cdots)
}{
\int Dz_1 e^{\rPPc \hat{g} \lb \Gamma^{(t)},\bar{h}_{\rm 1RSB} \rb}
},
\ee
which corresponds to \Req{ave_single_t}. 

The correspondence between \Reqs{SE_1RSB}{EOS_1RSB} is very clear: In both expressions, identical symbols denote the same quantities, and the effective field $\bar{h}_{\rm 1RSB}$ in the replica formula corresponds to $(\sqrt{B}z_1+C(x_0,z_0))$ in the 1RSB-SE equations; although the replica EOS do not explicitly contain MSE $\epsilon_{q_0}$, it is straightforward to obtain its expression using the formulas from $m$ and $q_0$, yielding
\be
\epsilon_{q_{0}}
=\E_{x_0,z_0} \lb x_0- \Ave{ \hat{x}(\Gamma,\bar{h}_{\rm 1RSB } )}_{z_1 \mid x_0,z_0} \rb^2 ,
\ee
matching to \Req{SE_1RSB_MSE}. Besides, it is known that the 1RSB free energy $f_{\rm 1RSB}$ in the replica approach and the free entropy $s$ derived from the Bethe free energy in the replica extension framework has the following relation~\cite{PhysRevLett.75.2847,Obuchi2015FiniteReplica}:
\be
- \rPPc f_{\rm 1RSB}=s.
\ee
This relation can be directly checked by inserting \Reqss{Gamma_1RSBEOS}{barq0_1RSBEOS} in \Req{f_1RSB_T0}. These correspondences support the correctness of our 1RSB message passing, and further clarify that the algorithm is closely connected to the global properties of the system.

One benefit of the replica formula in the current context comes from the variational structure of the free energy w.r.t. the order parameters. For computing $u^*$ in \Req{u^*-1RSBSE}, we need to differentiate $s$ w.r.t $\rPPc$, but thanks to the variational structure, on the replica formula we may leave the order parameters undifferentiated and take the partial derivative solely on $\rPPc$. This leads to the following formula:
\be
&&
u^*(\rPPc)
=\frac{\alpha}{2}\lbb \frac{Q-q_0}{Y}+\frac{\epsilon_{q_0}}{Y}-\frac{\rPPc \epsilon (Q-q_0)}{Y^2}\rbb
\no \\ &&
-\frac{1}{2}\lb \Gamma- 2 \rPPc (B+\bar{q}_0) \rb Q+\frac{1}{2}(B+\bar{q}_0)\chi-\rPPc \bar{q}_0q_0+\bar{m}m
-\E_{x_0,z_0}\lsb \Ave{\hat{g}(\Gamma,\bar{h}_{\rm 1RSB})}_{z_1\mid x_0,z_0} \rsb.
\ee
Hence, no numerical derivative is necessary to compute $u^*$. The complexity is computed by inserting this into \Req{complexity}.

\section{Computational issues in the 1RSB equations}\Lsec{Computational}
Let us compute
\be
\ul{\mc{Z}}(\Gamma,B,C)
=\int Dz e^{\rPPc \hat{g}(\Gamma,\sqrt{B}z+C)}
=\sum_{k \in \{\RI,\ldots,\RIV\}} L_k
\ee
where
\be
L_k=\int Dze^{\rPPc \hat{g}(\Gamma,\sqrt{B}z+C)}
\ind\lb 
\sqrt{B}z+C \in k
\rb,
\quad k \in \{\RI,\ldots,\RIV\}.
\ee
Let us specify each term:
\be
&&
L_{\RI}
=\int Dz e^{\rPPc \hat{g}} \ind\lb |\sqrt{B}z+C|>  a\lambda \Gamma \rb
=e^{
-\frac{1}{2}\rPPc \lambda^2(a+1)
}
\lbb
\int_{z_{1+}}^{\infty} Dz e^{\frac{1}{2}\frac{\ell}{\Gamma}h^2}
+
\int_{-\infty}^{z_{1-}} Dz e^{\frac{1}{2}\frac{\ell}{\Gamma}h^2}
\rbb,
\\ &&
L_{\RII}
=e^{
\frac{1}{2}\rPPc \frac{\lambda^2}{(a-1)}
}
\lbb
\int_{z_{2+}}^{z_{1+}} Dz 
e^{\frac{1}{2}\frac{\ell}{\Gamma-(a-1)^{-1}}\lb h-\frac{a\lambda }{a-1}\rb^2}
+
\int_{z_{1-}}^{z_{2-}} Dz 
e^{\frac{1}{2}\frac{\ell}{\Gamma-(a-1)^{-1}}\lb h+\frac{a\lambda }{a-1}\rb^2}
\rbb,
\\ &&
L_{\RIII}
=
\lbb
\int_{z_{3+}}^{z_{2+}} Dz 
e^{\frac{1}{2}\frac{\ell}{\Gamma}\lb h-\lambda \rb^2}
+
\int_{z_{2-}}^{z_{3-}} Dz 
e^{\frac{1}{2}\frac{\ell}{\Gamma}\lb h+\lambda \rb^2}
\rbb,
\\
 &&
L_{\RIV}
=
\int_0^{z_{3+}} Dz 
+
\int_{z_{3-}}^{0} Dz 
=H(z_{3+})-H(z_{3-}),
\ee
where $h=\sqrt{B}z+C$ and
\be
&&
z_{1 \pm}=\frac{1}{\sqrt{B}}(\pm a\lambda \Gamma -C)
, \quad
z_{2 \pm}=\frac{1}{\sqrt{B}}(\pm \lambda (1+\Gamma) -C)
, \quad
z_{3 \pm}=\frac{1}{\sqrt{B}}(\pm \lambda -C),
\\ &&
H(a)\coloneqq \int_0^{a}Dz=\frac{1}{2}\erf\lb \frac{a}{\sqrt{2}} \rb.
\ee
To handle this, we define
\be
W_0\lb a,b,\Lambda,k,\mu \rb=\int_{a}^{b}Dz e^{\frac{1}{2}\Lambda (kz-\mu)^2}
,\quad
W_{0\infty}\lb a,\Lambda,k,\mu \rb
=
W\lb a,\infty,\Lambda,k,\mu \rb.
\ee
Depending on $D=1-\Lambda k^2$, this integral has a different convenient form:
\be
W_0\lb a,b,\Lambda,k,\mu \rb
=
\left\{
\begin{array}{cc}
\frac{e^{\frac{\Lambda \mu^2}{2D}}}{2\sqrt{D}}
\lbb 
\erf\lb 
\frac{bD+k\mu \Lambda }{\sqrt{2D}}
\rb
-
\erf\lb 
\frac{aD+k\mu \Lambda }{\sqrt{2D}}
\rb
\rbb
,  & D>0,  \\
\frac{e^{\frac{\Lambda \mu^2}{2D}}}{2\sqrt{|D|}}
\lbb 
\erfi\lb 
\frac{aD+k\mu \Lambda }{\sqrt{2|D|}}
\rb
-
\erfi\lb 
\frac{bD+k\mu \Lambda }{\sqrt{2|D|}}
\rb
\rbb,
& D<0, \\
\frac{e^{\frac{1}{2} \Lambda \mu^2}}{k\mu \Lambda\sqrt{2\pi}} \lb e^{-ak \mu \Lambda}-e^{-bk \mu \Lambda}\rb,  &  D=0 \& \mu\neq 0, \\
\frac{b-a}{\sqrt{2\pi}},  &   D=0 \& \mu=0, 
\end{array}
\right.,
\ee
where 
\be
\erfi(x)=\frac{\erf(ix)}{i}=\frac{2}{\sqrt{\pi}}\int_0^{x}e^{t^2}dt.
\ee
In our case, $D=1-\Lambda k^2=1-\rPPc B/(\Gamma)$ or $1-\rPPc B/(\Gamma-(a-1)^{-1})$ and  it is expected that $D>0$ always holds. Hence we assume $D>0$ in the following. Also, the contribution from $\RI$ yields 
\be
W_{0\infty}\lb a,\Lambda,k,\mu \rb
=
\frac{e^{\frac{\Lambda \mu^2}{2D}}}{2\sqrt{D}}
\lbb 
1-
\erf\lb 
\frac{aD+k\mu \Lambda }{\sqrt{2D}}
\rb
\rbb.
\ee
Using these, we can write
\subbe
\be
&&
L_{\RI}=e^{-\frac{1}{2}\rPPc\lambda^2(a+1)}
\lbb
W_{0\infty}\lb z_{1+},\frac{\rPPc}{\Gamma},\sqrt{B},-C\rb
+
W_{0\infty}\lb -z_{1-},\frac{\rPPc}{\Gamma},\sqrt{B},C\rb
\rbb,
\\ &&
L_{\RII}
=e^{
\frac{1}{2}\rPPc \frac{\lambda^2}{(a-1)}
}
\Biggl\{
W_{0}\lb z_{2+},z_{1+},\frac{\rPPc}{\Gamma-(a-1)^{-1}},\sqrt{B},-C+\frac{a\lambda}{a-1}\rb
\no \\ &&
+
W_{0}\lb z_{1-},z_{2-},\frac{\rPPc}{\Gamma-(a-1)^{-1}},\sqrt{B},-C-\frac{a\lambda}{a-1}\rb
\Biggr\}
,
\\ &&
L_{\RIII}=
W_{0}\lb z_{3+},z_{2+},\frac{\rPPc}{\Gamma},\sqrt{B},-C+\lambda\rb
+
W_{0}\lb z_{2-},z_{3-},\frac{\rPPc}{\Gamma},\sqrt{B},-C-\lambda\rb,
\\ &&
L_{\RIV}
=H(z_{3+})-H(z_{3-}),
\ee
\subee

In computing $\Ave{\hat{x}}$, we have some integrals of the completing-square form. For example in $\RII$, 
\be
&&
\Ave{\hat{x}}
\overset{\RII}{\propto}
\frac{e^{
\frac{1}{2(a-1)}\rPPc \lambda^2
}}{
\Gamma-(a-1)^{-1}
}
\Biggl\{
\int_{z_{2+}}^{z_{1+}} Dz 
\lb h-\frac{a\lambda }{a-1}\rb
e^{\frac{1}{2}\frac{\ell}{\Gamma-(a-1)^{-1}}\lb h-\frac{a\lambda }{a-1}\rb^2}
\no \\ &&
+
\int_{z_{1-}}^{z_{2-}} Dz 
\lb h+\frac{a\lambda }{a-1}
\rb
e^{\frac{1}{2}\frac{\ell}{\Gamma-(a-1)^{-1}}\lb h+\frac{a\lambda }{a-1}\rb^2}
\Biggr\}.
\ee
This motivates us to define
\be
&&
W_j \lb a,b,\Lambda,k,\mu \rb
=\int_{a}^{b}Dz (kz-\mu)^j e^{\frac{1}{2}\Lambda (kz-\mu)^2}
,\quad
W_{j\infty}\lb a,\Lambda,k,\mu \rb
=
W_j\lb a,\infty,\Lambda,k,\mu \rb,
\\ &&
E_{j}(z,\Lambda,k,\mu)=\frac{(kz-\mu)^j e^{-\frac{1}{2}z^2+\frac{1}{2}\Lambda(kz-\mu)^2 } }{\sqrt{2\pi}}.
\ee
The cases $j=1,2$ are necessary for computing $\Ave{\hat{x}},\Ave{\hat{x}^2}$. According to the integration by parts, the case $j=1$ yields
\be
&&
W_1
=
\int_{a}^{b}Dz (kz-\mu) e^{\frac{1}{2}\Lambda (kz-\mu)^2}
=
-k \int_{a}^{b} (Dz)'  e^{\frac{1}{2}\Lambda (kz-\mu)^2}- \mu W_0
\no \\ &&
=-k \lsb \frac{e^{-\frac{1}{2}z^2 +\frac{1}{2}\Lambda (kz-\mu)^2}}{\sqrt{2\pi}} \rsb_{a}^{b}
+k \int_{a}^{b} Dz  e^{\frac{1}{2}\Lambda (kz-\mu)^2}\Lambda k(kz-\mu)
- \mu W_0
\no \\ &&
=
-k(E_0(b)-E_0(a))+\Lambda k^2W_1-\mu W_0
\no \\ &&
\therefore
W_1(a,b,\Lambda,k,\mu)
=-\frac{1}{1-\Lambda k^2}\lbb k(E_0(b)-E_0(a))+\mu W_0(a,b,\Lambda,k,\mu)\rbb.
\ee
Similarly
\be
&&
W_2
=
\int_{a}^{b}Dz (kz-\mu)^2 e^{\frac{1}{2}\Lambda (kz-\mu)^2}
=
 \int_{a}^{b} Dz \lb kz(kz-\mu)-\mu(kz-\mu)\rb  e^{\frac{1}{2}\Lambda (kz-\mu)^2}
 \no \\ &&
=- k \int_{a}^{b} (Dz)' (kz-\mu)  e^{\frac{1}{2}\Lambda (kz-\mu)^2}-\mu W_1
=-k \lsb (kz-\mu) \frac{e^{-\frac{1}{2}z^2 +\frac{1}{2}\Lambda (kz-\mu)^2}}{\sqrt{2\pi}} \rsb_{a}^{b}
\no \\ &&
+k \int_{a}^{b} Dz  \lbb ke^{\frac{1}{2}\Lambda (kz-\mu)^2}+(kz-\mu)e^{\frac{1}{2}\Lambda (kz-\mu)^2}\Lambda k(kz-\mu)  \rbb 
- \mu W_1
\no \\ &&
=
-k(E_1(b)-E_1(a))+\Lambda k^2W_2+k^2 W_0-\mu W_1
\no \\ &&
\hspace{-1cm}
\therefore
W_2(a,b,\Lambda,k,\mu)
=-\frac{1}{1-\Lambda k^2}\lbb k(E_1(b)-E_1(a))+\mu W_1(a,b,\Lambda,k,\mu)-k^2W_0(a,b,\Lambda,k,\mu)\rbb.
\ee
Using these functions, we can express the averaged quantities as
\be
&&
\ul{\mc{Z}}(\Gamma,B,C)
\Ave{\hat{x}}_{\Gamma,B,C}
= 
\int Dz \hat{x}(\Gamma,\sqrt{B}z+C)e^{\rPPc \hat{g}(\Gamma,\sqrt{B}z+C)}
\no \\ &&
= 
\sum_{k\in \{\RI,\ldots,\RIV \}}\int_{k} Dz \hat{x}(\Gamma,\sqrt{B}z+C)e^{\rPPc \hat{g}(\Gamma,\sqrt{B}z+C)}
\eqqcolon 
\sum_{k\in \{\RI,\ldots,\RIV \}}L_{1k}.
\ee
The explicit formulas are
\subbe
\be
&&
L_{1\RI}
=\frac{e^{
-\frac{1}{2}\rPPc \lambda^2(a+1)
}}{\Gamma}
\lbb
\int_{z_{1+}}^{\infty} Dz  h e^{\frac{1}{2}\frac{\ell}{\Gamma}h^2}
+
\int_{-\infty}^{z_{1-}} Dz h e^{\frac{1}{2}\frac{\ell}{\Gamma}h^2}
\rbb
\no \\ &&
=\frac{e^{
-\frac{1}{2}\rPPc \lambda^2(a+1)
}}{\Gamma}
\lbb
W_{1\infty}\lb z_{1+},\frac{\rPPc}{\Gamma},\sqrt{B},-C\rb
-
W_{1\infty}\lb -z_{1-},\frac{\rPPc}{\Gamma},\sqrt{B},C\rb
\rbb,
\\ &&
L_{1\RII}
=
\frac{e^{
\frac{1}{2}\rPPc \frac{\lambda^2}{(a-1)}
}}{
\Gamma-(a-1)^{-1}
}
\Biggl\{
W_1\lb z_{2+},z_{1+},\frac{\rPPc}{\Gamma-(a-1)^{-1}},\sqrt{B},-C+\frac{a\lambda}{a-1}\rb
\no \\ &&
+
W_1\lb z_{1-},z_{2-},\frac{\rPPc}{\Gamma-(a-1)^{-1}},\sqrt{B},-C-\frac{a\lambda}{a-1}\rb
\Biggr\},
\\ &&
L_{1\RIII}
=
\frac{1}{\Gamma}\lbb
\int_{z_{3+}}^{z_{2+}} Dz 
(h-\lambda)e^{\frac{1}{2}\frac{\ell}{\Gamma}\lb h-\lambda \rb^2}
+
\int_{z_{2-}}^{z_{3-}} Dz 
(h+\lambda)
e^{\frac{1}{2}\frac{\ell}{\Gamma}\lb h+\lambda \rb^2}
\rbb
\no\\ &&
=
\frac{1}{\Gamma}\lbb
W_{1}\lb z_{3+},z_{2+},\frac{\rPPc}{\Gamma},\sqrt{B},-C+\lambda \rb
+
W_{1}\lb z_{2-},z_{3-},\frac{\rPPc}{\Gamma},\sqrt{B},-C-\lambda \rb
\rbb,
\\ &&
L_{1\RIV}=0.
\ee
\subee
Very similarly
\be
&&
\ul{\mc{Z}}(\Gamma,B,C)
\Ave{\hat{x}^2}_{\Gamma,B,C}
= 
\sum_{k\in \{\RI,\ldots,\RIV \}}\int_{k} Dz \hat{x}^2(\Gamma,\sqrt{B}z+C)e^{\rPPc \hat{g}(\Gamma,\sqrt{B}z+C)}
\eqqcolon 
\sum_{k\in \{\RI,\ldots,\RIV \}}L_{2k}.
\ee
The explicit formulas are
\subbe
\be
&&
L_{2\RI}
=\frac{e^{
-\frac{1}{2}\rPPc \lambda^2(a+1)
}}{\Gamma^2}
\lbb
\int_{z_{1+}}^{\infty} Dz  h^2 e^{\frac{1}{2}\frac{\ell}{\Gamma}h^2}
+
\int_{-\infty}^{z_{1-}} Dz h^2 e^{\frac{1}{2}\frac{\ell}{\Gamma}h^2}
\rbb
\no \\ &&
=\frac{e^{
-\frac{1}{2}\rPPc \lambda^2(a+1)
}}{\Gamma^2}
\lbb
W_{2\infty}\lb z_{1+},\frac{\rPPc}{\Gamma},\sqrt{B},-C\rb
+
W_{2\infty}\lb -z_{1-},\frac{\rPPc}{\Gamma},\sqrt{B},C\rb
\rbb,
\\ &&
L_{2\RII}
=
\frac{e^{
\frac{1}{2}\rPPc \frac{\lambda^2}{(a-1)}
}}{
(\Gamma-(a-1)^{-1})^2
}
\Biggl\{
W_2\lb z_{2+},z_{1+},\frac{\rPPc}{\Gamma-(a-1)^{-1}},\sqrt{B},-C+\frac{a\lambda}{a-1}\rb
\no \\ &&
+
W_2\lb z_{1-},z_{2-},\frac{\rPPc}{\Gamma-(a-1)^{-1}},\sqrt{B},-C-\frac{a\lambda}{a-1}\rb
\Biggr\},
\\ &&
L_{2\RIII}
=
\frac{1}{\Gamma^2}\lbb
W_{2}\lb z_{3+},z_{2+},\frac{\rPPc}{\Gamma},\sqrt{B},-C+\lambda \rb
+
W_{2}\lb z_{2-},z_{3-},\frac{\rPPc}{\Gamma},\sqrt{B},-C-\lambda \rb
\rbb,
\\ &&
L_{2\RIV}=0.
\ee
\subee
Again similarly
\be
&&
\ul{\mc{Z}}(\Gamma,B,C)
\Ave{\partial_{C}\hat{x}}_{\Gamma,B,C}
= 
\sum_{k\in \{\RI,\ldots,\RIV \}}\int_{k} Dz \partial_C\hat{x}(\Gamma,\sqrt{B}z+C)e^{\rPPc \hat{g}(\Gamma,\sqrt{B}z+C)}
\no \\ &&
=
\frac{1}{\Gamma}L_{\RI}+\frac{1}{\Gamma-(a-1)^{-1}}L_{\RII}+\frac{1}{\Gamma}L_{\RIII}+0.
\ee

\paragraph{Summary of how to perform integration}
\subbe
\be
&&
\ul{\mc{Z}}(\Gamma,B,C)
=\sum_{k\in \{\RI,\ldots,\RIV \}}L_k
,\quad
\Ave{\partial_C \hat{x}}_{\Gamma,B,C}=\frac{\frac{L_{\RI}+L_{\RIII}}{\Gamma}+\frac{L_{\RII}}{\Gamma-(a-1)^{-1}}}{\sum_{k\in \{\RI,\ldots,\RIV \}}L_k},
\Leq{how2_Z-chi_1RSB}
\\ &&
\hspace{-0cm}
\Ave{\hat{x}}_{\Gamma,B,C}=\frac{\sum_{k\in \{\RI,\ldots,\RIV \}}L_{1k}}{\sum_{k\in \{\RI,\ldots,\RIV \}}L_k},
\quad
\Ave{\hat{x}^2}_{\Gamma,B,C}=\frac{\sum_{k\in \{\RI,\ldots,\RIV \}}L_{2k}}{\sum_{k\in \{\RI,\ldots,\RIV \}}L_k},
\Leq{how2_xave}
\\ &&
L_{\RI}=e^{-\frac{1}{2}\rPPc\lambda^2(a+1)}
\lbb
W_{0\infty}\lb z_{1+},\frac{\rPPc}{\Gamma},\sqrt{B},-C\rb
+
W_{0\infty}\lb -z_{1-},\frac{\rPPc}{\Gamma},\sqrt{B},C\rb
\rbb,
\\ &&
L_{\RII}
=e^{
\frac{1}{2}\rPPc \frac{\lambda^2}{(a-1)}
}
\Biggl\{
W_{0}\lb z_{2+},z_{1+},\frac{\rPPc}{\Gamma-(a-1)^{-1}},\sqrt{B},-C+\frac{a\lambda}{a-1}\rb
\no \\ &&
+
W_{0}\lb z_{1-},z_{2-},\frac{\rPPc}{\Gamma-(a-1)^{-1}},\sqrt{B},-C-\frac{a\lambda}{a-1}\rb
\Biggr\}
,
\\ &&
L_{\RIII}=
W_{0}\lb z_{3+},z_{2+},\frac{\rPPc}{\Gamma},\sqrt{B},-C+\lambda\rb
+
W_{0}\lb z_{2-},z_{3-},\frac{\rPPc}{\Gamma},\sqrt{B},-C-\lambda\rb,
\\ &&
L_{\RIV}
=H(z_{3+})-H(z_{3-}),
\\ &&
L_{1\RI}
=\frac{e^{
-\frac{1}{2}\rPPc \lambda^2(a+1)
}}{\Gamma}
\lbb
W_{1\infty}\lb z_{1+},\frac{\rPPc}{\Gamma},\sqrt{B},-C\rb
-
W_{1\infty}\lb -z_{1-},\frac{\rPPc}{\Gamma},\sqrt{B},C\rb
\rbb,
\\ &&
L_{1\RII}
=
\frac{e^{
\frac{1}{2}\rPPc \frac{\lambda^2}{(a-1)}
}}{
\Gamma-(a-1)^{-1}
}
\Biggl\{
W_1\lb z_{2+},z_{1+},\frac{\rPPc}{\Gamma-(a-1)^{-1}},\sqrt{B},-C+\frac{a\lambda}{a-1}\rb
\no \\ &&
+
W_1\lb z_{1-},z_{2-},\frac{\rPPc}{\Gamma-(a-1)^{-1}},\sqrt{B},-C-\frac{a\lambda}{a-1}\rb
\Biggr\},
\\ &&
L_{1\RIII}
=
\frac{1}{\Gamma}\lbb
W_{1}\lb z_{3+},z_{2+},\frac{\rPPc}{\Gamma},\sqrt{B},-C+\lambda \rb
+
W_{1}\lb z_{2-},z_{3-},\frac{\rPPc}{\Gamma},\sqrt{B},-C-\lambda \rb
\rbb,
\\ &&
L_{1\RIV}=0,
\\ &&
L_{2\RI}
=\frac{e^{
-\frac{1}{2}\rPPc \lambda^2(a+1)
}}{\Gamma^2}
\lbb
W_{2\infty}\lb z_{1+},\frac{\rPPc}{\Gamma},\sqrt{B},-C\rb
+
W_{2\infty}\lb -z_{1-},\frac{\rPPc}{\Gamma},\sqrt{B},C\rb
\rbb,
\\ &&
L_{2\RII}
=
\frac{e^{
\frac{1}{2}\rPPc \frac{\lambda^2}{(a-1)}
}}{
(\Gamma-(a-1)^{-1})^2
}
\Biggl\{
W_2\lb z_{2+},z_{1+},\frac{\rPPc}{\Gamma-(a-1)^{-1}},\sqrt{B},-C+\frac{a\lambda}{a-1}\rb
\no \\ &&
+
W_2\lb z_{1-},z_{2-},\frac{\rPPc}{\Gamma-(a-1)^{-1}},\sqrt{B},-C-\frac{a\lambda}{a-1}\rb
\Biggr\},
\\ &&
L_{2\RIII}
=
\frac{1}{\Gamma^2}\lbb
W_{2}\lb z_{3+},z_{2+},\frac{\rPPc}{\Gamma},\sqrt{B},-C+\lambda \rb
+
W_{2}\lb z_{2-},z_{3-},\frac{\rPPc}{\Gamma},\sqrt{B},-C-\lambda \rb
\rbb,
\\ &&
L_{2\RIV}=0.
\\ &&
z_{1 \pm}=\frac{1}{\sqrt{B}}(\pm a\lambda \Gamma -C)
, \quad
z_{2 \pm}=\frac{1}{\sqrt{B}}(\pm \lambda (1+\Gamma) -C)
, \quad
z_{3 \pm}=\frac{1}{\sqrt{B}}(\pm \lambda -C),
\ee
\subee

\paragraph{Summary of function definitions (such as $W,H$)}
\subbe
\be
&&
H(a)\coloneqq \int_0^{a}Dz=\frac{1}{2}\erf\lb \frac{a}{\sqrt{2}} \rb,
\\ &&
E_{j}(z,\Lambda,k,\mu)=\frac{(kz-\mu)^j e^{-\frac{1}{2}z^2+\frac{1}{2}\Lambda(kz-\mu)^2 } }{\sqrt{2\pi}},
\\ &&
D=1-\Lambda k^2,
\\ &&
W_0\lb a,b,\Lambda,k,\mu \rb
=
\left\{
\begin{array}{cc}
\frac{e^{\frac{\Lambda \mu^2}{2D}}}{2\sqrt{D}}
\lbb 
\erf\lb 
\frac{bD+k\mu \Lambda }{\sqrt{2D}}
\rb
-
\erf\lb 
\frac{aD+k\mu \Lambda }{\sqrt{2D}}
\rb
\rbb
,  & D>0  \\
\frac{e^{\frac{\Lambda \mu^2}{2D}}}{2\sqrt{|D|}}
\lbb 
\erfi\lb 
\frac{aD+k\mu \Lambda }{\sqrt{2|D|}}
\rb
-
\erfi\lb 
\frac{bD+k\mu \Lambda }{\sqrt{2|D|}}
\rb
\rbb,
& D<0 \\
\frac{e^{\frac{1}{2} \Lambda \mu^2}}{k\mu \Lambda\sqrt{2\pi}} \lb e^{-ak \mu \Lambda}-e^{-bk \mu \Lambda}\rb,  &  D=0 \& \mu\neq 0 \\
\frac{b-a}{\sqrt{2\pi}},  &   D=0 \& \mu=0 
\end{array}
\right.,
\\ &&
W_{0\infty}\lb a,\Lambda,k,\mu \rb
=
\frac{e^{\frac{\Lambda \mu^2}{2D}}}{2\sqrt{D}}
\lbb 
1-
\erf\lb 
\frac{aD+k\mu \Lambda }{\sqrt{2D}}
\rb
\rbb,
\\ &&
W_1(a,b,\Lambda,k,\mu)
=-\frac{1}{D}\lbb k(E_0(b,\Lambda,k,\mu)-E_0(a,\Lambda,k,\mu))+\mu W_0(a,b,\Lambda,k,\mu)\rbb,
\\ &&
W_{1\infty}(a,\Lambda,k,\mu)
=-\frac{1}{D}\lbb -k E_0(a,\Lambda,k,\mu)+\mu W_{0\infty}(a,\Lambda,k,\mu)\rbb,
\\ &&
W_2(a,b,\Lambda,k,\mu)
=-\frac{1}{D}
\Biggl\{
k(E_1(b,\Lambda,k,\mu)-E_1(a,\Lambda,k,\mu))
\no \\ &&
+\mu W_1(a,b,\Lambda,k,\mu)-k^2W_0(a,b,\Lambda,k,\mu)
\Biggr\},
\\ &&
W_{2\infty}(a,\Lambda,k,\mu)
=-\frac{1}{D}
\Biggl\{
-k E_1(a,\Lambda,k,\mu)
+\mu W_{1\infty}(a,\Lambda,k,\mu)-k^2W_{0\infty}(a,\Lambda,k,\mu)
\Biggr\}
.
\ee
\subee

We found that 1RSB-AMP or 1RSB-SE using these functions tend to suffer from NaN (not a number) in later steps of the algorithm. We inspected this and found that NaN  comes from the largeness of $C$ (that comes into $\mu,a,b$ of $W$ functions and $z,\mu$ in $E$ functions) leading the divergence of $e^{\frac{1}{2}\Lambda \mu^2}$. To prevent this, we considered to divide both the numerator and denominator in \Reqs{how2_Z-chi_1RSB}{how2_xave} by a common factor: The concrete idea is to divide by $e^{\frac{1}{2}\frac{\rPPc}{\Gamma-1/(a-1) }C^2}$. To implement this, the following modified functions and algorithms are introduced.
\paragraph{Summary of precision-aware function definitions (such as $W,H$)}
\subbe
\be
&&
\T{E}_{j}(z,\Lambda,k,\mu,\Lambda_2,C)=\frac{(kz-\mu)^j e^{-\frac{1}{2}z^2+\frac{1}{2}\Lambda(kz-\mu)^2 -\frac{1}{2}\Lambda_2 C^2 } }{\sqrt{2\pi}},
\\ &&
D=1-\Lambda k^2,
\\ &&
\T{W}_0\lb a,b,\Lambda,k,\mu ,\Lambda_2,C\rb
=
\left\{
\begin{array}{cc}
\frac{e^{\frac{\Lambda \mu^2}{2D} -\frac{1}{2}\Lambda_2 C^2}  }{2\sqrt{D}}
\lbb 
\erf\lb 
\frac{bD+k\mu \Lambda }{\sqrt{2D}}
\rb
-
\erf\lb 
\frac{aD+k\mu \Lambda }{\sqrt{2D}}
\rb
\rbb
,  & D>0  \\
\frac{e^{\frac{\Lambda \mu^2}{2D}-\frac{1}{2}\Lambda_2 C^2} }{2\sqrt{|D|}}
\lbb 
\erfi\lb 
\frac{aD+k\mu \Lambda }{\sqrt{2|D|}}
\rb
-
\erfi\lb 
\frac{bD+k\mu \Lambda }{\sqrt{2|D|}}
\rb
\rbb,
& D<0 \\
\frac{e^{\frac{1}{2} \Lambda \mu^2 -\frac{1}{2}\Lambda_2 C^2}}{k\mu \Lambda\sqrt{2\pi}} \lb e^{-ak \mu \Lambda}-e^{-bk \mu \Lambda}\rb,  &  D=0 \& \mu\neq 0 \\
e^{-\frac{1}{2}\Lambda_2 C^2}\frac{b-a}{\sqrt{2\pi}},  &   D=0 \& \mu=0 
\end{array}
\right.,
\\ &&
\T{W}_{0\infty}\lb a,\Lambda,k,\mu ,\Lambda_2,C\rb
=
\frac{e^{\frac{\Lambda \mu^2}{2D} -\frac{1}{2}\Lambda_2 C^2} }{2\sqrt{D}}
\lbb 
1-
\erf\lb 
\frac{aD+k\mu \Lambda }{\sqrt{2D}}
\rb
\rbb,
\\ &&
\T{W}_1(a,b,\Lambda,k,\mu,\Lambda_2,C)
=-\frac{1}{D}\biggl\{ k(\T{E}_0(b,\Lambda,k,\mu,\Lambda_2,C)-\T{E}_0(a,\Lambda,k,\mu),\Lambda_2,C)
\no \\ &&
+\mu \T{W}_0(a,b,\Lambda,k,\mu,\Lambda_2,C)
\biggr\}
,
\\ &&
\T{W}_{1\infty}(a,\Lambda,k,\mu,\Lambda_2,C)
=-\frac{1}{D}\lbb -k \T{E}_0(a,\Lambda,k,\mu,\Lambda_2,C)+\mu \T{W}_{0\infty}(a,\Lambda,k,\mu,\Lambda_2,C)\rbb,
\\ &&
\T{W}_2(a,b,\Lambda,k,\mu,\Lambda_2,C)
=-\frac{1}{D}
\Biggl\{
k(\T{E}_1(b,\Lambda,k,\mu,\Lambda_2,C)-\T{E}_1(a,\Lambda,k,\mu,\Lambda_2,C))
\no \\ &&
+\mu \T{W}_1(a,b,\Lambda,k,\mu,\Lambda_2,C)-k^2\T{W}_0(a,b,\Lambda,k,\mu,\Lambda_2,C)
\Biggr\},
\\ &&
\T{W}_{2\infty}(a,\Lambda,k,\mu,\Lambda_2,C)
=-\frac{1}{D}
\Biggl\{
-k \T{E}_1(a,\Lambda,k,\mu,\Lambda_2,C)
+\mu \T{W}_{1\infty}(a,\Lambda,k,\mu,\Lambda_2,C)
\no \\ &&
-k^2\T{W}_{0\infty}(a,\Lambda,k,\mu,\Lambda_2,C)
\Biggr\}
.
\ee
\subee

\paragraph{Summary of how to perform precision-aware integration}
\subbe
\be
&&
\ul{\mc{Z}}(\Gamma,B,C)
=\sum_{k\in \{\RI,\ldots,\RIV \}}\T{L}_k
,\quad
\Ave{\partial_C \hat{x}}_{\Gamma,B,C}=\frac{\frac{\T{L}_{\RI}+\T{L}_{\RIII}}{\Gamma}+\frac{\T{L}_{\RII}}{\Gamma-(a-1)^{-1}}}{\sum_{k\in \{\RI,\ldots,\RIV \}}\T{L}_k},
\Leq{how2_Z-chi_1RSB_precision}
\\ &&
\hspace{-0cm}
\Ave{\hat{x}}_{\Gamma,B,C}=\frac{\sum_{k\in \{\RI,\ldots,\RIV \}}\T{L}_{1k}}{\sum_{k\in \{\RI,\ldots,\RIV \}}\T{L}_k},
\quad
\Ave{\hat{x}^2}_{\Gamma,B,C}=\frac{\sum_{k\in \{\RI,\ldots,\RIV \}}\T{L}_{2k}}{\sum_{k\in \{\RI,\ldots,\RIV \}}\T{L}_k},
\Leq{how2_xave_precision}
\\ &&
\Lambda_2=\frac{\rPPc}{\Gamma-1/(a-1)},
\\ &&
\T{L}_{\RI}=e^{-\frac{1}{2}\rPPc\lambda^2(a+1)}
\lbb
\T{W}_{0\infty}\lb z_{1+},\frac{\rPPc}{\Gamma},\sqrt{B},-C,\Lambda_2,C\rb
+
\T{W}_{0\infty}\lb -z_{1-},\frac{\rPPc}{\Gamma},\sqrt{B},C,\Lambda_2,C\rb
\rbb,
\\ &&
\T{L}_{\RII}
=e^{
\frac{1}{2}\rPPc \frac{\lambda^2}{(a-1)}
}
\Biggl\{
\T{W}_{0}\lb z_{2+},z_{1+},\frac{\rPPc}{\Gamma-(a-1)^{-1}},\sqrt{B},-C+\frac{a\lambda}{a-1},\Lambda_2,C\rb
\no \\ &&
+
\T{W}_{0}\lb z_{1-},z_{2-},\frac{\rPPc}{\Gamma-(a-1)^{-1}},\sqrt{B},-C-\frac{a\lambda}{a-1},\Lambda_2,C\rb
\Biggr\}
,
\\ &&
\T{L}_{\RIII}=
\T{W}_{0}\lb z_{3+},z_{2+},\frac{\rPPc}{\Gamma},\sqrt{B},-C+\lambda,\Lambda_2,C\rb
+
\T{W}_{0}\lb z_{2-},z_{3-},\frac{\rPPc}{\Gamma},\sqrt{B},-C-\lambda,\Lambda_2,C\rb,
\\ &&
\T{L}_{\RIV}
=\lbb H(z_{3+})-H(z_{3-}) \rbb e^{-\frac{1}{2}\Lambda_2 C^2},
\\ &&
\T{L}_{1\RI}
=\frac{e^{
-\frac{1}{2}\rPPc \lambda^2(a+1)
}}{\Gamma}
\lbb
\T{W}_{1\infty}\lb z_{1+},\frac{\rPPc}{\Gamma},\sqrt{B},-C,\Lambda_2,C\rb
-
\T{W}_{1\infty}\lb -z_{1-},\frac{\rPPc}{\Gamma},\sqrt{B},C,\Lambda_2,C\rb
\rbb,
\\ &&
\T{L}_{1\RII}
=
\frac{e^{
\frac{1}{2}\rPPc \frac{\lambda^2}{(a-1)}
}}{
\Gamma-(a-1)^{-1}
}
\Biggl\{
\T{W}_1\lb z_{2+},z_{1+},\frac{\rPPc}{\Gamma-(a-1)^{-1}},\sqrt{B},-C+\frac{a\lambda}{a-1},\Lambda_2,C\rb
\no \\ &&
+
\T{W}_1\lb z_{1-},z_{2-},\frac{\rPPc}{\Gamma-(a-1)^{-1}},\sqrt{B},-C-\frac{a\lambda}{a-1},\Lambda_2,C\rb
\Biggr\},
\\ &&
\T{L}_{1\RIII}
=
\frac{1}{\Gamma}\lbb
\T{W}_{1}\lb z_{3+},z_{2+},\frac{\rPPc}{\Gamma},\sqrt{B},-C+\lambda ,\Lambda_2,C\rb
+
\T{W}_{1}\lb z_{2-},z_{3-},\frac{\rPPc}{\Gamma},\sqrt{B},-C-\lambda ,\Lambda_2,C\rb
\rbb,
\\ &&
\T{L}_{1\RIV}=0,
\\ &&
\T{L}_{2\RI}
=\frac{e^{
-\frac{1}{2}\rPPc \lambda^2(a+1)
}}{\Gamma^2}
\lbb
\T{W}_{2\infty}\lb z_{1+},\frac{\rPPc}{\Gamma},\sqrt{B},-C,\Lambda_2,C\rb
+
\T{W}_{2\infty}\lb -z_{1-},\frac{\rPPc}{\Gamma},\sqrt{B},C,\Lambda_2,C\rb
\rbb,
\\ &&
\T{L}_{2\RII}
=
\frac{e^{
\frac{1}{2}\rPPc \frac{\lambda^2}{(a-1)}
}}{
(\Gamma-(a-1)^{-1})^2
}
\Biggl\{
\T{W}_2\lb z_{2+},z_{1+},\frac{\rPPc}{\Gamma-(a-1)^{-1}},\sqrt{B},-C+\frac{a\lambda}{a-1},\Lambda_2,C\rb
\no \\ &&
+
\T{W}_2\lb z_{1-},z_{2-},\frac{\rPPc}{\Gamma-(a-1)^{-1}},\sqrt{B},-C-\frac{a\lambda}{a-1},\Lambda_2,C\rb
\Biggr\},
\\ &&
\T{L}_{2\RIII}
=
\frac{1}{\Gamma^2}\lbb
\T{W}_{2}\lb z_{3+},z_{2+},\frac{\rPPc}{\Gamma},\sqrt{B},-C+\lambda ,\Lambda_2,C\rb
+
\T{W}_{2}\lb z_{2-},z_{3-},\frac{\rPPc}{\Gamma},\sqrt{B},-C-\lambda ,\Lambda_2,C\rb
\rbb,
\\ &&
\T{L}_{2\RIV}=0.
\\ &&
z_{1 \pm}=\frac{1}{\sqrt{B}}(\pm a\lambda \Gamma -C)
, \quad
z_{2 \pm}=\frac{1}{\sqrt{B}}(\pm \lambda (1+\Gamma) -C)
, \quad
z_{3 \pm}=\frac{1}{\sqrt{B}}(\pm \lambda -C).
\ee
\subee

\paragraph{How to reduce 2-dim integration into 1-dim one}
In the SE equations,the two-dimensional integrations over $x_0,z_0$ are required. This can be converted into one-dimensional integration, which reduces the computational cost a lot. Here its prescription is explained. 

Since $x_0$ obeys the Bernoulli-Gaussian distribution \NReq{prior}, \Req{SE_1RSB^{(t)}chi} can be decomposed as 
\be
&&
\chi^{(t)}=
(1-\rho )
\E_{z_0}\lb 
\Ave{\partial_C\hat{x}(\Gamma^{(t)},\sqrt{B^{(t)}}z_1+C^{(t)}(0,z_0))}_{{\Gamma}^{(t)},B^{(t)},C^{(t)}}
\rb
\no \\ &&
+
\rho 
\E_{x_0,z_0}\lb 
\Ave{\partial_C\hat{x}(\Gamma^{(t)},\sqrt{B^{(t)}}z_1+C^{(t)}(x_0,z_0))}_{{\Gamma}^{(t)},B^{(t)},C^{(t)}}
\rb.
\ee
The first term is obtained by performing the 1-dim Gaussian integration over $z_0$. The second term requires the two-dim Gaussian integration literally, but since $(x_0,z_0)$ appears only through the function 
\be
C^{(t+1)}(x_0,z_0)
=\frac{\alpha }{\chi^{(t)}+\rPPc (Q^{(t)}-q_{0}^{(t)})}x_0
+\frac{\sqrt{\alpha \epsilon_{q_0}^{(t)}}}{\chi^{(t)}+\rPPc (Q^{(t)}-q_{0}^{(t)})}z_0
\overset{\textrm{d}}{=}
\frac{\sqrt{\alpha^2\sigma_x^2+\alpha \epsilon_{q_0}^{(t)} }}{\chi^{(t)}+\rPPc (Q^{(t)}-q_{0}^{(t)})}z_0
,
\ee
this two-dim Gaussian can be converted into a one-dimensional Gaussian integration over $z_0\sim \mc{N}(0,1)$. Hence we may write
\be
&&
\chi^{(t)}=
(1-\rho )
\E_{z_0}\lb 
\Ave {\partial_C\hat{x}(\Gamma^{(t)},\sqrt{B^{(t)}}z_1+C^{(t)}(0,z_0))
}^{(t)}
\rb
\no \\ &&+
\rho 
\E_{z_0}\lb 
\Biggl\langle
\partial_C\hat{x}
\lb \Gamma^{(t)},\sqrt{B^{(t)}}z_1+C^{(t)}\lb 0,\frac{\sqrt{\alpha^2\sigma_x^2+\alpha \epsilon^{(t-1)}_{q_0}}}{\sqrt{\alpha \epsilon^{(t-1)}_{q_0}}}z_0\rb 
\rb
\Biggr\rangle^{(t)}
\rb
\\ &&
\eqqcolon
(1-\rho )  (\chi^{(t)} \mid x_0=0)+\rho   (\chi^{(t)}\mid x_0\neq 0)
.
\ee
The same computation and notation are applied to $Q^{(t)},q_{0}^{(t)}$.

The situation is different for $\epsilon_{q_0}^{(t)}$ since $x_0$ appears not only through $C$. But some algebras yield
\be
&&
\epsilon_{q_0}^{(t)}
=
(1-\rho ) \E_{z_0}\lsb \lb \Ave{\hat{x}}^{(t)}\rb^2 \rsb
+
\rho 
\E_{x_0\sim \mc{N}(0,\sigma_x^2),z_0}\lsb \lb x_0-\Ave{\hat{x}}^{(t)}\rb^2 \rsb
\no \\ &&
=
(1-\rho ) \lb q_{0}^{(t)}\mid x_0=0\rb
+
\rho \sigma_x^2+\rho \lb q_{0}^{(t)}\mid x_0\neq 0\rb
-2\rho 
\E_{x_0\sim \mc{N}(0,\sigma_x^2),z_0}\lsb x_0\Ave{\hat{x}}^{(t)} \rsb 
\no \\ &&
=q_{0}^{(t)}
+
\rho \sigma_x^2
-2\rho  \sigma_x^2 
\E_{x_0\sim \mc{N}(0,\sigma_x^2),z_0}\lsb  \partial_{x_{0}}\Ave{\hat{x}}^{(t)} \rsb.
\ee
The last term is obtained by the integration by part. Thanks to \Req{m_diff_C}, the last factor can be simplified as
\be
&&
\E_{x_0\sim \mc{N}(0,\sigma_x^2),z_0}\lsb  \partial_{x_{0}}\Ave{\hat{x}}^{(t)} \rsb
=
\E_{x_0\sim \mc{N}(0,\sigma_x^2),z_0}\lsb  \Part{C^{(t)}}{x_0}{}\partial_{C^{(t)}}\Ave{\hat{x}}^{(t)} \rsb
\no \\ &&
=
\frac{\alpha}{\chi_{t-1}+\rPPc(Q_{t-1}-q_{0,t-1})}\E_{x_0\sim \mc{N}(0,\sigma_x^2),z_0}\lsb  
\Ave{\partial_{C^{(t)}}\hat{x}}^{(t)}+\rPPc\lb \lb \Ave{\hat{x}^2}^{(t)}-\Ave{\hat{x}}^{(t)}\rb^2\rb
 \rsb
\no \\ &&
=
\frac{\alpha}{\chi_{t-1}+\rPPc(Q_{t-1}-q_{0,t-1})}
\lbb  
\lb \chi^{(t)}\mid x_0\neq 0\rb
+
\rPPc
\lb 
\lb Q^{(t)}\mid x_0\neq 0\rb
-
\lb q_{0}^{(t)}\mid x_0\neq 0\rb
\rb
\rbb.
\ee
In summary, 
\be
&&
\epsilon_{q_0}^{(t)}
=q_{0}^{(t)}
+
\rho \sigma_x^2
-2 \frac{\alpha \rho  \sigma_x^2}{\chi_{t-1}+\rPPc(Q_{t-1}-q_{0,t-1})}
\lbb  
\lb \chi^{(t)}\mid x_0\neq 0\rb
+
\rPPc
\lb 
\lb Q^{(t)}\mid x_0\neq 0\rb
-
\lb q_{0}^{(t)}\mid x_0\neq 0\rb
\rb
\rbb.
\ee
By this way, the numerical integration gets easier.

\paragraph{Behavior in the large $C$ limit}
In the large $C$ limit, we expect the contribution from $\RI$ dominates. In the case of $\Ave{\hat{x}}$ and $C \to +\infty$, we have
\be
&&
\Ave{\hat{x}}_{\Gamma,B,C}\approx \frac{L_{1\RI}}{L_{\RI}}.
\\ &&
L_{\RI}=e^{\frac{1}{2}\rPPc\lambda^2 (a+1)}
\lbb 
W_{0\infty}\lb z_{1+},\frac{\rPPc}{\Gamma},\sqrt{B},-C\rb
+
W_{0\infty}\lb -z_{1-},\frac{\rPPc}{\Gamma},\sqrt{B},C\rb
\rbb,
\\ &&
L_{1\RI}=\frac{e^{\frac{1}{2}\rPPc\lambda^2 (a+1)}}{\Gamma}
\lbb 
W_{1\infty}\lb z_{1+},\frac{\rPPc}{\Gamma},\sqrt{B},-C\rb
-
W_{1\infty}\lb -z_{1-},\frac{\rPPc}{\Gamma},\sqrt{B},C\rb
\rbb,
\\ &&
z_{1\pm } \to -\infty \therefore 
W_{0,1\infty}\lb -z_{1-},\frac{\rPPc}{\Gamma},\sqrt{B},C\rb
\propto
\int_{-z_{1-}}^{\infty} Dz(\cdots)
\to 0,
\\ &&
\therefore
\Ave{\hat{x}}_{\Gamma,B,C}\approx 
\frac{1}{\Gamma}
\frac{W_{1\infty}\lb z_{1+},\frac{\rPPc}{\Gamma},\sqrt{B},-C\rb
}{W_{0\infty}\lb z_{1+},\frac{\rPPc}{\Gamma},\sqrt{B},-C\rb
}
\\ &&
\approx 
\frac{1}{\Gamma}
\frac{
\int_{-\infty}^{\infty}Dz(\sqrt{B}z+C)e^{\frac{1}{2}\frac{\rPPc}{\Gamma}(\sqrt{B}z+C)^2}
}{
\int_{-\infty}^{\infty}Dz e^{\frac{1}{2}\frac{\rPPc}{\Gamma}(\sqrt{B}z+C)^2}
}
=
\frac{C}{\Gamma}
+
\frac{\sqrt{B}}{\Gamma}
\frac{
\int_{-\infty}^{\infty}Dz z e^{\frac{1}{2}\frac{\rPPc}{\Gamma}(\sqrt{B}z+C)^2}
}{
\int_{-\infty}^{\infty}Dz e^{\frac{1}{2}\frac{\rPPc}{\Gamma}(\sqrt{B}z+C)^2}
}
\\ &&
=
\frac{C}{\Gamma}
+
\frac{\rPPc BC}{\Gamma^2(1-B\rPPc/\Gamma)}
.
\ee
The second term yields a bias significantly deviating from the signal term $C/\Gamma$. Hence, $\Ave{\hat{x}}_{\Gamma,B,C}$ has, if it is regarded as an estimator to $x_{0}$, a strong bias as long as $B$ does not vanish. 

\bibliographystyle{unsrt}
\bibliography{cs}

\end{document}